%% file: ese_main.tex
\RequirePackage{fix-cm}
\documentclass[smallcondensed]{svjour3}     
\smartqed  
\usepackage{graphicx}

\usepackage{mathptmx}      
\usepackage[round]{natbib}

\usepackage{latexsym}

\usepackage{amsmath,amssymb}
\usepackage{commath}

\usepackage[caption=false]{subfig}

\usepackage{nameref,hyperref}

\usepackage[table]{xcolor}
\usepackage{epstopdf}
\AppendGraphicsExtensions{.tiff}

\usepackage{array}
\usepackage{multirow} 

\usepackage{tikz-qtree}

\newcounter{RQCounter}
\newcommand{\RQ}[2]{%
\vspace{0.05in}                                                                                                               
\refstepcounter{RQCounter} \label{#1}
	\vspace{0.04in}
    \noindent \textbf{RQ\arabic{RQCounter}.}~{\em #2  }                                                                                                        
\vspace{0.05in}                                                                                                               
}

\newenvironment{myquote}[1]%
  {\list{}{\leftmargin=#1\rightmargin=#1}\item[]}%
  {\endlist}

\begin{document}

\title{Studying the Difference Between Natural and Programming Language Corpora
}


\author{Casey Casalnuovo         \and
        Kenji Sagae \and
        Prem Devanbu 
}


\institute{Casey Casalnuovo \at
               Department of Computer Science, University of California, Davis, CA, USA \\
              \email{ccasal@ucdavis.edu}           
           \and
           Kenji Sagae \at
              Department of Linguistics, University of California, Davis, CA, USA \\
              \email{sagae@ucdavis.edu}
           \and
           Prem Devanbu \at
              Department of Computer Science, University of California, Davis, CA, USA \\
              \email{ptdevanbu@ucdavis.edu}
}

\date{Received: date / Accepted: date}

\maketitle

\begin{abstract}
Code corpora, as observed in large 
software systems, are now known
to be far more repetitive and predictable than natural language corpora.
But why?  Does the difference simply arise from the syntactic limitations of programming
languages? Or does it arise from the differences in
authoring decisions made by the writers of these natural and programming language texts?
We conjecture that the differences are \emph{not} entirely due to syntax, but also 
from the fact that reading and writing code
is \emph{un}-natural for humans, and requires substantial mental effort; so, people prefer
to write code in ways that are familiar to both reader and writer. To support this argument,
we present results from two sets of studies: 1) a first set aimed at
attenuating the effects of syntax, and 2) a second, aimed at measuring
repetitiveness of text written in other 
settings (\emph{e.g.} second language, technical/specialized jargon), which are
also effortful to write. 
We find find that this repetition in source code is not entirely the result of grammar constraints,
and thus some repetition must result from human choice.  While the evidence we find of similar repetitive behavior
in technical and learner corpora does not conclusively show that such language is used by humans to
mitigate difficulty, it is consistent with that theory. 

\keywords{Language Modeling \and Programming Languages \and Natural Languages \and Syntax \& Grammar \and Parse Trees \and Corpus Comparison}
\end{abstract}


\section{Introduction}
\label{sec:intro}
\input{intro.tex}
\section{Theory}
\label{sec:theory}
\input{theory.tex}

\section{Materials and methods}
\label{sec:method}
\input{method.tex}
\section{Results}
\label{sec:result}
\input{result.tex}
\section{Discussion}
\label{sec:discussion}
\input{discussion.tex}

\begin{acknowledgements}
We would like to thank Professors Charles Sutton, Zhendong Su, Vladimir Filkov, and Raul Aranovich, along with the UC Davis DECAL and NLP Reading groups for comments and feedback on this research.
We also would like to especially thank Vincent Hellendoorn for his feedback and input on our experiment between parse trees in Java and English.
We also acknowledge support from NST Grant \#1414172, Exploiting the Naturalness of Software.
\end{acknowledgements}

\bibliographystyle{spbasic}      
\bibliography{codenlp.bib}   

\end{document}

%% file: intro.tex
Source code is often viewed as being primarily intended for machines 
to interpret and execute.
However, source code is not just an interlocutory medium between human and machine, but also a form of communication between humans - a view advocated by Donald Knuth:

\begin{myquote}{0.3cm}
\emph{\small Instead of imagining that our main task is to instruct a computer 
what to do, let us concentrate rather on explaining to human beings what we 
want a computer to do~\citep{knuth1984literate}}. 
\end{myquote}

\begin{figure}[t]
\centering
   \includegraphics[width=1.0\columnwidth]{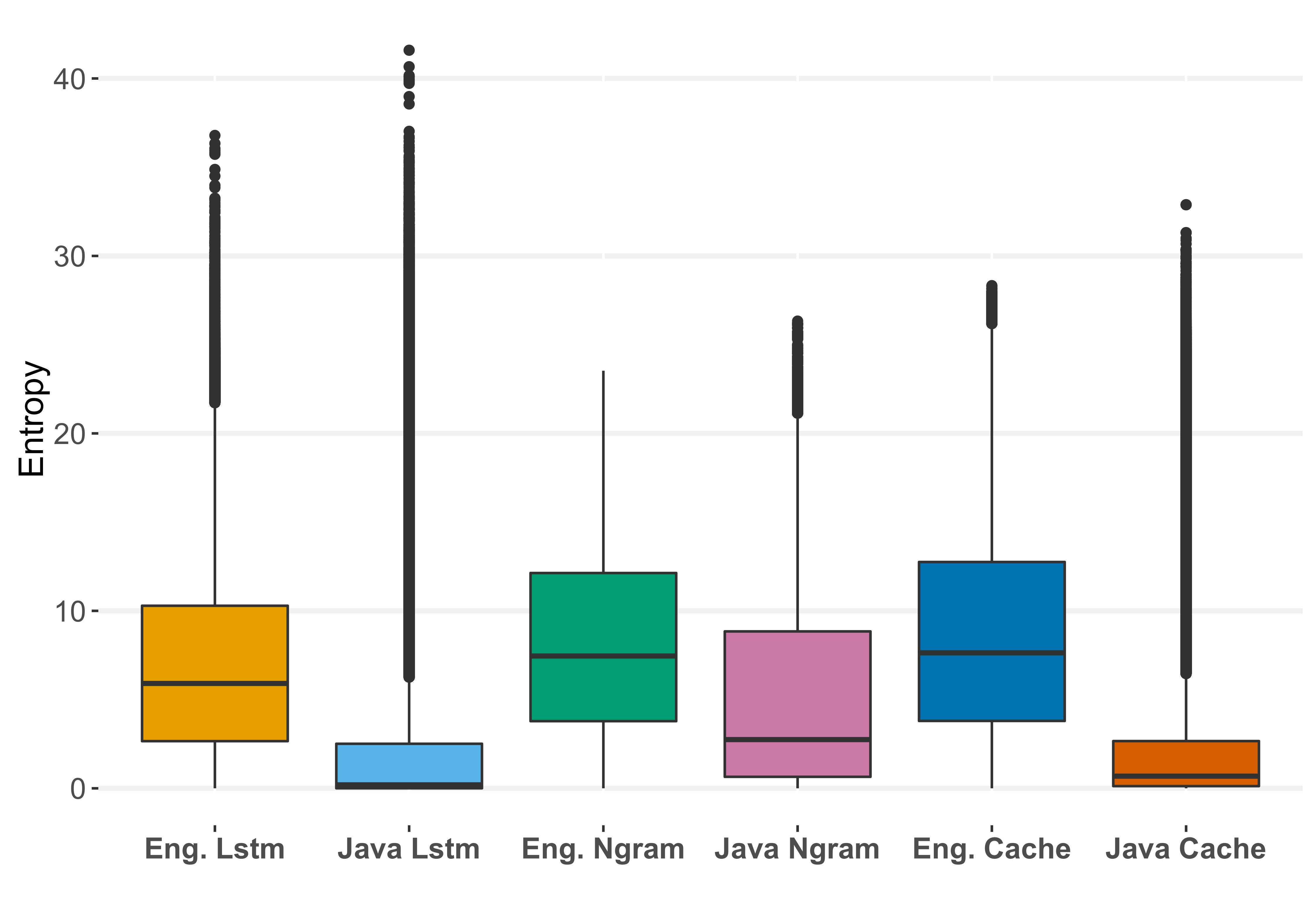}
  \caption{Entropy comparisons of English and Java corpora from 3 different language models}
  \label{fig:IntroEntropy}
\end{figure}

Software development is largely a team effort; 
code that cannot be understood and maintained will be likely not endure. 
It's  well known that most development time is spent in maintenance
rather than \emph{di novo} coding~\citep{SoftMaintLehman}.
Thus it's very reasonable to consider source code as a form of human communication,
amenable to the same sorts of statistical language models  (LM)
developed for natural language.
This hypothesis was originally conceived by Hindle et al. \citep{Hindle2012}, 
who showed that LM designed for natural language were actually \textit{more effective} for code, than in their original context. 
Hindle \emph{et al} used basic ngram language models to capture repetition in code;
subsequent, more advanced models, tuned for modular structure~\citep{Tu2014,hellendoorn2017deep}, 
and deep learning approaches such as LSTMs \citep{hochreiter1997long} 
yield even better results. 
Fig~\ref{fig:IntroEntropy} demonstrates this difference on corpora of Java and English, 
using the standard entropy measure~\citep{manning1999foundations} over a held-out test set. 
A lower entropy value indicates that a token was less surprising for the language model.
These box plots display the entropy for each token in the test set, and show that (regardless of model)
Java is more predictable than English\footnote{Precise details on the datasets and language models will be presented later their respective sections.}.

But why is code more predictable? 
The difference could either arise from a) \emph{inherent syntactic} 
differences between natural and programming languages
or b)
the \textit{contingent authoring} choices made by authors.
Source code grammars are unambiguous, for ease of parsing; 
this limitation might account for the greater predictability of code. 
But there may be other reasons; 
perhaps source code is more  domain-specific; perhaps
developers
deliberately limit their constructions to a smaller set of highly reused forms, 
just to deal with the great cognitive challenges of code reading and writing.

\smallskip

This leaves us with 2 questions:
\begin{enumerate}
\item How much does programming language \emph{syntax} influence repetitiveness in coding?
and 
\item
What are the \emph{contingent} factors  (not constrained by syntax) that play a role in code repetitiveness?
 \end{enumerate}
We address the first question, with experiments breaking down the syntactic differences between
source code and natural language. The second question is very open-ended; to constrain it, we consider a variant: 
\begin{enumerate}
  \setcounter{enumi}{1}
  \item  \emph{Is repetitiveness observed in code also observed in other natural language 
corpora that similarly required significant effort from the creators}? 
\end{enumerate}

We address this question, 
with corpora of text that are similarly ``effortful'' for the writers (or readers, or both) or have potentially
higher costs of miscommunication: we consider English as a second language and in specialized corpora 
such as legal or technical writing. To summarize our results, we find:
\begin{itemize}
\item The differences between source code and English, observed previously in Java hold true in many different programming and natural languages. 
\item
Programming language corpora are more similar to each other than to English, although Haskell appears somewhat more like English than the others. 
\item  Even when accounting for grammar and syntax in different ways, Java is statistically significantly more repetitive than English. 
\item  ESL (English as a Second language) corpora, as well as technical, imperative, and legal corpora, do exhibit repetitiveness  
similar to that seen in code corpora. 
\end{itemize}

These suggest that differences observed between natural and programming languages are not entirely due to grammatical limitations,
and that code is also more repetitive due to contingent facts -- i.e. humans \emph{choose} to write code more repetitively than English.

%% file: theory.tex
By \emph{syntax}, we mean the aspects of language related
to structure and grammar, rather than  meaning.
Both code (an artificial language) and natural language have syntactic constraints. 
Code has intentionally simplified grammar, to facilitate language
learning, and to enable
efficient parsing by compilers. 
Human languages 
have evolved naturally; grammars for natural languages 
are ``just'' theories of linguistic phenomena, that aren't always
followed, and in general, are more complex, non-deterministic, and ambiguous 
than code grammars. 

A language's syntax constrains the set of valid utterances. 
The more restrictive the grammar, the less choice in utterances. 
Thus,  it's possible that the entropy differences between code and NL 
arise entirely out of the more restrictive grammar for code.
If so, the observed differences
aren't a result of conscious choice  by humans
to write code more repetitively;  it's just the grammar. 

However, if we could explicitly account for the syntactic differences between English
and code, and \emph{\underline {still}} find that code is repetitive, 
then the unexplained difference could well 
arise from deliberate choices made by programmers. 
Below, we explore a couple theories of why the syntax of source code may be more repetitive than the syntax of natural language.

\subsection{Syntactic Explanations}
\label{sec:syntax_theory}

\subsubsection{Open And Closed Vocabulary Words }
\label{sec:open_closed_voc_theory}

Languages evolve; 
with time, certain word categories grow more than others. 
We can call categories of words where new words are easily and frequently added \emph{open category}
(\emph{e.g.}, nouns, verbs, adjectives).
As the corpus grows, we can expect to see more and more open category words. 
\emph{Closed category} vocabulary, however is limited; 
no matter how big the corpus, the set of distinct words in these categories 
is fixed and limited.\footnote{While this category is very rarely updated, there could be
unusual and significant changes in the language - for instance a new preposition or conjunction in English.}
In English, closed category words include conjunctions, articles, and pronouns.  
This categorization of English vocabulary 
is well-established~\citep{bradley1978computational}, and  
we adapt this analogously for source code. 

In code, 
reserved words, like \textit{for}, \textit{if}, or \textit{public} form a closed set of language-specific keywords whose usage is syntactically limited. The
arithmetic and logical operators (which combine elements in code like
conjunctions in English) also constitute closed vocabulary.
Code also has 
\textit{punctuation}, like ``;" which demarcates sequences of expressions, statements, etc. 
These categories are slightly different from those studied by Petersen et al., \citep{petersen2012languages} who
consider a kernel or core vocabulary, and an unlimited vocabulary to which new words were added.
Our definitions are tied to syntax rather than semantics, hingeing on the type of word (e.g. noun vs conjunction or identifier vs reserved word)
rather than how core the meaning of the word is to the expressibility of the language.
Closed vocabulary words are necessarily part of the kernel lexicon they describe, but open category words will appear in both the kernel and
unlimited vocabulary. 
For example, the commonly used iterator \emph{i} would be in the kernel vocabulary in most programming languages, but  other identifiers like \emph{registeredStudent} could fall under Petersen's unlimited lexicon.

Closed vocabulary tokens relate intimately 
to the underlying structure and grammar of a language, 
whereas open vocabulary tokens relate more to the content.
A fixed number of (closed category)
markers defining how to structure the content of a 
message in a language are sufficient, since they relate more
to structure than to topic or content. 
In contrast, new nouns, verbs, adverbs, and adjectives in English, 
or types and identifiers in Java are constantly invented to express new ideas
in new contexts. 
Since closed category words relate more to syntax, we would expect that
the corpus that remains after removing them (\emph{viz.,} the open-category corpus)
would be more reflective of content,
and less of the actual syntax. Thus  analyzing the open-category corpus
(for code and English) would allow us to judge the repetitiveness that arises
more from content-related choices made by the authors, rather than merely
from syntax \emph{per se}. Removal of closed category words to focus on content
rather than form is then similar to the removal of \emph{stop words} 
(frequently occurring words that are considered of no or low value to a particular task)
in natural language processing.

Thus, our first experiment addresses the question:

\RQ{keywords}{How much does removing closed category words 
affect the difference in repetitiveness and predictability 
between Java and English?}

\subsection{Ambiguity in Language}
\label{sec:ambig_theory}

Grammatical structure transcends keyword or closed-category word usage.
Programming language grammars are intentionally unambiguous, 
whereas natural languages are rife with grammatical ambiguity. 
Compilers must be able to easily parse source code;  
syntactic ambiguity in code also impedes reading \& debugging. 
For example, in the C language, there are constructs that 
produce undefined behavior (See Hathhorn et al.  \citep{hathhorn2015defining}).
Different compilers might adopt different semantics, 
thus vitiating portability.

Various theories for explaining 
the greater ambiguity in natural language 
have been proposed.
One camp, led by Chomsky, asserts that 
ambiguity in language arises from NL being adapted not for purely communicative 
purposes, but for cognitive efficiency \citep{chomsky2002interview}.

Others have argued that ambiguity is desirable for communication. 
Zipf \citep{zipf49} argued that ambiguity arises from a trade off between speakers and listeners:
ambiguity reduces speaker effort. In the extreme case if one word expressed all possible meanings then ease of speaking would be minimized; however, listeners would prefer less ambiguity.  
If humans are able to disambiguate more easily, then some  ambiguity could naturally arise. Others argue ambiguity could arise from memory limitations or applications in inter-dialect communication \citep{wasow2005puzzle}.
A variant of Zipf's argument is presented by Piantadosi et al. \citep{piantadosi2012communicative}:
since ambiguity is often resolvable from context, 
efficient language systems will allow ambiguity in some cases. 
They empirically demonstrated that words which are more frequent and shorter in length, tend to possess more meanings than infrequent and longer words.

The ambiguity is widely prevalent in natural language, both in word meaning and in sentence structure.
Works like ``take'' have \emph{polysemy}, or many meanings. 
Syntactic structure (even without polysemic words) can lead to ambiguity. 
One popular example of ambiguous sentence structure is that of prepositional attachment.
Consider the sentence: 

\begin{myquote}{0.3cm}
\emph{They saw the building with a telescope.}
\end{myquote}
\noindent
There are two meanings, depending on where the phrase \textit{with a telescope} attaches: did they see using the telescope, or is the telescope mounted on the building? 
Both meanings are valid, where one or the other may be preferred based on the context.

Such ambiguous sentences
can be resolved using 
a \emph{constituency parse tree} or \emph{CPT} -- representing natural language
in a way similar to how an \emph{AST} represents source code.
~A \emph{CPT} is built from nested units, building up to a root node that represents the whole sentence (typically represented with \emph{S} or \emph{ROOT}).
The terminal nodes are the words of the original sentence, and the non-terminals include parts of speech (nouns/verbs) and phrase labels 
(noun phrases, verb phrases, prepositional phrases, etc).
While there is no definitive set of non-terminals used of labeling English sentences, some 
sets are very commonly used, such as the one designed for the Penn Treebank \citep{Marcus:1993}.

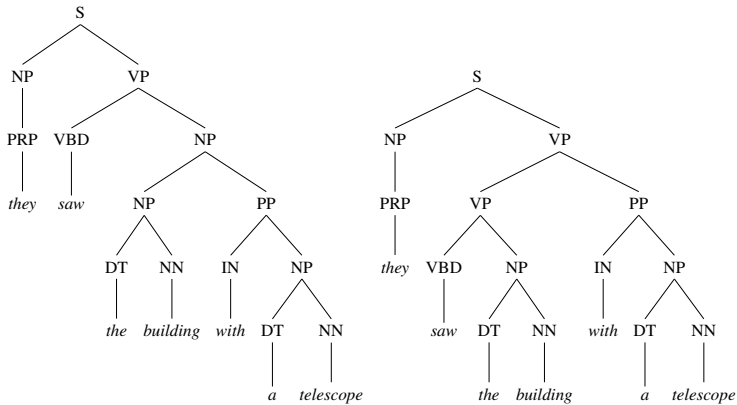
\begin{figure}[ht]
\centering
\begin{tikzpicture}[scale=.8]
\Tree[.S [.NP [.PRP \textit{they} ]]
	[.VP [.VBD \textit{saw} ]
	[.NP [.NP [.DT \textit{the} ] [.NN \textit{building} ]]
	[.PP [.IN \textit{with} ] [.NP [.DT \textit{a} ] [.NN \textit{telescope} ]]]]]]
\end{tikzpicture}
\begin{tikzpicture}[scale=.8]
\Tree[.S [.NP [.PRP \textit{they} ]]
	[.VP [.VP [.VBD \textit{saw} ] [.NP [.DT \textit{the} ] [.NN \textit{building} ]]]
	[.PP [.IN \textit{with} ] [.NP [.DT \textit{a} ] [.NN \textit{telescope} ]]]]]
\end{tikzpicture}
  \caption{Two parse trees for the sentence \textit{They saw the building with a telescope}.
  The tree on the left corresponds to the the reading that the telescope is part of the building; on the right, to the reading
  that the viewing was done with a telescope}
  \label{fig:AmbParse}
\end{figure}

A \emph{CPT} \emph{fully} resolves syntactic ambiguities:
\emph{e.g.,} consider Fig. \ref{fig:AmbParse}, which shows the two possible \emph{CPTs} 
for our example sentence. 
While the raw text is ambiguous, each of the 
\emph{CPTs} fully resolve and clarify the different possible
meanings; only one meaning is possible for a given \emph{CPT}. \
In source code, however, the syntactic structure is unambiguous, given
the raw tokens. 

Source code syntax is represented using a  similar hierarchical
construction: the \emph{abstract syntax tree} or \emph{AST}.
However, \emph{ASTs} differ from \emph{CPTs} in that they exclude some
tokens of the original text, that are inferable from context.
Both trees unambiguously represent
structure in natural language and source code.
In section \ref{sec:EqPartTree}, we will discuss how we modified these slightly to further improve their comparability.

Using such trees, we can revisit the question of whether the greater repetitiveness and predictability of source code arises merely from simpler, unambiguous syntactic structure. 
Once converted to a tree based form, code and NL are on equal footing, with all ambiguity vanquished; the syntactic structure is fully articulated.
On this equal footing, then, is code \emph{still} more repetitive and predictable than English? 
This leads us to our next research question:

\RQ{trees}{When parse trees are explicitly included for English and Java, to what degree are the differences in predictability accounted for? }

\subsection{Explanations From Contingent Factors}
\label{sec:semantic_theory}

One hypothesis for why  code is more repetitive than NL is that 
\emph{humans find reading and writing code  harder to read and write than NL}. Code has precise denotational and operational semantics, and computers cannot deal
automatically with casual errors like human listeners. 
 As a result, natural software is not merely constrained by simpler grammars; programmers 
may further deliberately limit their coding choices to manage the added challenge of
dealing with the semantics of code. 

Cognitive science research suggests that developers process software in similar ways to natural language, but do so with less fluency.  
Prior work suggests \citep{siegmund2014understanding} that some of the parts of the brain used in natural language comprehension are shared when understanding source code.
Though there is overlap in brain regions used, eye tracking studies have been used to show that
humans read source code differently from natural language \citep{busjahn2015eye,jbara2017programmers}.
Natural language tends to be read in a linear fashion.
For English, this would be left-to-right, top-to-bottom.
Source code, however, is read non-linearly.
People's eyes jump around the code while reading, following function invocations to their definitions,
checking on variable declarations, etc.
Busjahn et al. \citep{busjahn2015eye} found this behavior in both novices and experts;
but also found that experts seem to 
improve in this reading style over time.
In order to reduce this reading effort, developers might choose to
write code in a simple, repetitive, idiomatic style, much 
more so than in natural language.

This hypothesis concerns the motivations of programmers, and is difficult to test directly. 
We therefore seek corpus-based evidence in different kinds of natural language. 
Specifically, we would like to examine corpora that are more difficult for their writers to produce and readers to understand
than general natural language.  Alternatively, we also would like corpora where, like code, the cost of miscommunication is higher. Would such corpora evidence a more repetitive style? 
To this end, we consider a few specialized types of English corpora: \emph{1) corpora produced by non-fluent language learners} and \emph{2) corpora written in a technical style or imperative style}.

\subsubsection{Native vs Language Learners}
\label{sec:efl_theory}
Attaining fluency in a second language is difficult. 
If humans  manage greater language difficulty by deploying
more repetitive and templated phrasing, then
we might find evidence for this in English as a Foreign language \emph{(EFL)} corpora.

Use of templated and repetitive language appears in linguistic research through the concept of \textit{formulaic sequences} \citep{schmitt2004formulaic}.
These are word sequences that appear to be stored and pulled from memory as a complete unit, rather than being constructed from the grammar.
Such sequences come in many forms, one of the most common being concept of \emph{idioms}, but the key point is that they
are intended to convey information in a quick and easy manner \citep{schmitt2004formulaic}.
This theory is backed by empirical evidence, as both native and non-native readers have been found to read such phrases faster than non-formulaic language constructs \citep{Conklin2008}.
Several studies have found that language learners acquire and use these sequences as a short hand to express themselves
more easily, and thus use them more excessively than native speakers \citep{schmitt2004formulaic, de2000repetitive,magali2012LearnerFormulaic}.
We can see such use as an adaption for novices increased difficulty with the language.
If we can statistically capture the patterns in written corpora of language learners and see similar trends as in source code,
it would be consistent with the hypothesis that source code is more repetitive because it is more cognitively difficult.
Therefore we ask the following questions:

\RQ{ESL}{Do english foreign language learners produce writing that resembles code patterns more closely than general English?}

\subsubsection{Technical and Imperative Style}
\label{sec:tech_theory}

Tied into alternative cognitive explanations for the observed differences between programming and natural languages
is the question of style.
Source code is a technical production; if writing in a technical style is more difficult, we would expect other technical corpora
to be more repetitive and predictable.

Differences between general language usage and specialized/technical language usage,
 have long been a focus of linguists \citep{gotti2011investigating}.
Prior attempts to categorize the features of specialized language \citep{sager1980english},
are often characterized by somewhat contradictory forces.
Gotti cites Hoffman who gives 11 properties desirable in technical language, including unambiguousness, objectivity, brevity, simplicity, consistency, density of information, etc. 
The desire for a lack of ambiguity contradicts with the desire for concise and information dense text, as the meaning is also intended to be clear \citep{hoffmann1984seven,gotti2011investigating}.
Moreover, technical language is also heavily decided by the intended audience, ranging a spectrum from communication to laypeople 
(either for educational or general public dissemination) to communication between experts, which often includes highly unambiguous 
mathematical formulations \citep{gotti2011investigating}.
Expert to expert communication
is characterized by usage of unexplained terminology, or jargon, which
can be efficient \citep{varantola1986special,gotti2011investigating}.
Moreover, technical language is marked by compound noun phrases that may be easier for language models to detect.
Salager et al. found that compared to the 0.87\% rate of compounds in general English, technical language had them appear at a rate of 15.37\% \citep{salager1983compound}.
Additionally, longer sentences are associated with technical language, especially legal language, with
increased length sometimes suggested as arising from a need for greater precision \citep{gotti2011investigating}.
However, this claim of precision in legal language is disputed, as Danet points out that for being supposedly 
precise, laws often require extensive interpretation \citep{danet1980language}.
Though there is evidence of political gamesmanship making the language overly verbose and complex,
legal language and technical language in general are still driven in part by the need for
precision and reduced ambiguity.
Such language can be seen as more difficult or labored than general language, 
and we would expect it to feature more code-like properties.

If we consider language transactions as an optimization of cognitive effort between 
speaker and listener \citep{zipf49, piantadosi2012communicative}, then it is useful to consider how the type of language will shift
the balance in one direction or the other. 
In fact, psycholinguistic research suggests that a reader's or listener's cognitive load
increases when faced with certain types of ambiguity and increased entropy in language \citep{hale2003}.
 In language where there is a high cost when the listener misinterprets the speaker, then
these theories would predict the language would become less ambiguous, which would be reflected in language models.
In code, there is a very high cost of misinterpretation, and thus the grammar does not typically permit ambiguity (barring undefined
behavior in languages like C).  Thus, in theory, contexts in natural language with a high cost will also more closely resemble code. 
Technical language is one such area where clear communication is important, but imperative language is another.
When humans write instructions or give commands, if the reader or listener misinterprets the commands, there is presumably a higher
cost than in the case of merely descriptive language.  Therefore, we would also expect such corpora to exhibit more code-like behavior.

\RQ{tech}{Do technical and imperative language, seemingly more difficult and with higher cost of misinterpretation than general and domain specific language, exhibit more code-like properties?}

\subsection{Statistical Language Models and Entropy}
\label{sec:langmodel}

A \textit{Statistical Language Model} assigns a probability to utterances in a language.
These models are estimated on a representative training corpus, 
and typically work by 
by estimating the probabilities of a token in a given context.
Let us define an utterance  as a sequence of tokens $S = t_1, t_2, ...,t_n$.
For each token $t_i$ in the sequence, we have a corresponding \textit{context} $C(t_i)$.
The exact definition of the context will depend on what language model is being used.
In \textit{ngram} models, the context is defined as the preceding 
$n$ tokens; in neural models such as an forward LSTM, all previous tokens are available as potential context\footnote{Bidirectional LSTMs can make use of context both before and after a token.}.
Then, we can define the probability of the sequence relative a language model $L_M$ as:

\begin{gather}
\label{Eq:LmGen}
P(S;L_M) = \prod_{i=1}^{n}P(t_i|C(t_i); L_M)
\end{gather}

\noindent
Eq. \ref{Eq:LmGen} defines the probability of the sequence as the product of the probabilities of each token in the
sequence, given the context of the token and the language model.
Typically, instead of using the raw probabilities, Eq. \ref{Eq:LmGen} is represented in the form of \textit{entropy}.
Formally, the average entropy per token in $S$, $\bar{H}$ is defined as:
\begin{gather}
\label{Eq:ent}
\bar{H}(S;L_M) = -\frac{1}{\norm{S}}*\sum_{i=1}^{n}log(P(t_i|C(t_i); L_M))
\end{gather}

Originally proposed by Shannon \citep{shannon1948mathematical}, 
who later used it to predict the next letter in a sequence of English \citep{shannon1951prediction}, 
entropy models the amount of information conveyed by a message.
That is, if the message where to be translated to binary, what is the fewest number of bits required to encode it in the language model?
The fewer the bits are needed  encode the message, the less information (and thus more repetitive/predictable) the message. 
In the context of language models, entropy indicates how \textit{unexpected} a token is, and acts as measure of how successful
the language model is in capturing the underlying relevant features that characterize the grammar,
vocabulary usage, and ideas of the text.  

Different types of models capture different kinds of repetitiveness, so
considering the entropy of a text under multiple language models gives greater insight into the features of a text. 
We thus explore predictability and repetition using basic ngram models, ngram cache models 
that focus on capturing local repetition, and LSTM models capable of capturing long term dependencies in the text.

N-gram models are the simplest: here, the context $C(t_i)$ is equivalent to the past $n$ tokens in 
the sequence.  For example, the probability of a sentence in a trigram model would be:

\begin{gather}
\label{eq:ngramModel}
P(S) = \prod_{i=3}^{n}P(t_i|t_{i-2},t_{i-1})
\end{gather}

\noindent
Note that we can pad the start of a sequence with buffer tokens in order to produce a probability value
for the initial tokens.  Thus, in the above example $t_3$ would be the actual first token in the sequence.

Ngram models capture the 
global repetitiveness of a corpus, but  source code has additional \emph{local} 
repetitiveness. 
These local patterns  are modeled in a local \textit{cache}, 
and this type of model as an \textit{ngram cache model}.
Tu et al. originally observed this effect in Java code \citep{Tu2014}, and Hellendoorn et al. have recently extended the idea of a cache to have multiple layers
of nesting \citep{hellendoorn2017deep}.
It is notable the ngram cache models don't show improvement over ngram models in English. 
Formally, Eq. \ref{eq:cacheModel} shows the basic cache model as described by Tu et al.

\begin{gather}
\label{eq:cacheModel}
\begin{split}
P(t_i|h, cache) = \lambda * P_{ngram}(t_i|h) + (1 - \lambda) * P_{cache}(t_i|h) \\
0<= \lambda <= 1
\end{split}
\end{gather}

The cache model interpolates between two ngram models $P_{ngram}$ and $P_{cache}$.
The first is the regular ngram model as described in \ref{eq:ngramModel}.
The second ngram model is built using counts built from the local \textit{cache}.
Details on this model and how $\lambda$ is selected can be found in Zhaopeng et al \citep{Tu2014}.

Finally, we also use  Long Short Term Memory Network, or LSTM \citep{hochreiter1997long}.
Unlike traditional feedforward models, these recursive neural networks 
(RNNs) allow models to leverage variable-length contexts \citep{mikolov2010recurrent}.
LSTMs extend RNNs by adding the ability to choose to remember some of the prior elements of the sequence\footnote{A good explanation of the details of LSTM cell structure can be found at: \url{http://colah.github.io/posts/2015-08-Understanding-LSTMs/}}.
This ``selective memory" allows LSTMs to learn longer contexts than the fixed ngram models.

LSTMs and RNNs have been applied to both natural \citep{mikolov2010recurrent, sundermeyer2012lstm} and programming languages \citep{white2015toward, khanh2016deep}.
We include LSTMs to compare and contrast their ability to learn natural and programming languages, but also to leverage their greater context when modeling our
linearized parse trees.
Much larger ngram models are needed to capture the text of these trees, but the selective learning of the LSTM is greater able to capture the repetition in them.
We provide more details on these in sections \ref{sec:EqPartTree} and \ref{sec:Lm}.

\subsection{Zipfian Distributions in Natural Language and Code}
\label{sec:zipf_theory}

Zipf famously observed that the distribution of the vocabulary of natural language is made up of a few highly frequent words with a long tail of very rare words \citep{zipf49}.
The original formula indicates a power-law relationship between 
the rank of a word and its frequency.
By rank, we mean that the most frequent word receives rank 1 (or 0), then the next most frequent gets rank 2, and so on.
Then, the frequencies of this words following roughly this formula:

\begin{gather}
f \approx \frac{C}{r^{\alpha}}
\end{gather}

Here, $f$ represents the frequency of the word, $C$ is a constant, $r$ is the word rank, and $\alpha$ is the power used
to fit the line (originally observed as being close to 1) to the data.  This law was improved slightly to better fit very frequent 
and very rare words by Mandelbrot soon after \citep{mandelbrot1953informational}.
He proposed an additional constant $b$, which was better able to account for high frequency words in natural language
texts:

\begin{gather}
f \approx \frac{C}{(r+b)^{\alpha}}
\end{gather}

However, the power law can only approximate the frequency patterns of language.
More precise models of word frequency
include a bipartite function known
as the Double Pareto; this plot of the distribution
 has an observable bend in log-log plots of natural language data \citep{ferrer2001two,gerlach2013stochastic,piantadosi2014zipf, mitzenmacher2004brief}, between
two slopes. 
The first slope is associated with the most frequent words, called a kernel lexicon, and a second rate of decrease among the less common words, belonging to an unlimited lexicon.
When new vocabulary is added to a natural language, they are added to the unlimited lexicon at a decreasing rate over time \citep{petersen2012languages}.
By accounting for these two vocabularies in modeling, very accurate simulations of natural language vocabulary frequency and growth
can be captured \citep{gerlach2013stochastic}.
Notably, the decreasing need for additional words observed in natural language \citep{petersen2012languages}, is not true in
source code, as developers make up new identifiers for new files, which is why cache models are so much more effect in code \citep{Hindle2012, Tu2014}.

Power laws and other related distributions (exponential, lognormal, etc) have been examined in regards to many source code features
of interest: class methods and fields, dependency and function call graphs, etc \citep{louridas2008power,concas2007power,baxter2006understanding}.
Of closest interest to our work are two papers \citep{zhang2008exploring, pierret2009empirical} source code lexical tokens against Zipf law's in the
same manner as natural language.
Both find that source code unigrams do largely follow Zipfian patterns, both in Java \citep{zhang2008exploring} and in several additional languages \citep{pierret2009empirical}.
Zhang explores dividing Java tokens into five categories and remarks on the similarity of java keywords to the natural language concept of stop words. 
However, neither paper explores the the Zipf curves of programming languages directly with natural language for comparison purposes.
We will use Zipf curves in addition to language models so that some language features can be confirmed in a environment agnostic to the choices of a particular language model.

%% file: method.tex
\subsection{Data}
\label{sec:Data}

We collected many different kinds of natural and programming language corpora.  
Below, we shall describe how each were collected, along with any modifications made
to them for our experiments.

\subsubsection{Programming Language Corpora}
\label{sec:PLCorpora}

\begin{table}
\caption{Summary of the size and vocabulary of the programming language corpora}
\center
  \begin{tabular}{| c | c | c | c |  }
    \hline
    Language & \# of Tokens & \# of Unique Tokens & Projects \\ \hline
    Java & 16797357 & 283255 & 12 \\ \hline
    Haskell & 19113708 & 473065 &  100 \\ \hline
    Ruby & 17187917 & 862575 &  15 \\ \hline
    Clojure & 12553943 & 563610 &  561  \\ \hline 
     C & 14172588 & 306901 & 10 \\ \hline
 \end{tabular}
 \label{tb:PLdata}
\end{table}

We focus most on Java and English; however, we empirically confirm that
the Java/English difference also applies to several programming languages.
We selected corpora of both functional and non-functional languages.
We gather source from OSS projects written in Java, Haskell, Ruby, Clojure, and C.
We chose Java and Ruby due to their popularity on GitHub and Java in particular due to its past use as a research subject for ngram models \citep{Hindle2012,Allamanis2013}.
We also add C as well due to its historical significance as a procedural language. 
Haskell and Clojure  are among the most popular functional languages on Github.
Two requirements were used when selecting projects for our corpora:
(1) the combined size of the projects chosen for each language were roughly equivalent and
(2) the projects did not overlap too much in shared domain or source code.

Due to differences between the more and less popular languages, 
we cannot adopt exactly the same selection criteria for each language.
On Github, developers mark projects they want to follow with stars.\footnote{\url{https://help.github.com/articles/about-stars/}}
These stars are a proxy for popularity \citep{tsay2014influence}, which we use to choose projects
in very popular languages like Java, Ruby, and C.
For these languages, we manually selected the projects by examining the list of most starred projects and carefully reading the project descriptions.
We chose projects such that they were both popular, and that their descriptions indicated that the project purpose did not overlap in domain.\footnote{One exception for Java is the Eclipse project, which was not hosted on GitHub, but is selected for significance within the Java community}

The functional languages, Haskell and Clojure, are not as popular.
After the few most popular projects, the code size of each new project drops drastically.
As having significantly smaller training data can negatively affect model performance, we decided that having corpora be roughly equivalent in size was
more important than domain diversity.
Many more projects are needed to provided sufficient data.
This makes manually selecting diverse projects unfeasible, especially as the smaller projects often lack meaningful descriptions.

We thus use an automated process that focuses first on collecting a sufficient amount of data, but still apply some constraints to filter out less meaningful
projects and avoid projects that share code.
First, we use GHTorrent \citep{ghtorrent:paper} to obtain a list of all non forked projects in the language on Github, and select those with over 100 commits.
Any project whose name directly contains the name of another project on the list is removed.
We then parsed the git logs to verify the GHTorrent results and remove any projects under the commit threshold or with only 1 contributor.

Finally, as we wish to avoid projects including that share significant amounts of exactly copied code, we remove projects that share overly similar directory structures.
For each project, we build a set of names, where the each name is a source code file and the directory immediately above it.
Then, we use the Jaccard index to compare these sets of names.
This index takes the intersection of the two sets and divides it by their union.
Any pair of projects that share more than 10\% of the their names are thus excluded.
In deciding which of the two projects to keep, we remove one if it is an obvious fork of the the other, or if it conflicts with several projects.
Otherwise we pick whichever project is larger in bytes, or if they are the same, delete one arbitrarily.

Then, for the projects selected for each programming language, we selected all files associated with the primary file type for
that language.  We took \emph{.java}, \emph{.clj}, \emph{.hs}, \emph{.rb}, and \emph{.c}/\emph{.h} files for Java, Clojure, Haskell, Ruby, and C respectively.
We use the Pygments syntax highlighting library\footnote{\url{http://pygments.org/}} in python to divide the code in tokens, and separate them with spaces,
ignoring comments and removing indentation and other whitespace.
Additionally, we treat the content of strings as three units, giving a token to the opening and closing quotes, but removing all spacing within the string and
count it as one individual token.
For example of what one tokenized line looks like, the line \emph{return EpollSocketTestPermutation.INSTANCE.socket();} is represented as
\emph{return   EpollSocketTestPermutation . INSTANCE . socket ( ) ;}.

Table \ref{tb:PLdata} shows the size in tokens and projects of the resulting corpora.
We see that all the language sizes fall in roughly the same order of magnitude, though the number
of projects needed to achieve the size varies.

\subsubsection{English Corpora}
\label{sec:EngCorpora}

\begin{table*}
\centering
\caption{Summary of the size and vocabulary of the English and other natural language corpora}
\begin{tabular}{ | l | c | c | c | }
\hline
Category & Corpus & \# of Tokens & \# Unique Tokens\\ \hline
\multirow{2}{*}{General English} & Brown & 1161192 & 56057 \\
 & 1-Billion Sample & 16852300 & 241138\\ \hline
\multirow{4}{*}{Specialized English Texts} & NASA & 245788 & 10880 \\
 & US Code & 2237992 & 29535 \\ 
 & Scifi & 1541361 & 40715 \\
 & Shakespeare & 969583 & 33273\\
 & Recipes & 1388875 & 14328 \\ \hline
\multirow{2}{*}{English as a Foreign Language} & Gachon & 3052797 & 40180 \\
 & Teccl &  2096018 & 35842 \\ \hline
\multirow{2}{*}{Other Natural Languages} & German & 17007990 & 710301 \\
 & Spanish &  16955041 & 453133 \\ \hline
\end{tabular}
  \label{tab:EngCorpora}
\end{table*}

We drew on a variety of natural language corpora to capture general characteristics of writing,
those specific to writing produced by English language learners, and the differences in English
technical and non-technical language.
We will describe each corpus in turn below, but
summaries of all English (and other natural language) corpora are located in Table \ref{tab:EngCorpora}.

First, for general purpose English, we began with the topically balanced Brown Corpus \citep{kuvcera1967computational}, provided by the NLTK project \citep{bird2006nltk}.
While well balanced, this corpus is small for modern statistical language modeling, so we also used as a general English corpus a 1 billion token benchmark corpus \citep{ChelbaMSGBK13}.
As noted previously, it is important that the language models are explored to roughly equivalent sized training sets, and 1 billion tokens is orders of magnitude larger
than our code corpora.
Thus, we select a random sample of this corpus, ending up with approximately 17 million tokens -- about the same size as the programming language corpora.

Although English is our primary example of natural language, we also consider two other natural language corpora, German and Spanish, to verify that our results are not specific to English and rather apply to other natural languages.
These are only used to in the initial experiment, aimed at seeing how well the comparison of the differences in repetitiveness of Java and English holds 
across various programming and natural languages.
The German and Spanish corpora were selected from a sample of files from the unlabeled datasets
from the ConLL 2017 Shared Task \citep{conll17}, which consist of web text obtained from CommonCrawl.\footnote{\url{http://commoncrawl.org}}
Like the 1 billion token English corpus, we selected a random subsample to make these corpora size comparable with our other corpora. In this sample, we excluded files from the Wikipedia translations, as we observed Wikipedia formatting mixed in with some of the files.
Summaries of the vocabulary token counts of these corpora are also in Table \ref{tab:EngCorpora}.

To test hypotheses about language difficulty and repetition, we used two english language learner corpora.
The Gachon \citep{carlstrom2013gachon} corpus is a collection of primarily Korean, but also some Chinese and Japanese
English language learners.
The Gachon corpus covers a range of just over 25K 100 to 150 words answers to 20 essay questions.
It has meta information including the years a student has studied the language, the their native language, and their TOEIC language proficiency score\footnote{\url{https://www.ets.org/toeic}}.
While this corpus contains explicit information about the writer's language proficiency, it does suffer from a confounding effect of being limited in domain to merely 20 topics.
Domain specificity is known to make corpora more predictable and repetitive \citep{Hindle2012}.
Therefore, we include the Teccl Corpus (Ten-thousand English Compositions of Chinese Learners) \citep{teccl2015} as another example of \emph{EFL} for robustness.
Unlike the Gachon corpus, Teccl covers a much wider range of topics (the authors estimate around 1000).
It consists of a wide range of writers in both geographically and in current education level.

The question of technical and imperative language is also confounded with the possibility of restricted domain.
Therefore we selected five corpora, two technical corpora, two non-technical corpora with potentially restricted domain, and a corpus of instructions in the form
of cooking recipes.
The two non-technical corpora came from literature: a corpus
of Shakespeare's works \citep{norvig2009natural}, restricted in domain by having the same author, and a corpus complied of 20
classic science fiction novels from the Gutenberg corpus\footnote{\url{https://www.gutenberg.org/}}, which all fall under the same literary genre.  

For the technical and imperative language corpora, we selected a corpus of NASA directives, a corpus of legal language, and a corpus of cooking recipes.
The NASA directives were scraped from the NASA website.
Directives share similarities with source code requirement documents, a written English equivalent to source code. 
Source code requirements explain in detail what is expected from a software application, and the requirements documents of the NASA CM1
and Modis projects have been used in many requirements studies \citep{hayes2005improving, sundaram2005baselines}.
However, the requirements documents for the two NASA projects often used in these studies are only about 1.2K words for Modis, and 22K words
for CM1.
Language models typically require far more words, we mined the more general NASA directives, creating a corpus approximately 245K words
long.

Legal language shares qualities with code in that it must also be prescriptive and precise.
Like code variables and functions regularly reference other parts of the code, so to do references within legal text.
For this purpose, we downloaded the US Legal Code \footnote{\url{http://uscode.house.gov/download/download.shtml}}.
The US legal code consists of 54 major title sections relating to the general permanent federal law of the United States. 

Finally, we use a recipe corpus as a study of imperative language meant to be expressed clearly and with specific purpose.
This corpus comes from the text found in the million recipe corpus \citep{salvador2017learning}.
Like source code, recipes are series of instructions, though the degree of precision required in the writing is lower.
In order to make this corpus comparable in size to our technical corpora, we selected a random sample of the recipes such that the
total text would be about 1 million tokens in size.
The full corpus contains images, ingredients, and instructions associated with each recipe.
For our purposes, we only considered the \emph{instructions} text for each recipe as input into our models.

\subsubsection{Parse Tree Corpora}
\label{sec:ParseTreeCorpora}
  
For our parse tree comparison experiment,  we needed to extract an abstract
syntax tree for a software corpus, and represent it in a similar fashion to natural language constituency trees (as described below). 
This experiment was limited to our Java and English data. 
When comparing the parse trees, we first selected constituency parse trees for written English from
the Penn Treebank \citep{Marcus:1993}, which includes sections from the Brown Corpus and the Wall Street journal corpora.
Then, we used a modified version of the Eclipse Abstract Syntax Tree parser to transform all the files in our Java corpus to English.
Since the Java trees could be automatically created, we randomly sampled from these Java files in order to make the corpora roughly size 
equivalent in token count to the Penn Treebank trees.
Details on the modifications made to make the two trees more comparable are described in section \ref{sec:EqPartTree}.
 
Table \ref{tab:TreeCorpora} shows the sizes of the resulting corpora.
We see that the trees have roughly the same number of non-terminal tokens, but that the number of distinct rules
is much larger in English than in Java.
Likewise, the Java trees have about half as many terminal tokens.

  \begin{table}
\centering
\caption{Summary of corpora token counts and vocabulary for the modified English and Java parse trees}
  \begin{tabular}{ | c | c | c |}
    \hline
     & Java Trees & English Trees \\ \hline
    All Tokens & 11267469 & 11354764 \\ \hline
    Terminal Tokens & 2191014 & 1740902 \\ \hline
    Simplified Non-Terminal Vocabulary Size & 81 & 93 \\ \hline
  \end{tabular}
  \label{tab:TreeCorpora} 
\end{table}

\subsection{Comparing Language Repetition and Predictability}
\label{sec:MeasureRepetition}

We present two general methods for measuring the repetition of language corpora.
The first involves the reporting the entropy values of each token given a language model as described theoretically in section \ref{sec:langmodel}.
The details of the modeling and the representation of the results can be found in section \ref{sec:Lm}.

However, the redundancy of corpora can also be modeled using a adaptation of the Zipf plot \citep{zipf49}.
In a standard Zipf plot, we count all occurrences of a word in a text and assign each word a rank based on frequency.  
The x-axis is the rank of the word, and the y-axis is its frequency.
When plotted in log-scale, this relationship appears roughly linear.

We modify this plot in two ways.  
First, we normalize the frequencies on the y-axis to a percentage to make different corpora more comparable.  
Second, we extend the idea of a Zipf plot beyond merely individual words.
Instead of just looking at the frequencies of individual words, we can also look at the frequencies of sequences of words.  
If we count bigrams, trigrams, or higher order ngrams, the distribution of phrase usage of a text becomes
apparent.
In more repetitive texts the most repetitive phrases take up proportionately more of the text.
On a log-log plot, we can visually approximate this effect by examining how steeply the roughly power law slope of the data descends.
Less creative texts begin higher on the y-axis and descend to their less frequent phrases more quickly.
Note that we are not interested in fitting a precise distribution to this data, such as double pareto, power law, or otherwise.
It is the relative slopes of the data that is important.
Once normalized, corpora with steeper slopes are more repetitive; those with shallower slopes are less repetitive.

This alternative way of measuring corpus repetition is useful as it averts some of threats from using language models.
To use an LSTM the vocabulary size must be limited to remove very infrequent words, making the results for infrequent
tokens or sequences somewhat artificial.  There is no such limitation in the Zipf plots, which increases the robustness
of the overall observations.

\subsection{Measuring Open Category Words}
\label{sec:MethodOpenClose}

To test the hypothesis that differences in closed category words account for most differences between source code and English,
we remove the closed vocabulary words from a corpus, and leave behind just sequences of open vocabulary words.
Removed are elements most closely
tied to the language syntax; arguably, what remains are \emph{content words}.
These most closely model the sequence of \emph{ideas} expressed by the text. 

How do we determine what tokens qualify?
For English, we use a list of 196 words and contractions, along with a list of 30 punctuation markers, derived from a published NLTK stop word list \citep{bird2006nltk}.
For our programming languages, we use the Pygments' type categorization (implemented with regular expressions) to remove non-identifier words, keeping references to types, classes (when applicable), variables, and function names.
Specifically, we labelled as open category tokens that Pygments had marked as Token.Name (but not the subtype Token.Name.Builtin), Token.Keyword.Type, Token.Literal.String, or Token.Number with a few modifications.
These modifications involved some small changes to keyword lists and are intended to make the closed category words more consistent across the different
programming languages.
For example, we classified the boolean (\emph{true/false} and \emph{null} literal values as closed category.
We also extended the list of what Pygments considered keywords in Haskell\footnote{We add \emph{\textbackslash}, \emph{proc}, \emph{forall}, \emph{mdo}, \emph{family}, \emph{data}, and \emph{type}.}, Ruby\footnote{We add \emph{\_\_ENCODING\_\_}, \emph{\_\_END\_\_}, \emph{\_\_FILE\_\_}, and \emph{\_\_LINE\_\_}.}, and Clojure\footnote{We add  \emph{recur},  \emph{set!},  \emph{moniter-enter},  \emph{moniter-exit},  \emph{throw},  \emph{try},  \emph{catch},  \emph{finally}, and \emph{/}, along with some operators Pygments had classified as Token.Names}. 
These labels only approximate the open category words, but they do remove operators, separators, punctuation, and most keywords.
If these sequences of content words are more repetitive in source code than in natural language, 
this would be consistent with the theory that the repetition in code is not wholly due to syntactic constraints.
Below is are examples of what part of these filtered sequences would look like in Java and English respectively:

\textit{... InputStream in FileInputStream file ByteArrayOutputStream out ByteArrayOutputStream byte buf byte 8192 ... }

\textit{... Now 175 staging centers volunteers coordinating get vote efforts said Obama Georgia spokeswoman Caroline Adelman ...}

One consideration for these open category words in code is the question of how to handle literal values.  In the case of strings,
an argument could be made that many of them would qualify as being natural language, leading to a dual language corpus.
We compared the code corpus open category words both with and without the literal values included, but found little difference
in the overall trends from our language models and Zipf models, though the exact size of the differences changed.
Presented in this paper are the results of the corpora with the literal
values included.

A potential threat to this experiment results from the fact that English open and closed category words are fairly well defined, 
but the concept is much less clear in the context of programming languages.
Though Pygment's provides a good approximation (which we attempt to tweak further to improve), there are some few ambiguous cases.
Some language elements are very common and difficult to extend without strictly being on the official list of reserved words, 
or could be construed as part of a larger category that \emph{is} open category, such as primitive types like \emph{int} in Java can be seen as belonging to
the larger open category of \emph{types}\footnote{In particular, we
called these primitives types open category to be consistent with how other programming languages like Haskell treat their types.}.
We argue that these edge cases are infrequent enough and the size of the effects observed in our experiments are large enough that drawing the boundaries between open and
closed differently would only slightly impact our results.

\subsection{Creating Equivalent Parse Trees in Java and English}
\label{sec:EqPartTree}

While the syntax of Java and English strings can be unambiguously represented with a tree data structure, the
trees themselves are not structurally identical; there are a few potentially confounding factors.
For one, the Java grammar is explicitly defined, where as English grammar can be at best estimated via human or automated techniques.
The Java trees are also \emph{abstract}; they do not represent every token present in the original text.
Many tokens such as the \emph{punctuation} of code (e.g. brackets, semicolons, dot operators, etc) and some reserved keywords are not explicitly listed in the tree.
In contrast, the constituency parse trees of English are \emph{concrete} -- all tokens from the raw text are preserved in the tree.
Moreover, if the vocabulary size of the set of non-terminals is radically different, then comparisons between the trees may not be fair.
A larger number of non-terminals allows for more potential choices and a higher upper bound on variation and repetition, which may be more predictive of terminal symbols.
Finally, the syntax trees in Java and English represent different granularities of objects.
In Java a complete \emph{AST} describes an entire file; in English, the tree describes a sentence.
Thus, the code \emph{ASTs} are both encompass for tokens and have longer paths from the root to the leaves.

How can we account for some of these differences and create a more fair comparison between the two trees?
First, we use a highly reliable English constituency parse -- that from the Penn Treebank \citep{Marcus:1993} (PTB).
This includes about 200 files of the 500 file Brown corpus, with an additional text from the Wall Street Journal.
All parses have been manually corrected by linguists to ensure accuracy; Indeed, 
PTB is a standard choice for training/evaluating other automated syntax parsers for English \citep{de2006generating,petrov2006learning,andor2016globally}. 
Automated methods for creating parses of English exist \citep{de2008stanford,petrov2016announcing}, but they are not always accurate.
To focus on the actual grammatical structure rather than an approximation, we choose
the human annotated parse trees as our corpus. 

Second, we modify both trees to make them more comparable.
For Java, we modify the tree to be \emph{concrete} instead of abstract.  
We created a new category, called \emph{PUNCTTERMINAL} for all terminal tokens typically missing from an \textit{AST}, 
giving a total of 81 non-terminal tags.
These new nodes are inserted into the syntax tree such that during preorder traversal,
the terminals will appear in the same order as in the original text --
 a feature that is already true of the constituency parse trees.
In the English parse trees, we consider the effect of reducing the size of set of non-terminal tags to be closer to the size of Java's
The PTB includes tags with multipart labels (for example \emph{NP-2}, \emph{PP-TMP}, 
\emph{ADVP-TMP-PRD}.
We reduce this set by taking only the first portion, such as \emph{ADVP}, which represents the 
more generic grammatical \emph{category} of an adverbial phrase, leaving out additional tags that 
reflect grammatical \emph{function} (e.g. \emph{ADVP-TMP} reflects that the adverbial phrase serves a 
temporal function).
Once this reduction has been performed, we have a total of 93 syntactic categories for English. 
To verify whether this reduction could unfairly penalize the language models ability to learn the grammar, we consider results on both the original unmodified tags and on the simplified tags that are size equivalent to Java.
In our plots, we will refer to the modified English trees as \emph{simplified}.
That the English trees capture sentences and the Java trees capture files remains an intrinsic difference between them and a possible threat, but these changes
at least make the trees contain similar organization and content.

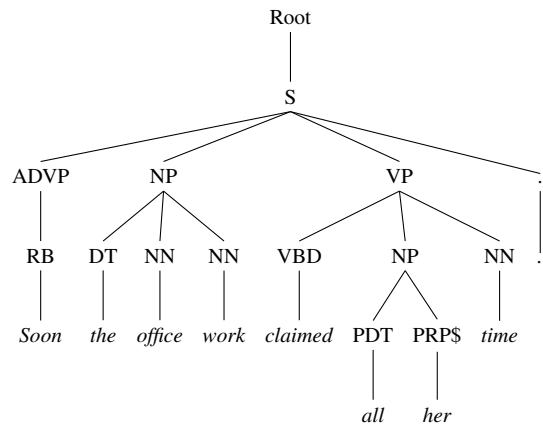
\begin{figure*}[t]
\centering
\begin{tikzpicture}
\Tree [.Root [.S [.ADVP [.RB \textit{Soon} ]]
	[.NP [.DT \textit{the} ] [.NN \textit{office} ] [.NN \textit{work} ]]
	[.VP [.VBD \textit{claimed} ] [.NP [.PDT \textit{all} ] [.PRP\$ \textit{her} ]] [.NN \textit{time} ]]
	[.{.} \textit{.} ]]]
\end{tikzpicture}
  \caption{A example CPT from one of the sentences from the Penn Treebank along with the reduced tag sets}
  \label{fig:ModTreesCPT}
\end{figure*}

\begin{figure*}[t]
\centering
\begin{tikzpicture}[scale=.95]
\Tree [.{...} [.{\#EnumDeclaration} 
	[.{\#Modifier} \textit{public} ]
	[.{\bf \#PT} \textit{\bf enum} ]
	[.{\#SimpleName} \textit{JavaDocOutputLevel} ]
	[.{\bf \#PT} \textit{\bf \{} ]
	[.{\#ECD} [.{\#SimpleName} \textit{VERBOSE} ]]
	[.{\bf \#PT} \textit{\bf ,} ]
	[.{\#ECD} [.{\#SimpleName} \textit{QUIET} ]]
	[.{\bf \#PT} \textit{\bf \}} ]]]
\end{tikzpicture}
  \caption{An example of part of a modified AST capturing a single line of Java. The bolded tags correspond to nodes added to ensure the tree contains all tokens from the original text (PT = PunctTerminal, ECD = EnumConstantDeclaration)}
  \label{fig:ModTreesJPT}
\end{figure*}
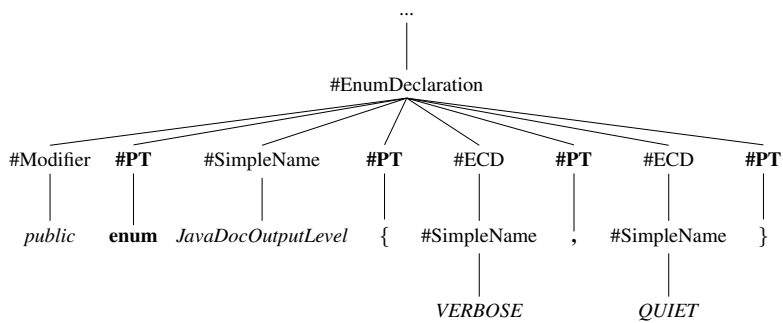

Figures \ref{fig:ModTreesCPT} and \ref{fig:ModTreesJPT} display examples of what each of these trees look like for English
and Java respectively.  Note how the changes to the Java tree ensure that both trees produce the original text in the same
left to right order.  The tags used for English are described by the Penn TreeBank \citep{Marcus:1993}, and the tags for the Java
AST come from the eclipse ASTNode class.\footnote{\url{https://help.eclipse.org/neon/index.jsp?topic=\%2Forg.eclipse.jdt.doc.isv\%2Freference\%2Fapi\%2Forg\%2Feclipse\%2Fjdt\%2Fcore\%2Fdom\%2FASTParser.html}}

To measure the entropy values of the terminal tokens, we convert the trees to a linear form.
We traverse the tree in preorder form, which presents the non-terminals as context for the terminal
symbols and also retains the order of the words originally in the text.
Then, we apply our language models to this linearized parse tree.
How do we know that our language models will be able to capture the syntax of the grammar in this form?
Some prior work with LSTMs has shown the ability to capture meaningful information from \emph{CPTs} \citep{vinyals2015grammar},
they are a reasonable choice to model 
the terminal token predictability here.
Importantly, by examining \emph{only} the entropy values of the terminal tokens, we account for the differences in the complexity of the grammars\footnote{Indeed, when running LSTMs over just the nonterminals, we see that the Java grammar is more predictable than the English grammar.}.
Given the extra information the grammar provides, we can see how the differences in terminal choice between Java and English changes.
The more the gap reduces, the more the differences in the language can be attributed to the grammar instead of some other contingent factor.
Finally, while the theoretical grounding for capturing the grammar's of the trees has only been found with neural models like our LSTM, 
we also include results from the simpler ngram and cache models for completeness.

\subsection{Modeling Details}
\label{sec:Lm}

The parameters of our ngram models were estimated using KenLM \citep{heafield2011kenlm} with modified Kneyser-Ney smoothing \citep{kneser1995improved},
based off of the code used by Tu et al. \cite{Tu2014}.
For the raw texts of all English and programming language corpora, we use a trigram model as the base.
When comparing the parse trees, we use instead a 7-gram model to capture more information about the sparser context.
This was determined empirically by modeling parse trees with ngram models from 2 to 9-grams, and observing no further improvement after the 7-gram level.
In our cache models, we use a 5000 token window cache with 10 tokens of context.
Our LSTM models are implemented in Tensorflow \citep{abadi2016tensorflow}, with a minibatch size of 20, 1 hidden layer of 300 units,
a maximum of 13 training epochs, no dropout, and a learning rate of 1.0.
Additionally, to see the effect of scaling the LSTM models to a larger one for our parse tree experiment, we also used a model with 2 hidden layers of size 650, 
a dropout rate of .5,  and a maximum of 39 training epochs.
We will prefer to these models as the \emph{small} and \emph{medium} sized LSTM models going forward.
These settings are similar to those used by Hellendoorn et al \citep{hellendoorn2017deep}.

Our corpora tend to have large vocabularies, which need to be limited
in order for the LSTM models to complete within a reasonable timeframe.
Likewise, new unseen tokens can always appear in the test set.
This is especially true in source code, where new variable names can be easily created and used in new localized contexts \citep{Tu2014}.
Therefore, ngram language models use smoothing \citep{chen1998empirical}, which reserves some probability mass for unseen words. 
We limit our vocabulary size to  the most frequent 75000 distinct tokens,  
replacing the least frequent words with with a special ``unknown" token (\textit{UNK}).

For the LSTM models, we split each code corpus at the file level with 70\% of files in the training set, and 15\% each in the validation and test sets.
We do that same for the natural language corpora if they come with files; otherwise, we divide them into small chunks which are combined into training, validation, and test sets with
the same splits.
The ngram models do not use a validation set, so we combine the validation and training sets when training them.
Otherwise, the test sets for a particular corpus are the same for each type of language model, with one exception.
We found that giving the vocabulary capped version of the Java parse tree to the KenLM model caused an error.
Therefore, the test sets of the LSTM and ngram models for the Java parse tree are not exactly comparable.
As we are primarily concerned with the LSTM results and cross language comparison, this is does not have an impact on our results.
The English parse tree did not need to be capped as its vocabulary was small enough, so these comparisons are unaffected.

When comparing the results of the language models, we report the per-token estimated
 entropy values. 
This forms a distribution of entropy values, which we compare visually with box plots and quantitatively
in a pairwise fashion.
The distributions of entropy are often long tailed, 
contra-indicating the t-test, so we instead us
the non-parametric Mann Whitney U Test (also commonly referred to as a Wilcox test), to compare the distributions.
We report the significance of the test, a 99\% confidence interval for the true difference in the median value of the distributions, and a 
effect size \emph{r}, which can be interpreted similarly to a Cohen's D value \citep{field2009discovering}.
These tests and confidence intervals where implemented in R using the \emph{coin}\citep{coinRPackage2006} package, and plots were created using \emph{ggplot}\citep{ggplotRPackage2009}.

There are several potential threats to the validity to consider in our modeling choices.
While we have used several language models and tried to use random sampling to make each corpora comparable,
we cannot say how the results might change with a much larger corpus.
For some corpora, like general Java or English, one can easily get billion token corpora.
But for more specialized corpora or less popular programming languages, the pool of what is available is much smaller, and
limits how much we can use from the larger corpora.
Otherwise, the effects observed in the models could simply result from larger amounts of training data.
We selected training, validation, and test sets randomly, but a different split could produce different results.
A more robust method would be to use 10-fold cross validation, but given the number of corpora and the training time necessary
to train the LSTM models, this was not feasible.

%% file: result.tex
Below, we discuss our results comparing various corpora of Code and English.
The structure is as follows.
\begin{enumerate}
\item
First, we examine if the Java-English difference  is consistent in other
programming languages and natural languages. 
\item
Then, we will compare the open category content words of each programming language with those in English
to see how repetitive the content of English and programming languages are.
\item
We explore the syntactic structure of Java and English to see what
parts of the structure of each contributes to differences in repetition.
\item
Finally, we compare source code with out English language learner and technical corpora to see if the expected characteristics
of each make them more code-like.
\end{enumerate}

\subsection{Repetition in Natural Languages and Various Programming Languages}
\label{sec:ResultRawText}

 \begin{figure}[ht]
\centering
\subfloat[3 grams]{\includegraphics[width=.49\textwidth]{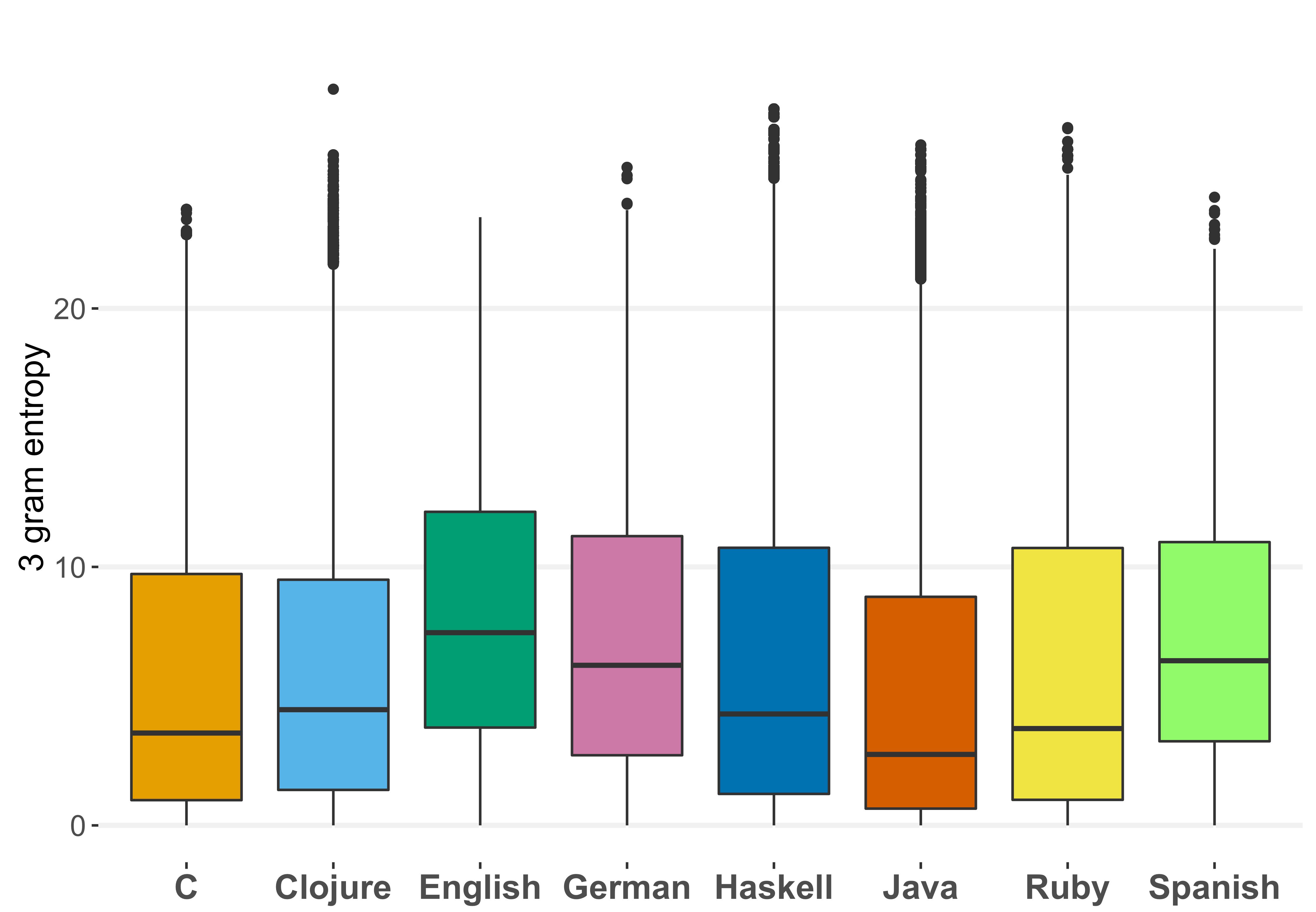}} 
\subfloat[3 grams with cache]{\includegraphics[width=.49\textwidth]{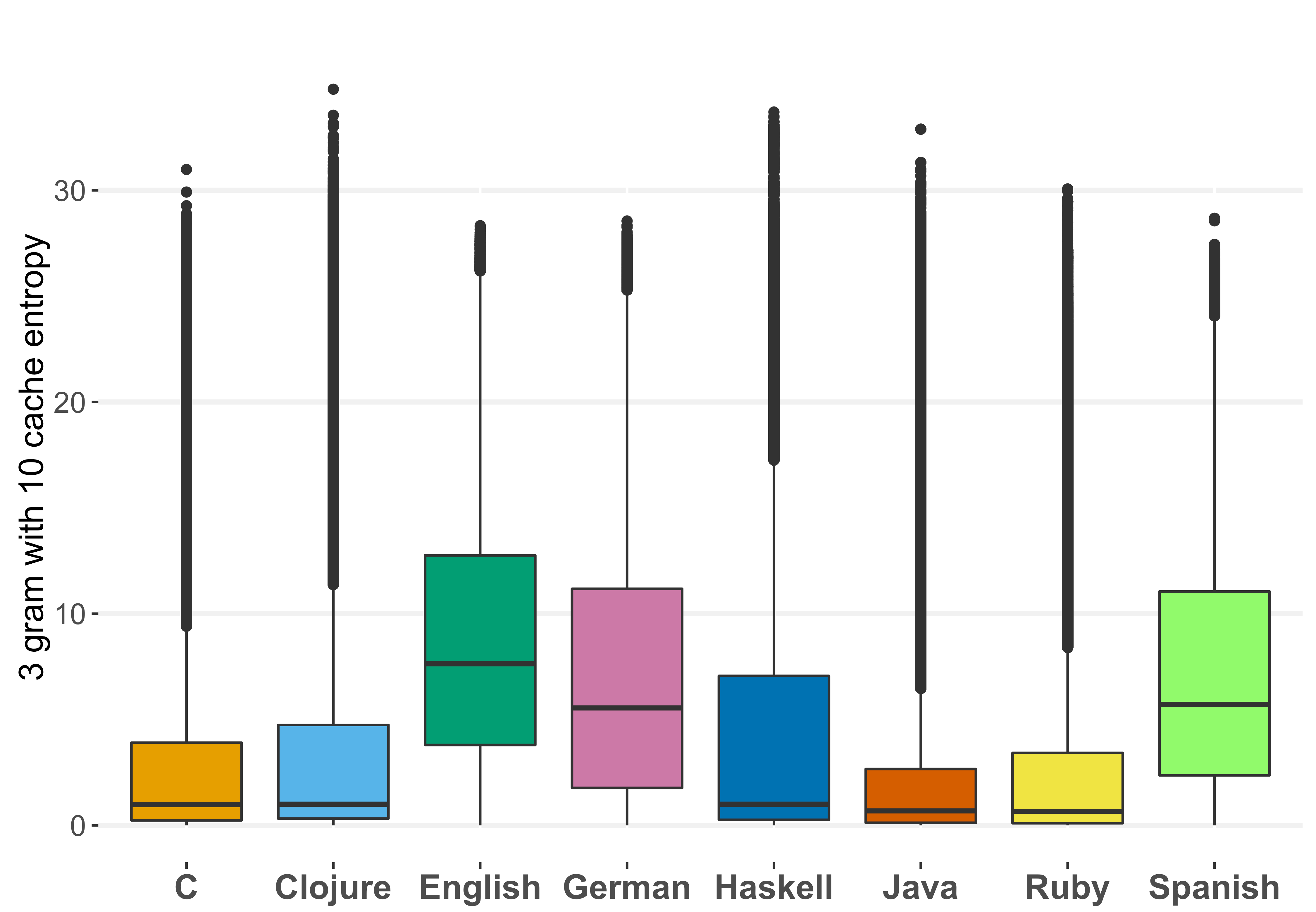}}\\
\subfloat[LSTM (Small)]{\includegraphics[width=.49\textwidth]{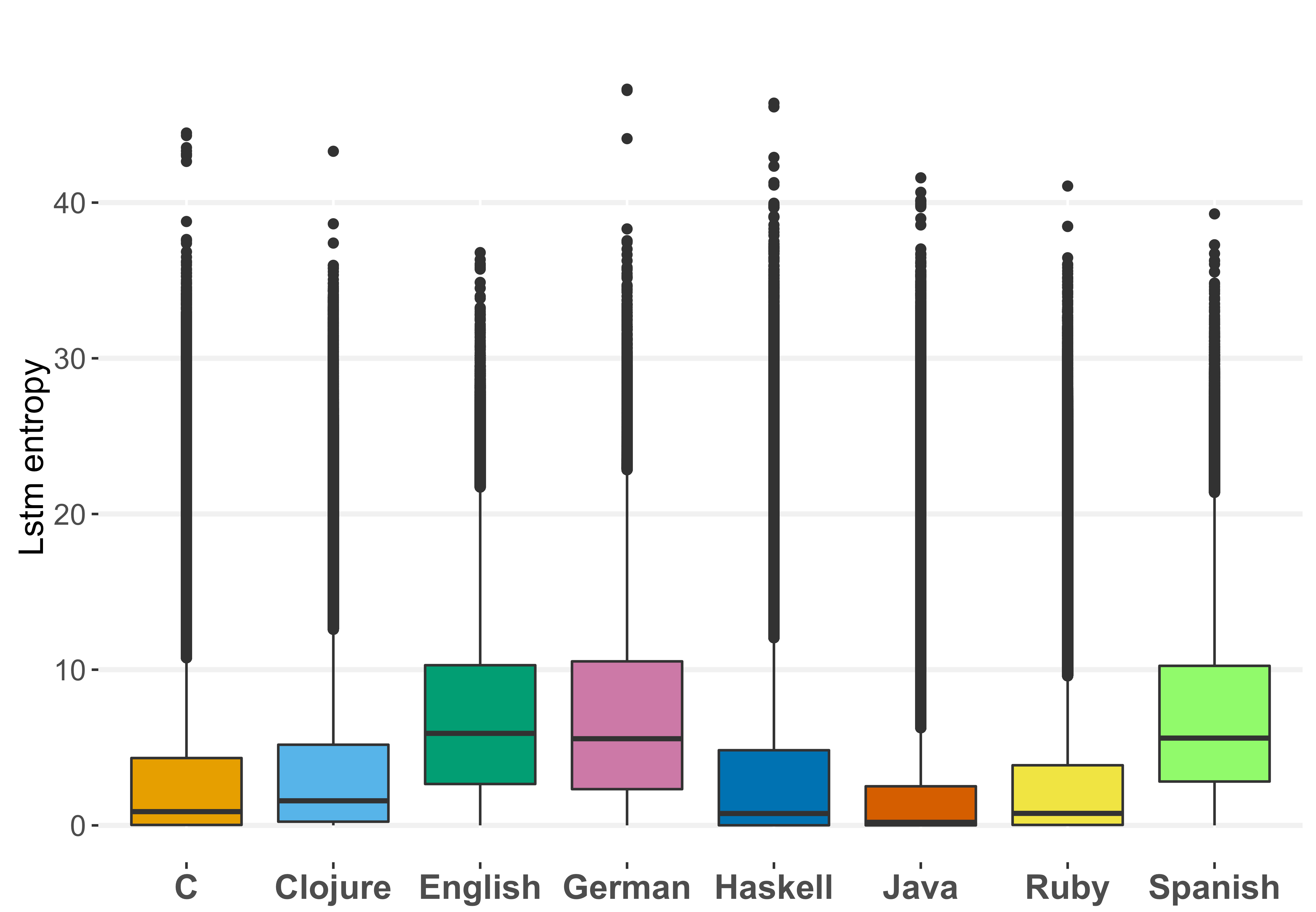}}
  \caption{Entropy score distributions for each of our programming and natural language corpora, using ngram, ngram-cache, and lstm models.  Each data point used in the box plot is the entropy score for one of the tokens in the test set}
  \label{fig:PLVsEngEnt}
\end{figure}

\begin{table*}[ht]
\centering
\caption{Summary of non-parametric effect sizes and 99\% confidence intervals (in bits) comparing each code and natural language corpus with English a baseline.  Numbers are marked with * if $p < .05$, ** if $p < .01$, *** if $p < .001$ from a Mann Whitney U test}
\begin{tabular}{| c | c  c  c |}
\hline
Language \textless  English&Ngram&Cache&LSTM\\ \hline
\multirow{2}{*}{German}  & (-0.921, -0.897) & (-1.585, -1.56) & (-0.182, -0.161) \\ 
 & $0.088^{***}$ & $0.145^{***}$ & $0.02^{***}$  \\ \hline
\multirow{2}{*}{Spanish}  & (-0.662, -0.639) & (-1.38, -1.355) & (-0.055, -0.035) \\ 
 & $0.064^{***}$ & $0.127^{***}$ & $0.005^{***}$  \\ \hline
\multirow{2}{*}{Java}  & (-2.974, -2.951) & (-5.422, -5.398) & (-4.292, -4.272) \\ 
 & $0.285^{***}$ & $0.562^{***}$ & $0.564^{***}$  \\ \hline
\multirow{2}{*}{C}  & (-2.586, -2.559) & (-4.93, -4.901) & (-3.581, -3.557) \\ 
 & $0.242^{***}$ & $0.488^{***}$ & $0.44^{***}$  \\ \hline
\multirow{2}{*}{Clojure}  & (-2.138, -2.115) & (-4.755, -4.728) & (-3.075, -3.053) \\ 
 & $0.203^{***}$ & $0.479^{***}$ & $0.372^{***}$  \\ \hline
\multirow{2}{*}{Ruby}  & (-2.338, -2.314) & (-5.12, -5.095) & (-3.691, -3.671) \\ 
 & $0.219^{***}$ & $0.516^{***}$ & $0.469^{***}$  \\ \hline
\multirow{2}{*}{Haskell}  & (-2.059, -2.036) & (-4.148, -4.139) & (-3.443, -3.423) \\ 
 & $0.191^{***}$ & $0.405^{***}$ & $0.431^{***}$  \\ \hline
\end{tabular}
\label{tab:EngCodeEffect}
\end{table*}

Fig. \ref{fig:PLVsEngEnt} displays the entropy distributions over all tokens from various language models for Java, Haskell, Ruby, Clojure, C, English, German, and Spanish. We make the following
observations. 
We replicate the prior results comparing Java to English (e.g. \citep{Hindle2012}), 
across many programming and natural languages. 
Regardless of the language model used, 
all of the programming languages are more predictable than English and the other natural language corpora..
Examining Table \ref{tab:EngCodeEffect}, which quantifies the differences seen in the box plots, we see that these differences are significant.
In fact, the programming languages are usually several bits more predictable than  English.
The other natural languages, German and Spanish, are somewhat more predictable than English with ngram models, but about the same with the longer
context of the LSTM model. 
The non-parametric effect sizes of the differences between programming languages and English vary from \emph{small} to \emph{medium}.

\begin{table*}[ht]
\centering
\caption{Summary of non-parametric effect sizes and 99\% confidence intervals of the difference (in bits) of language.  The columns compare how many bits higher the entropy of model on the left is from the one on the right.  Numbers are marked with * if $p < .05$, ** if $p < .01$, *** if $p < .001$ from a Mann Whitney U test}  
  \begin{tabular}{| c |  c  c  |}
    \hline
    Language & Ngram $>$ Cache & Cache $>$ LSTM \\ \hline
     \multirow{2}{*}{English} & (-0.218, -0.193) & (1.459, 1.484)\\ 
    & $0.020^{***}$ & $0.140^{***}$  \\ \hline
      \multirow{2}{*}{German} & (0.488, 0.511) & (-0.01,  0.003)\\ 
    & $0.051^{***}$ & $0.001^{}$  \\ \hline
      \multirow{2}{*}{Spanish} & (0.582, 0.604) & (-0.049, -0.027)\\ .
    & $0.060^{***}$ & $0.004^{***}$  \\ \hline
    \multirow{2}{*}{Java} & (1.240,1.255) & (0.148, 0.152) \\ 
    & $.275^{***}$ & $.178^{***}$   \\ \hline
   \multirow{2}{*}{Haskell} & (1.484, 1.504) & (.265, .272)  \\
   & $.220^{***}$ & $.158^{***}$  \\ \hline
      \multirow{2}{*}{Ruby} & (1.770, 1.795) & (0.004, 0.005)  \\
   & $0.327^{***}$ & $0.033^{***}$  \\ \hline
      \multirow{2}{*}{Clojure} & (2.006, 2.023) & (-0.038, -0.032)\\
   & $.318^{***}$ & $.024^{***}$  \\ \hline
      \multirow{2}{*}{C} & (1.418, 1.442) & (0.091, 0.096)  \\
   & $.271^{***}$ & $.082^{***}$ \\ \hline
  \end{tabular}
  \label{tab:modelEffect} 
\end{table*}

Tab. \ref{tab:modelEffect} also shows the improvement when a cache model
is used to capture of the \emph{locality} of the corpus. 
As expected, the basic trigram models perform the worst on all the code corpora.
The cache improves all of the programming languages.
For natural language, the cache has no effect in English, as previously reported \citep{Tu2014}. 
However, in German and Spanish, there is a small cache effect, 
much smaller than seen in any programming language.
Our small 1 layer LSTM models improve over both the ngram and cache models significantly (with the exception of Clojure, in which the cache model is very slightly better).

Interestingly, Haskell is the least predictable of the programming languages in all the models. 
Haskell is an expressive, higher-order polymorphic language with a high degree of reuse, 
and programs are often much smaller than equivalent programs in other languages; this
expressivity may contribute to lower repetitiveness. 
For instance the popular \emph{quicksort} algorithm can be implemented in only 2 lines of Haskell code while being
typically longer in other languages.\footnote{See https://rosettacode.org/wiki/Sorting\_algorithms/Quicksort\#Haskell.  Compare Haskell's implementation with others on the page.}

 \begin{figure}[ht]
\centering
\subfloat[Unigrams]{\includegraphics[width=.49\textwidth]{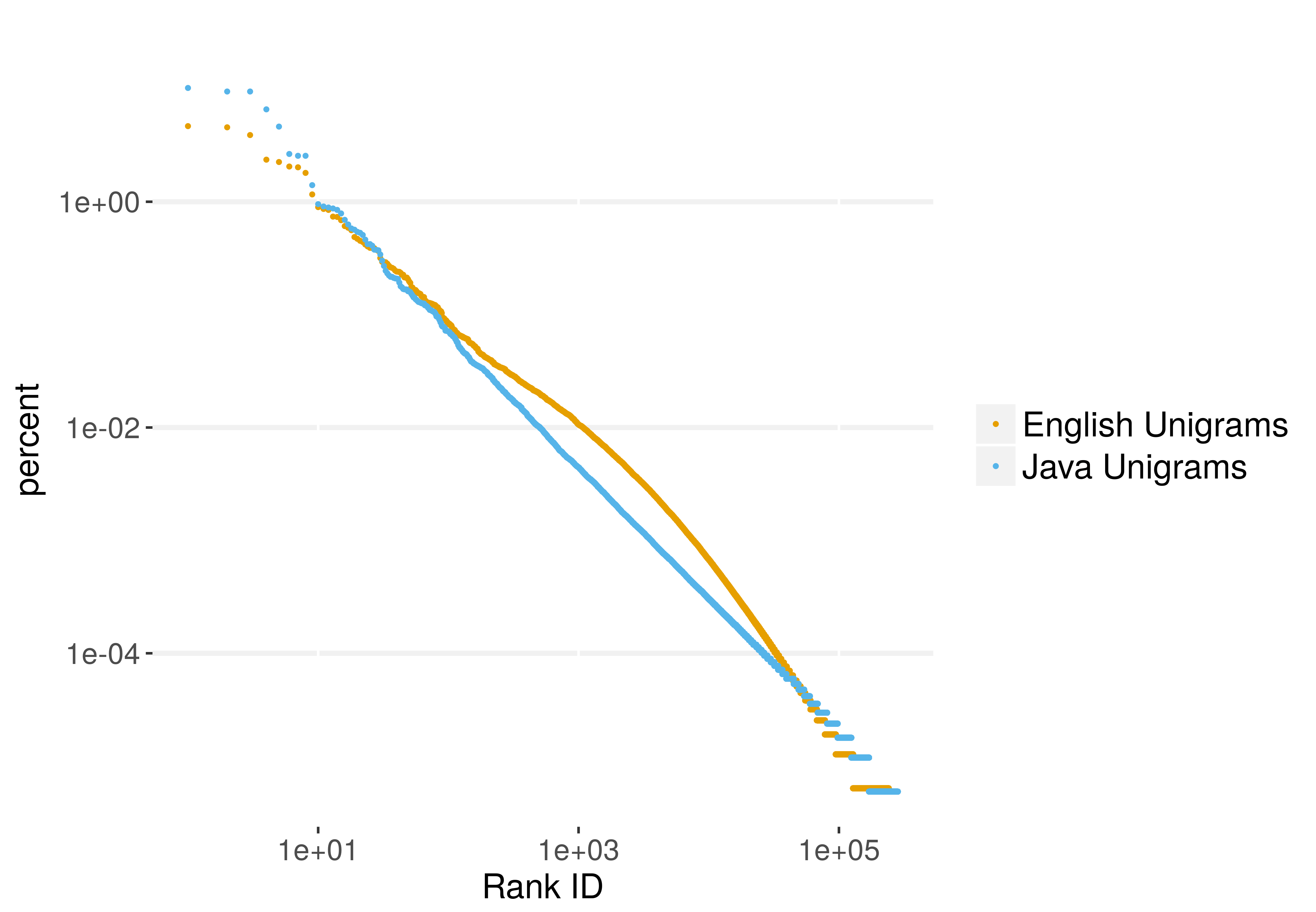}} 
\subfloat[Bigrams]{\includegraphics[width=.49\textwidth]{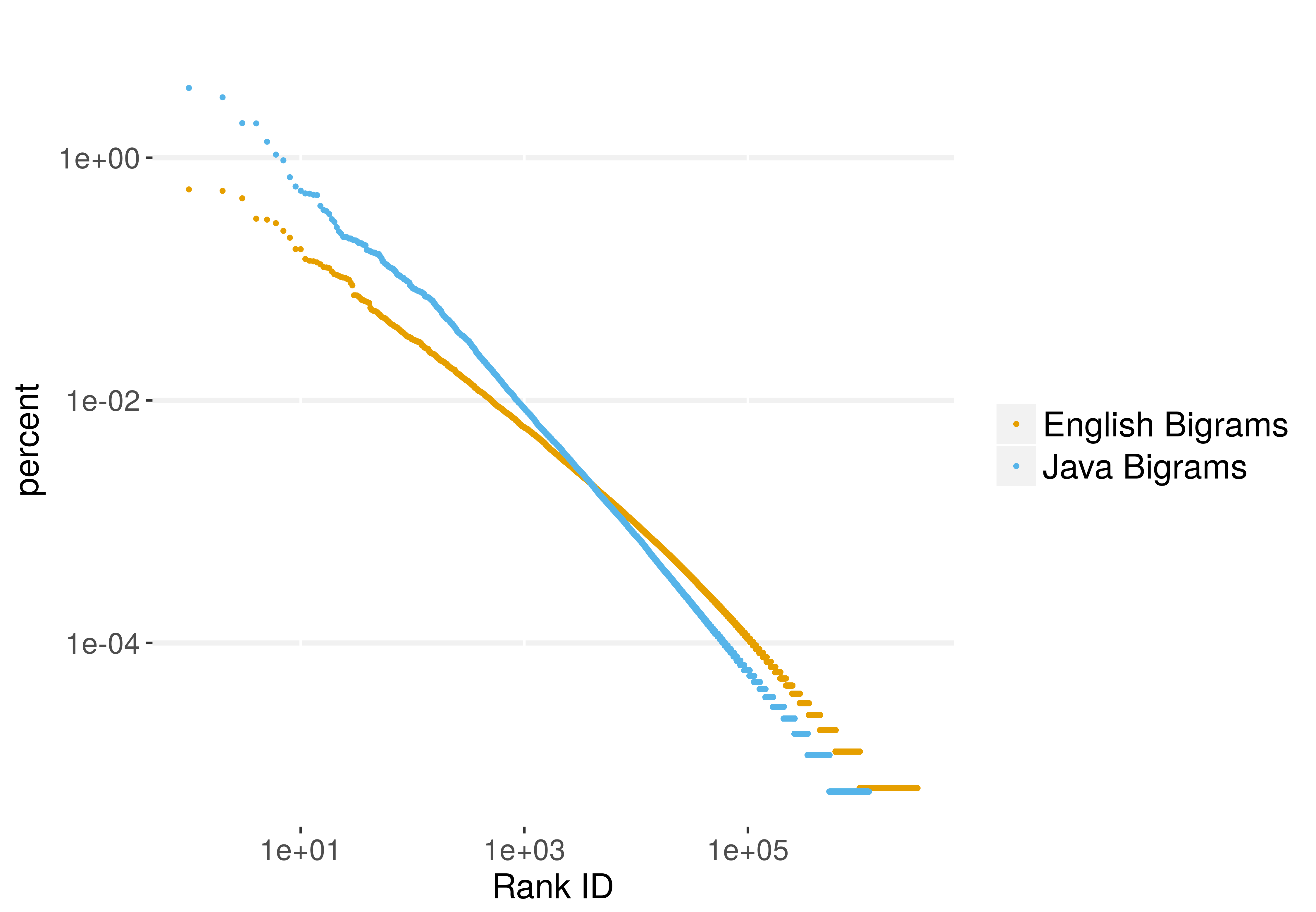}}\\
\subfloat[Trigrams]{\includegraphics[width=.49\textwidth]{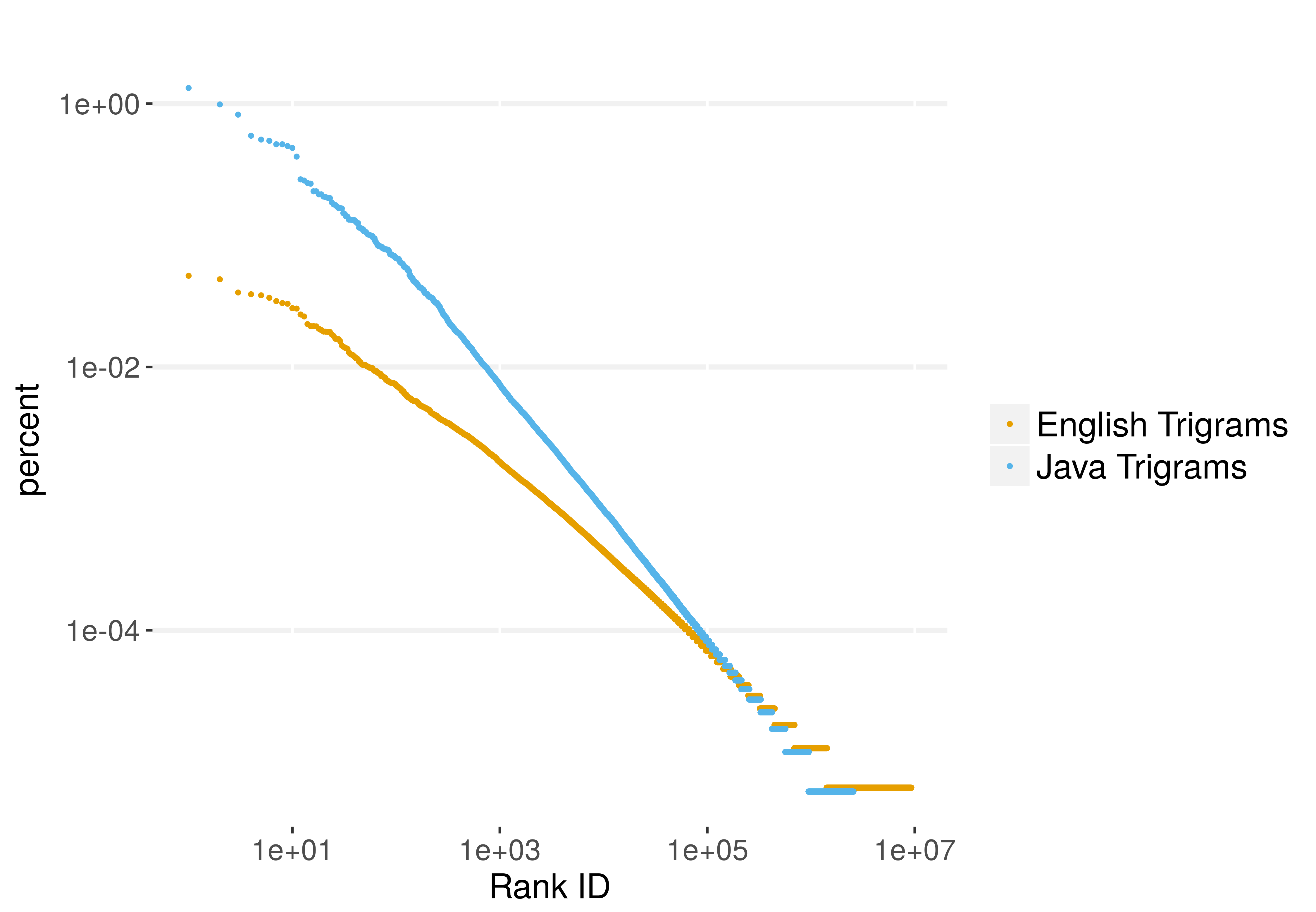}}
  \caption{Comparison of slopes for Zipf plots of Java and English unigrams, bigrams, and trigrams.  The axes are in log scale.  Higher percentages in low ranks indicate a more repetitive corpus, as can be seen by the diverging slopes between Java and English}
  \label{fig:IntroZipf}
\end{figure}

Fig.~\ref{fig:IntroZipf} contains Zipf curves for only Java and English for unigrams, bigrams, and trigrams. The increased repetition of source code over English widens the 
gap between the slopes as the length of the n-gram increases; longer sequences
are repeated even more in Java than in English.  
However, the English curve exhibits a noticeable bend that the Java unigram curve lacks.
This behavior agrees with past studies of such curves in English \citep{ferrer2001two,gerlach2013stochastic,piantadosi2014zipf, mitzenmacher2004brief}, as is better modeled with a bipartite double pareto curve as previously described in \ref{sec:zipf_theory}.

\begin{figure}[ht]
\centering
\subfloat[Unigrams]{\includegraphics[width=.49\textwidth]{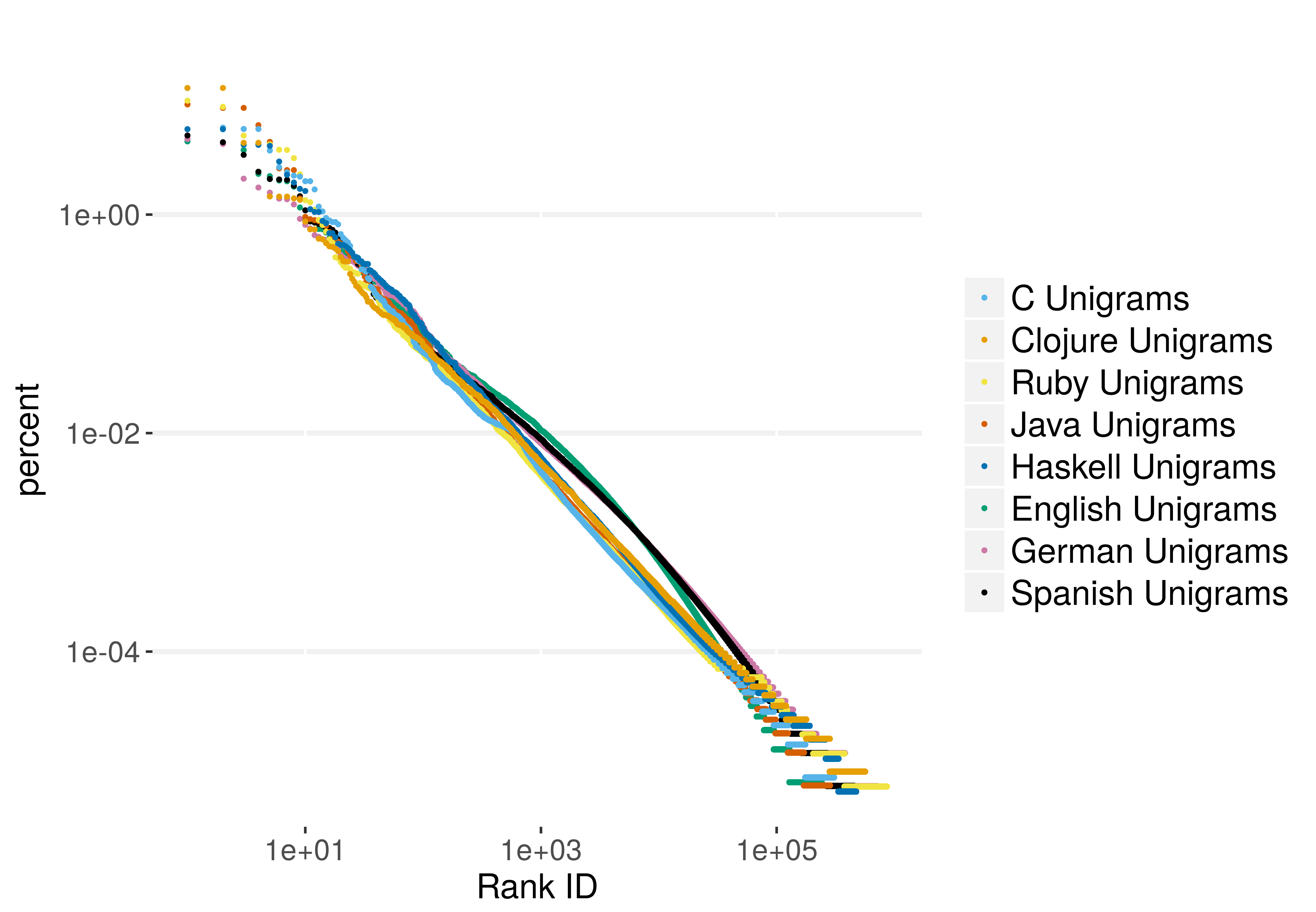}} 
\subfloat[Bigrams]{\includegraphics[width=.49\textwidth]{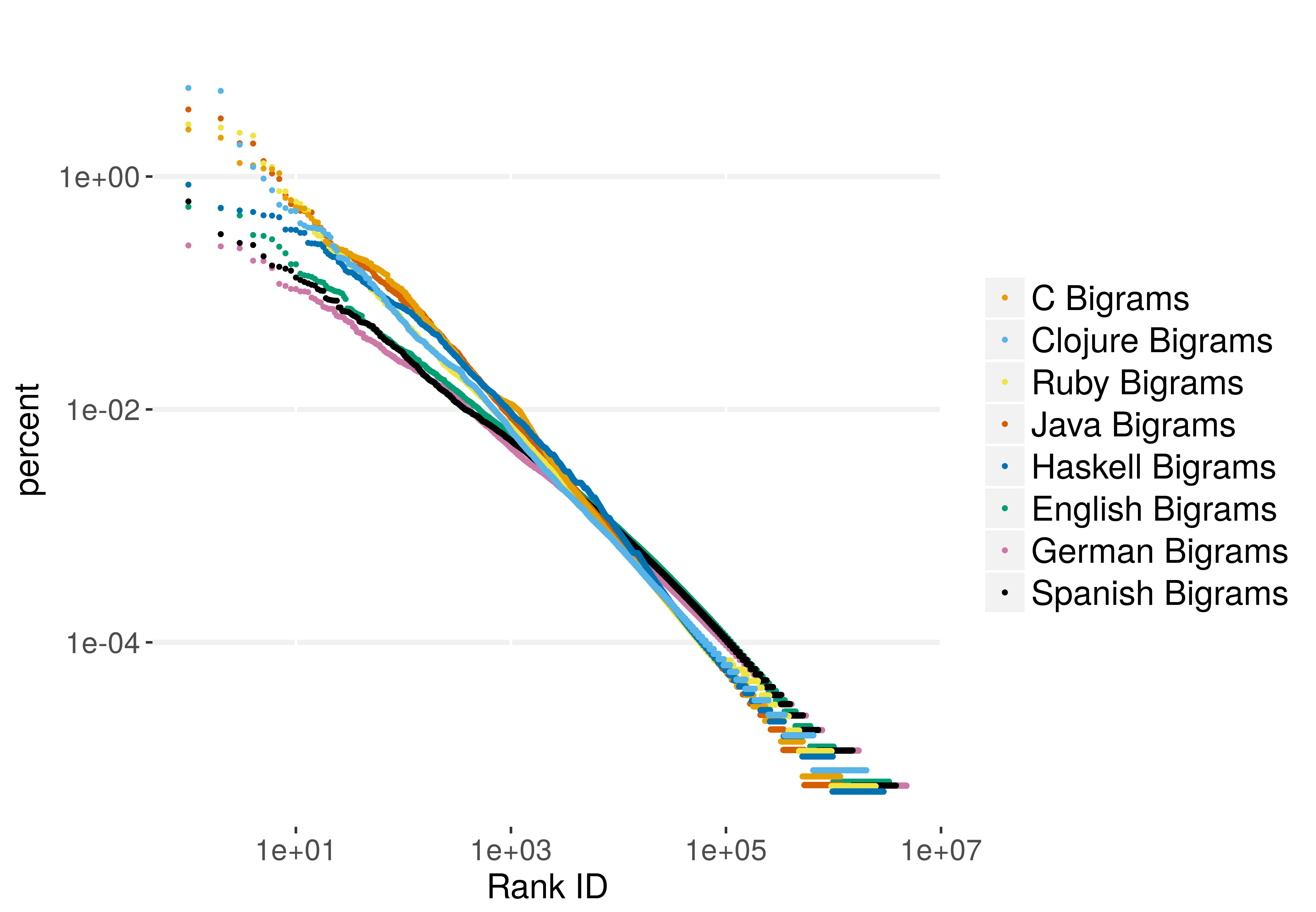}}\\
\subfloat[Trigrams]{\includegraphics[width=.49\textwidth]{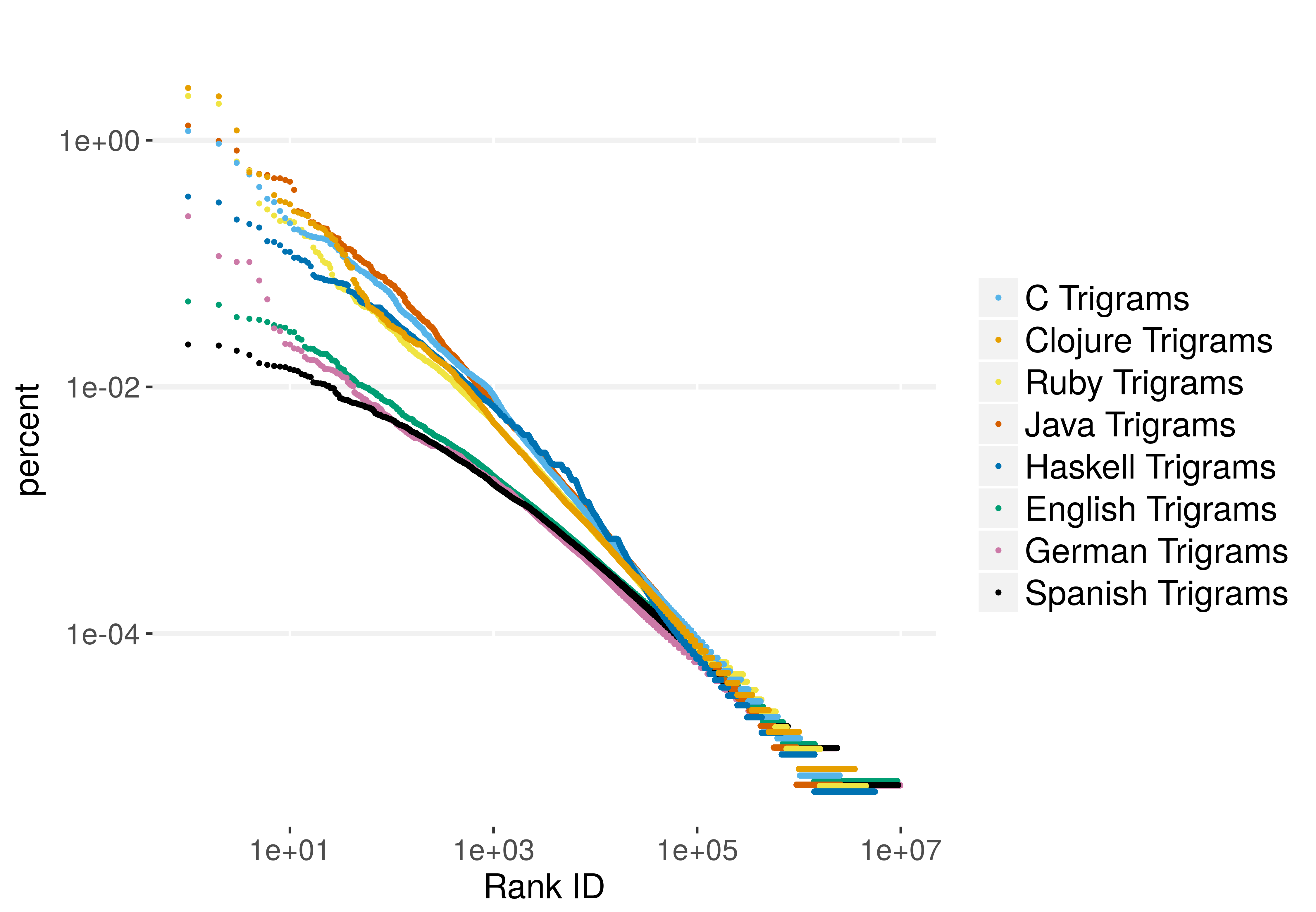}}
  \caption{Unigram, bigram, and trigram Zipf Slopes for all 5 of our different programming languages as compared to our 3 natural language corpora.  The other programming
  and natural languages exhibit similar behavior to Java and English}
  \label{fig:PLVsEngZipf}
\end{figure}

We extend the Fig.~\ref{fig:IntroZipf} Zipf plots to cover all our programming and natural languages in Fig. \ref{fig:PLVsEngZipf}, and a range of behaviors are observed.
All of the programming languages have steeper slopes than the natural language corpora, but not all exhibit the same level of repetition.
Similar to the behavior seen in the ngram models, Haskell bigrams and trigrams fall midway between the natural languages and the other source code languages.
Haskell therefore, is less repetitive and predictable than other programming languages, but not so much as natural language.
The other programming languages are more closely grouped together, with no clear distinction between them.
From here on, while comparing programming \emph{vs} natural languages, we use 
English as a proxy for other natural languages.

\subsection{Modeling just the Open Vocabulary Words}
\label{sec:ResultOpenVoc}

Table~\ref{tab:Corpora} shows the size of two corpora after tokenization before and after closed category word removal.
Three of the programming language corpora (Haskell, Ruby, and Clojure) exhibit a similar amount of closed category word usage as English,
with C and Java having about 10-15\% less proportionately.
Existing work by Allamanis et al. has shown closed category tokens in code to be much more predictable than identifiers \citep{Allamanis2013},
But since English does not have proportionately more open category words than code, we cannot attribute the additional ease of predicting
programming languages simply to an increased amount of closed category tokens.
However, the difference could still result if these closed category tokens are far more predictable in code than in English.
As we shall see shortly, this is not the case.

\begin{table}[ht]
\centering
\caption{Summary of the fraction of open category tokens to all tokens in English and programming languages}
  \begin{tabular}{ | l | c  c |}
    \hline
    & All Tokens &  Open Category Tokens   \\ \hline
    English & 15708917 & 8340284 (53.1\%) \\ \hline
    Java & 16797357 &  6469474 (38.5\%) \\ \hline
    Haskell & 19113708 & 10803544 (56.5\%) \\ \hline
    Ruby & 17187917 & 8992955 (52.3\%)\\ \hline
    Clojure & 12553943 & 6286549 (50.1\%)\\ \hline
    C &  14172588 &  5846097 (41.2\%) \\ \hline
  \end{tabular}
  \label{tab:Corpora}
 \end{table}

\begin{figure*}[ht]
\centering
\subfloat[Unigrams]{\includegraphics[width=.49\textwidth]{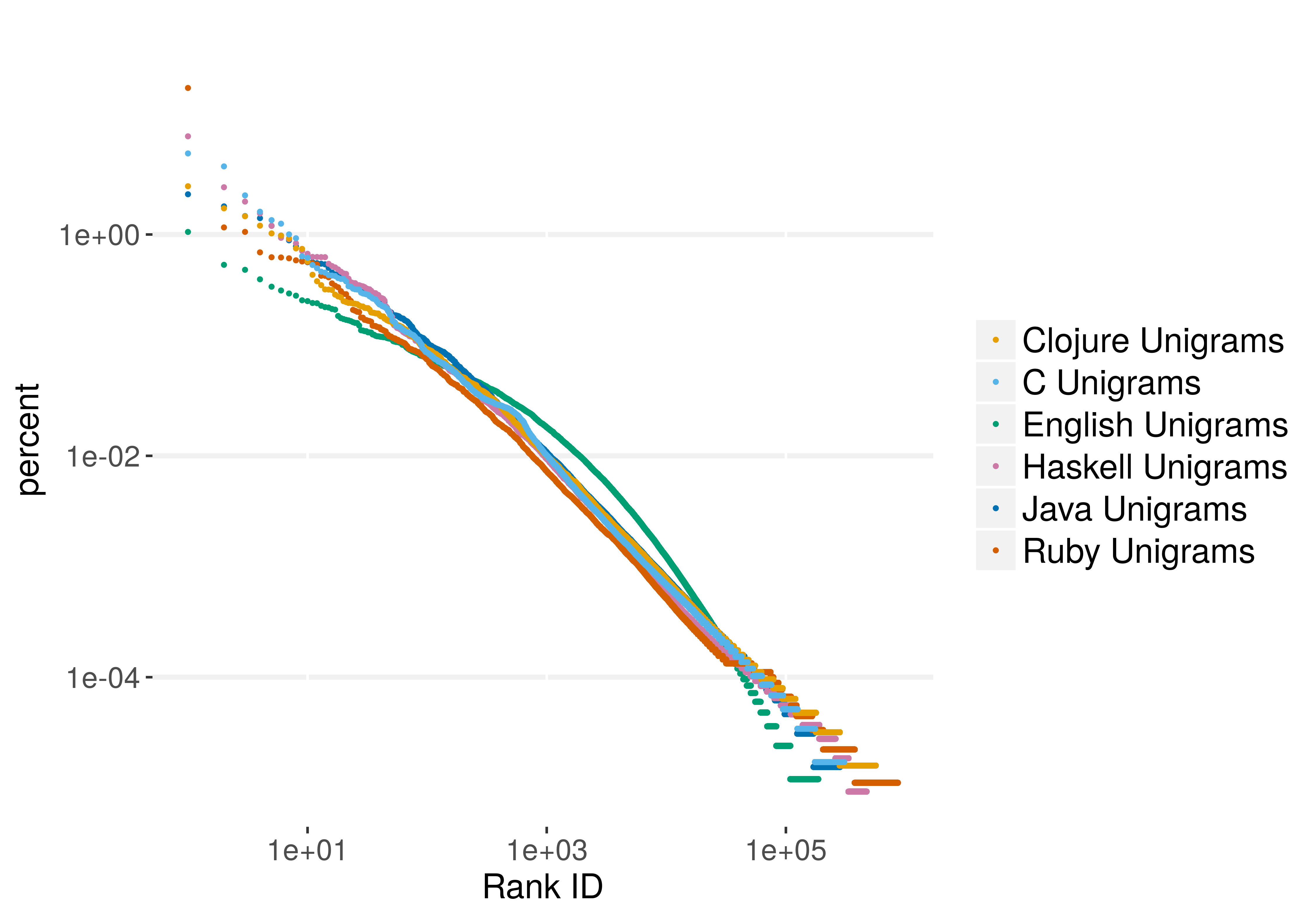}} 
\subfloat[Bigrams]{\includegraphics[width=.49\textwidth]{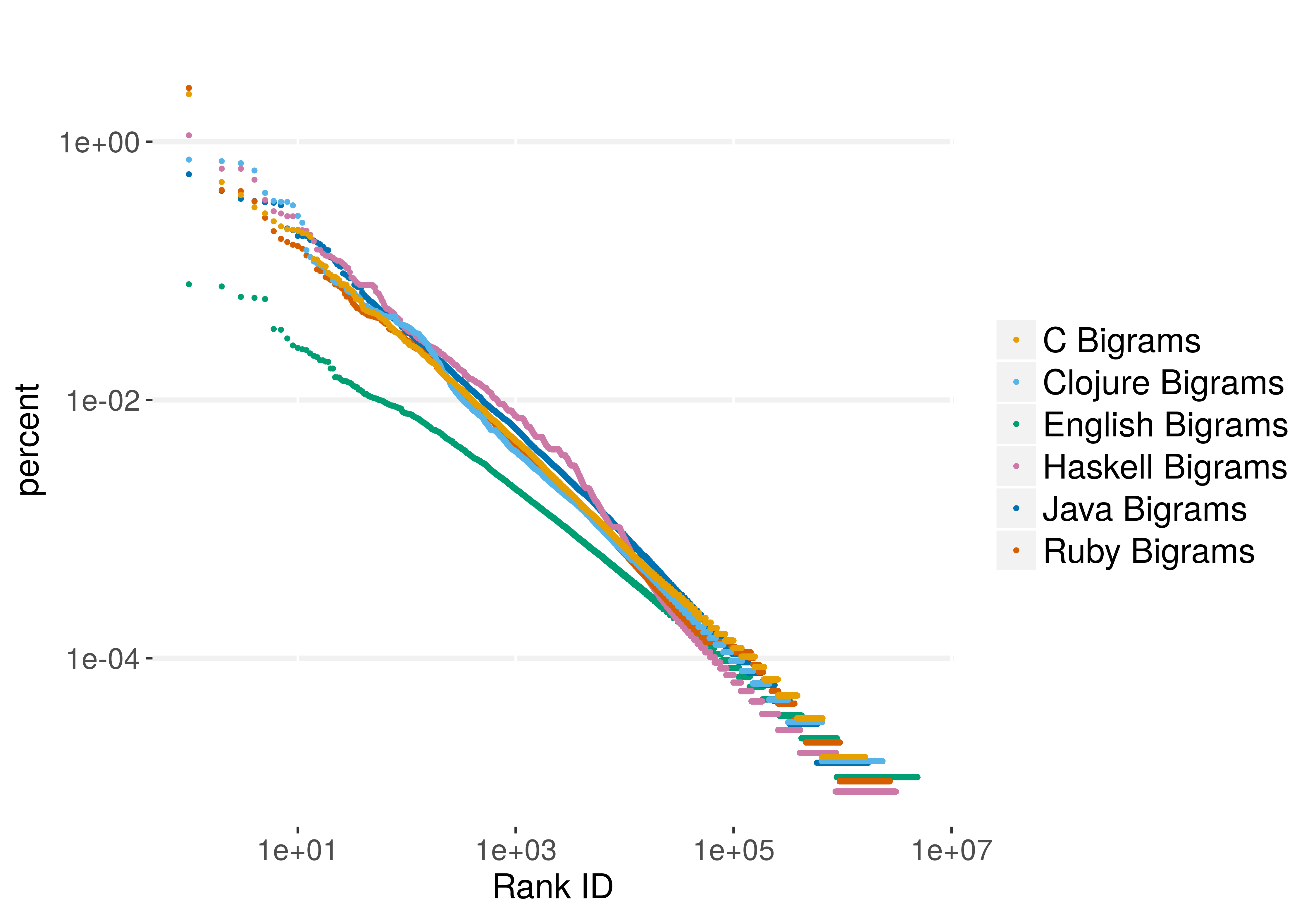}}\\
\subfloat[Trigrams]{\includegraphics[width=.49\textwidth]{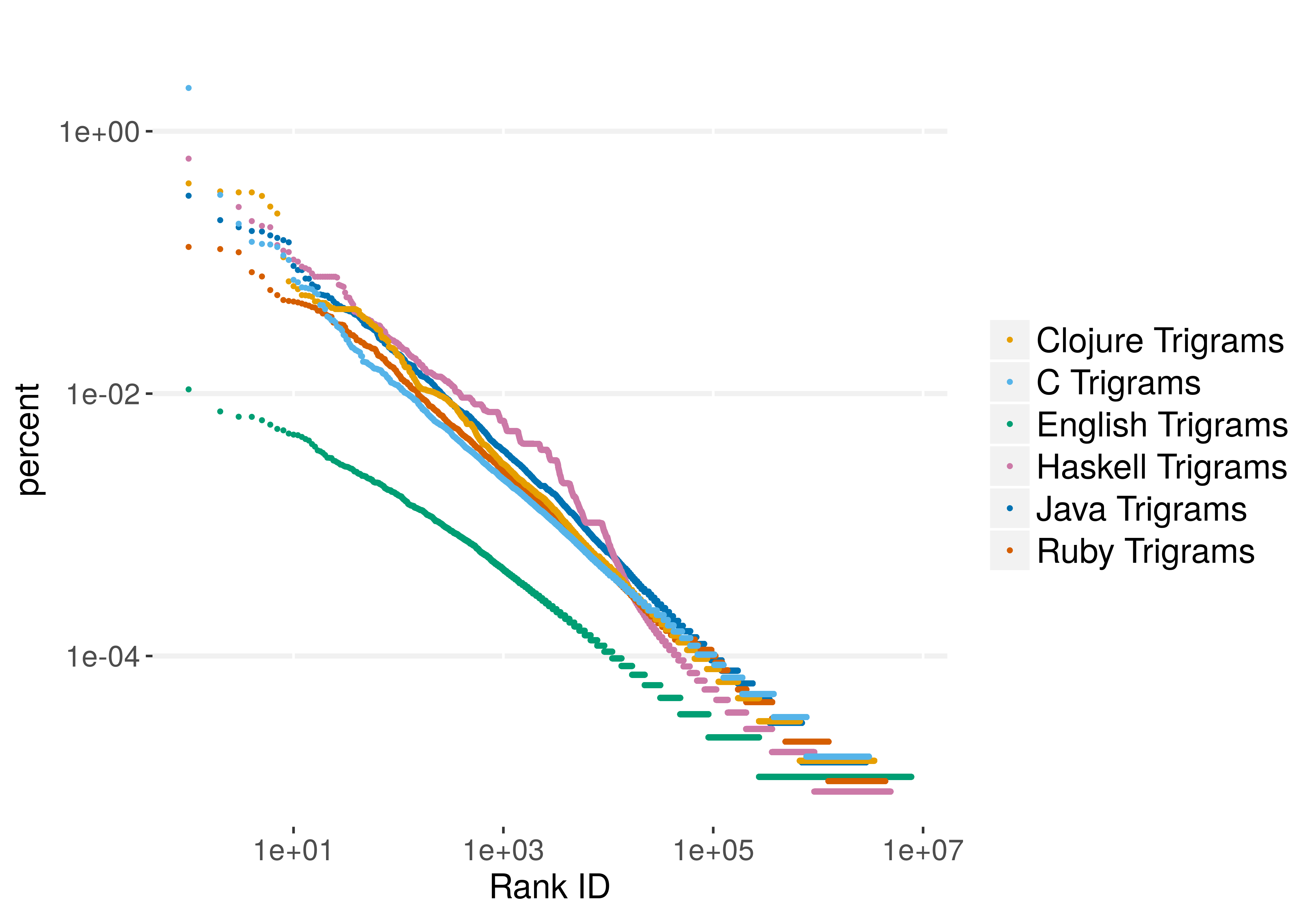}}
  \caption{Unigram, bigram, and trigram Zipf plots comparing English open category words with programming language open category words}
  \label{fig:nameZipf}
\end{figure*}

Fig.~\ref{fig:nameZipf} shows the Zipf slopes of the of the open category-only
unigrams, bigrams, and trigrams.
The unigrams in code are roughly equivalent to that of English, except for the curved nature 
of the Zipf line.
As we move from unigrams to bigrams and then trigrams, we see a similar separation in the Zipf plots lines as
was seen in the full texts.
In all programming languages, the open category word-sequences are more repetitive than English, though
the amount of repetition varies.

 \begin{figure}[ht]
\centering
\subfloat[3 grams]{\includegraphics[width=.49\textwidth]{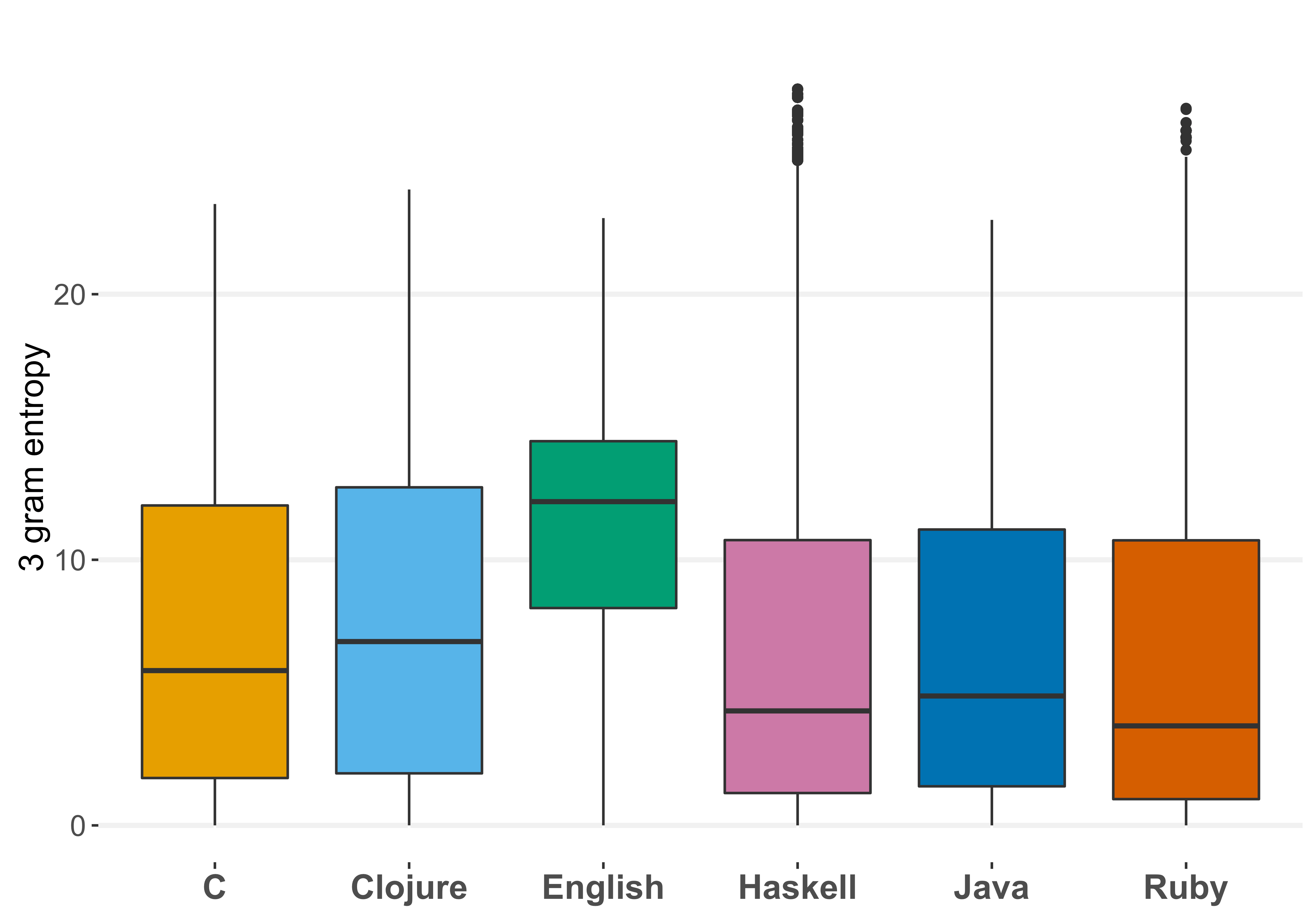}} 
\subfloat[3 grams with cache]{\includegraphics[width=.49\textwidth]{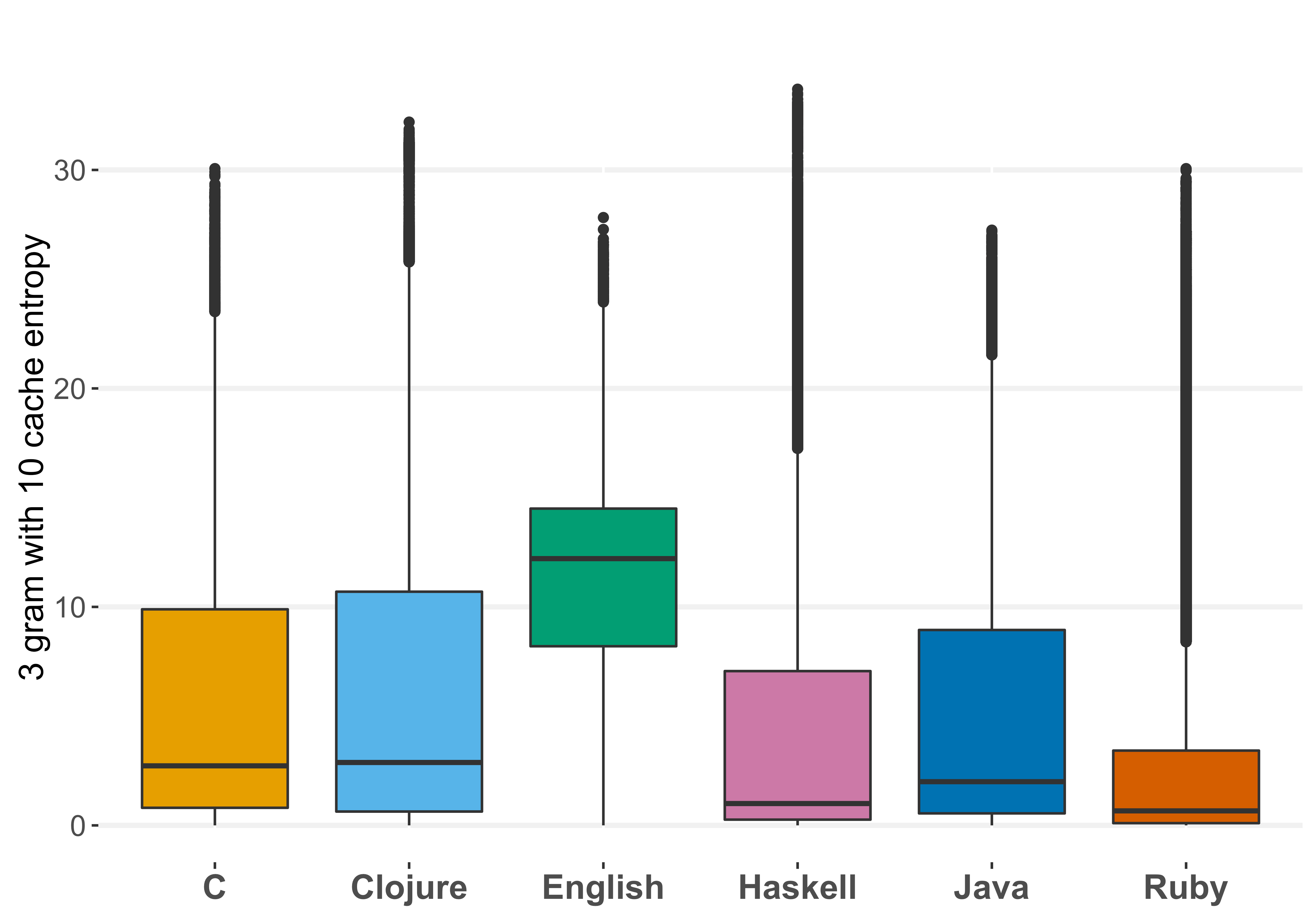}}\\
\subfloat[LSTM (Small)]{\includegraphics[width=.49\textwidth]{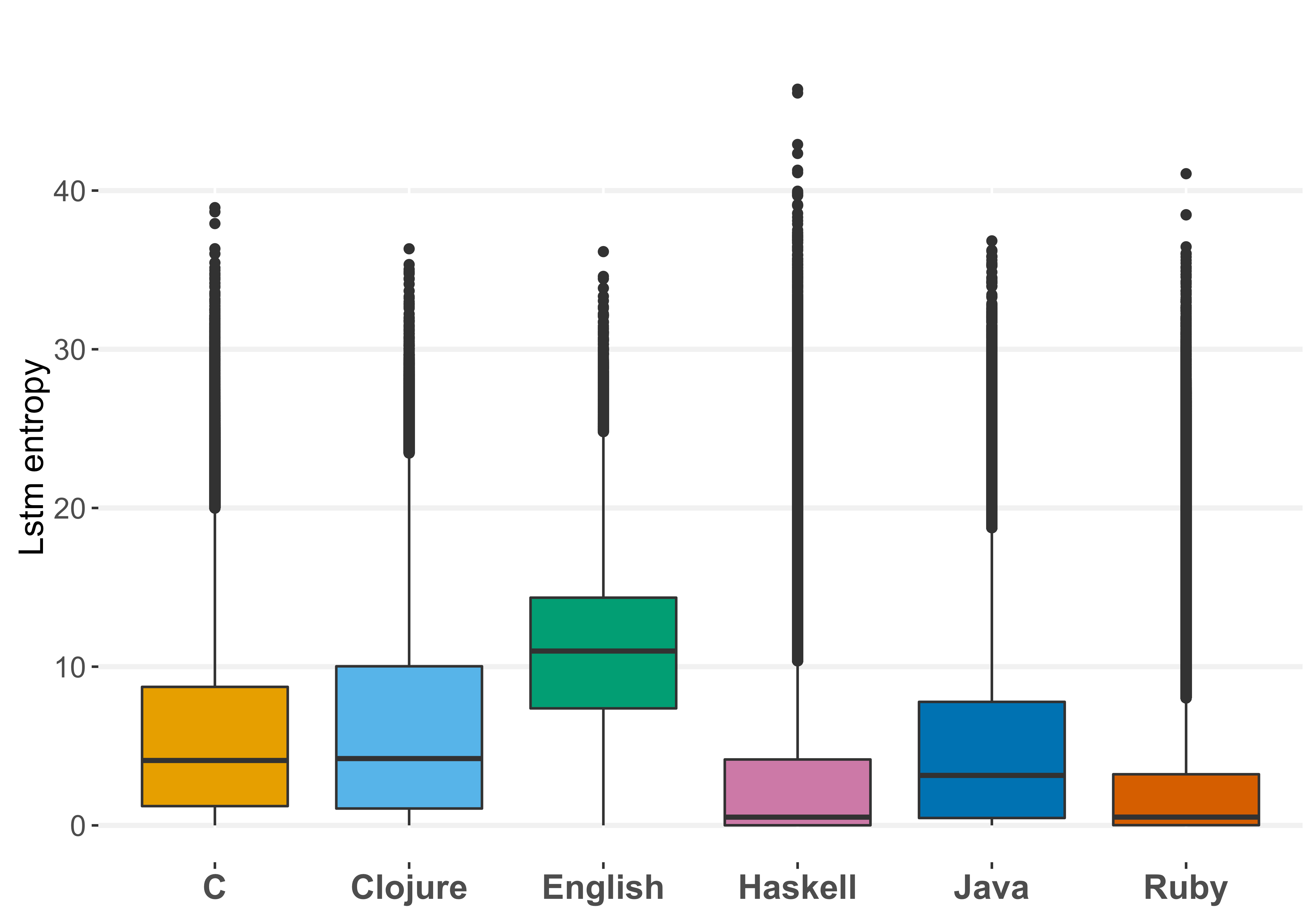}}
  \caption{Entropy distribution comparisons of English and the programming language open category words from an ngram, cache, and LSTM model}
  \label{fig:nameEntropy}
\end{figure}


\begin{table*}[ht]
\centering
\caption{Summary of non-parametric effect sizes and 99\% confidence intervals (in bits) comparing the median of the entropy distribution of open category English words
with those of several programming languages. 
Numbers are marked with * if $p < .05$, ** if $p < .01$, *** if $p < .001$ from a Mann Whitney U test}
\begin{tabular}{| c | c  c  c |}
\hline
Language \textless  English&Ngram&Cache&LSTM\\ \hline
\multirow{2}{*}{Java}  & (-5.507, -5.462) & (-7.377, -7.335) & (-6.618, -6.58) \\ 
 & $0.403^{***}$ & $0.5^{***}$ & $0.505^{***}$  \\ \hline
\multirow{2}{*}{C}  & (-4.715, -4.673) & (-6.858, -6.811) & (-5.826, -5.784) \\ 
 & $0.335^{***}$ & $0.446^{***}$ & $0.435^{***}$  \\ \hline
\multirow{2}{*}{Clojure}  & (-4.112, -4.065) & (-6.641, -6.594) & (-5.463, -5.42) \\ 
 & $0.306^{***}$ & $0.444^{***}$ & $0.397^{***}$  \\ \hline
\multirow{2}{*}{Ruby}  & (-6.22, -6.185) & (-9.61, -9.58) & (-8.941, -8.916) \\ 
 & $0.398^{***}$ & $0.581^{***}$ & $0.618^{***}$  \\ \hline
\multirow{2}{*}{Haskell}  & (-5.857, -5.823) & (-8.372, -8.34) & (-8.628, -8.603) \\ 
 & $0.382^{***}$ & $0.505^{***}$ & $0.58^{***}$  \\ \hline
 \end{tabular}
\label{tab:EngCodeNameEffect}
\end{table*}

Fig.~\ref{fig:nameEntropy} confirms this intuition of content word repetition in source code; the open category words of English are more predictable than those in programming languages. 
Table \ref{tab:EngCodeNameEffect} quantifies these differences with Wilcox tests, showing that the difference for all distributions is significant and varies from
\emph{small} to \emph{medium} effect sizes.
Java, Haskell, and C open category words tend to be more predictable, while Ruby and Clojure names are more difficult to predict.
Haskell's content words are much easier to predict relative to the other languages.
This is notable, especially since Haskell was the most difficult language to predict as a whole.
This means the entropy must be caught up in Haskell's syntax, matching the intuition of Haskell as an information dense language.

When contrasting the median difference in entropy, all of the programming language open category words are at least nearly 4 bits more predictable than the English ones, and the difference is often substantially higher.
In fact, the median difference between the programming languages content words and English context words
is larger than when considering all tokens, though the size of this increase varies.
Additionally, note that when compared to the distributions of entropy of the full corpora seen in Fig~\ref{fig:IntroEntropy}, the the open category words are less predictable, as expected from existing research \citep{Allamanis2013}.
Finally, note that if we exclude the literal values in the code corpora from open category words, we get similar results, though the size of the difference is less, though still larger than in the raw text.
So, while content words are in general less predictable, code content words not only easier to predict than English content words, but also difference in predictability is accentuated!

\subsection{Parse Tree Results}
\label{sec:TreeResult}

 \begin{figure}[ht]
\centering
\subfloat[7 grams]{\includegraphics[width=.49\textwidth]{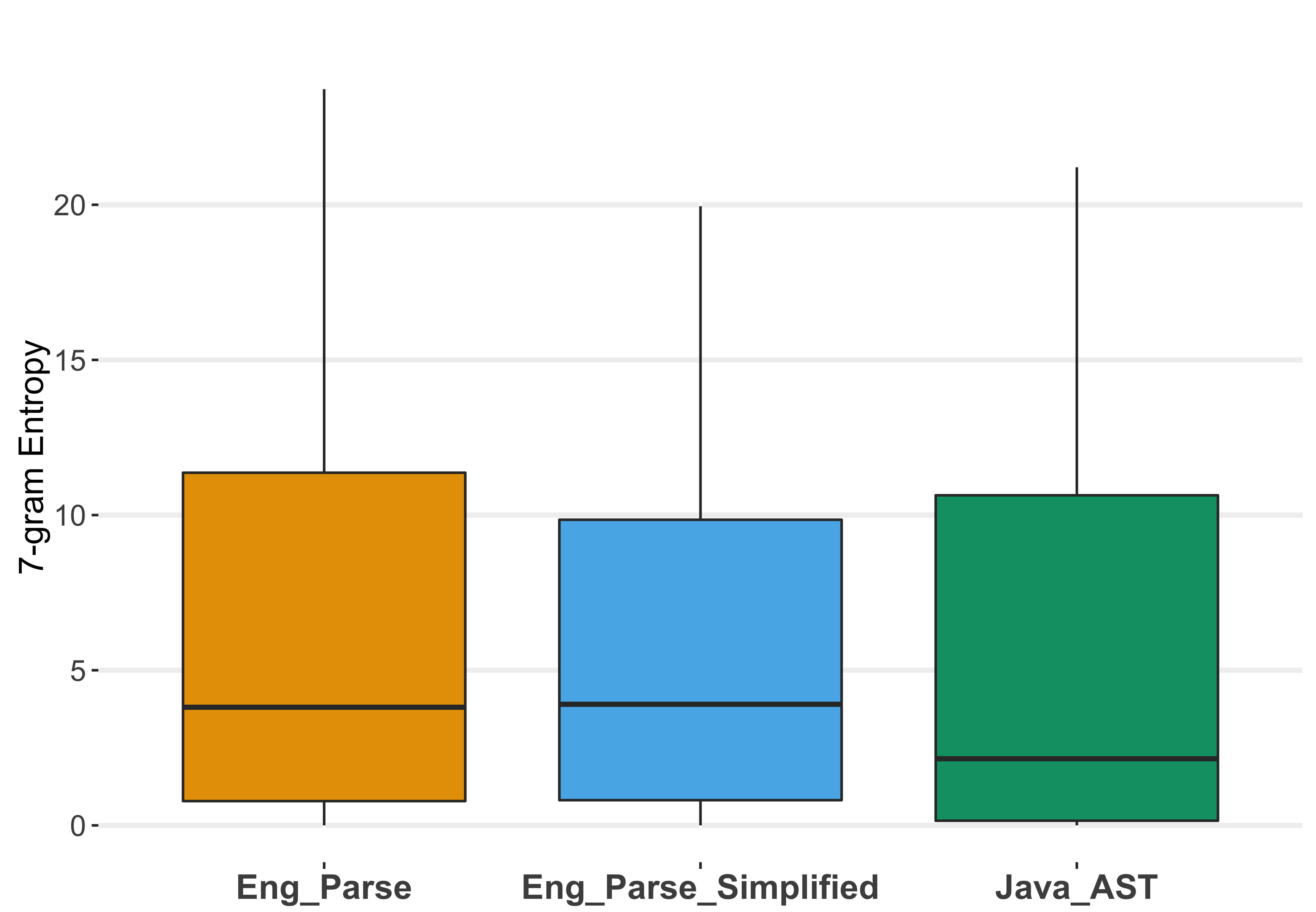}} 
\subfloat[7 grams with cache]{\includegraphics[width=.49\textwidth]{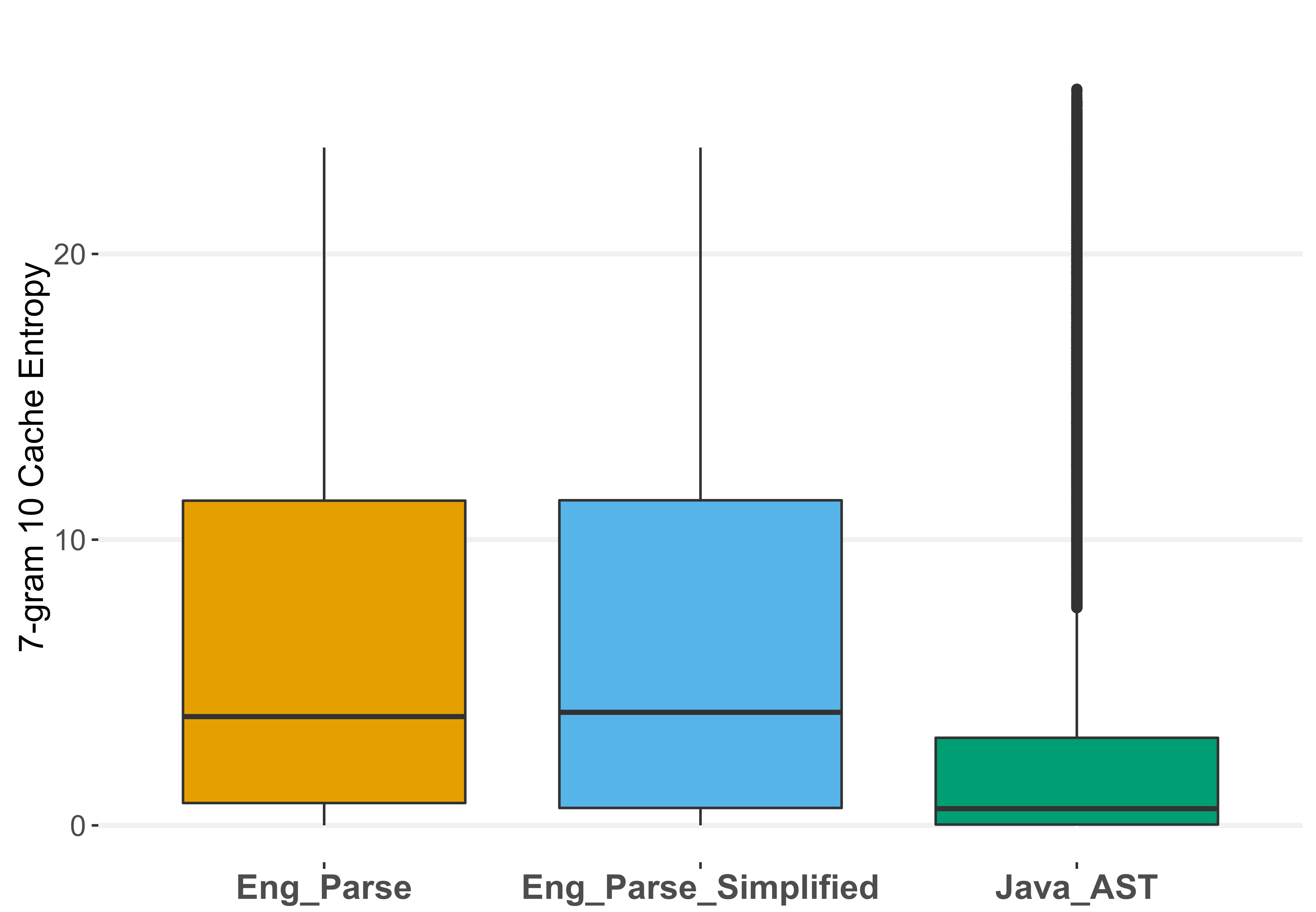}}\\
\subfloat[LSTM (Small)]{\includegraphics[width=.49\textwidth]{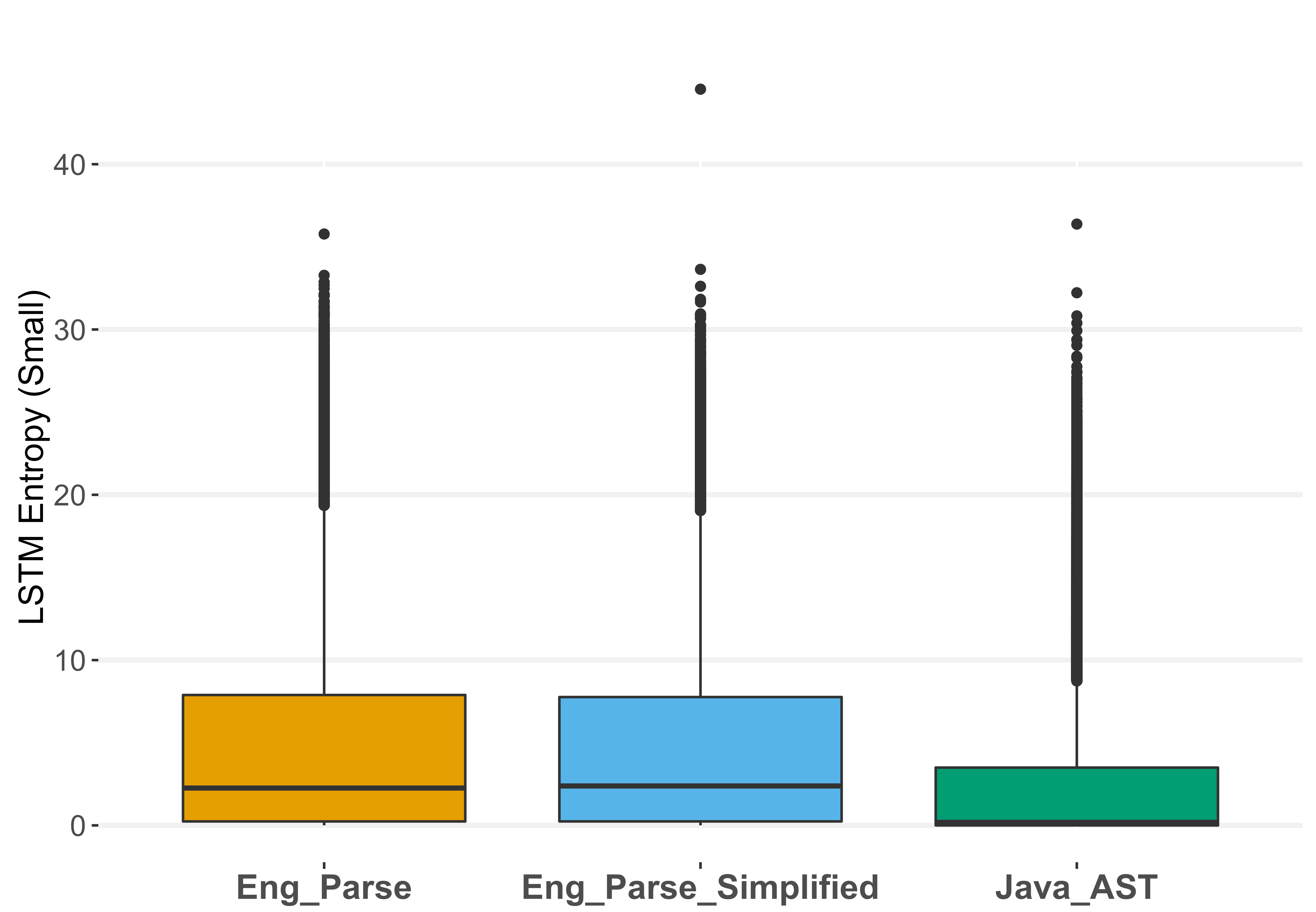}}
\subfloat[LSTM (Medium)]{\includegraphics[width=.49\textwidth]{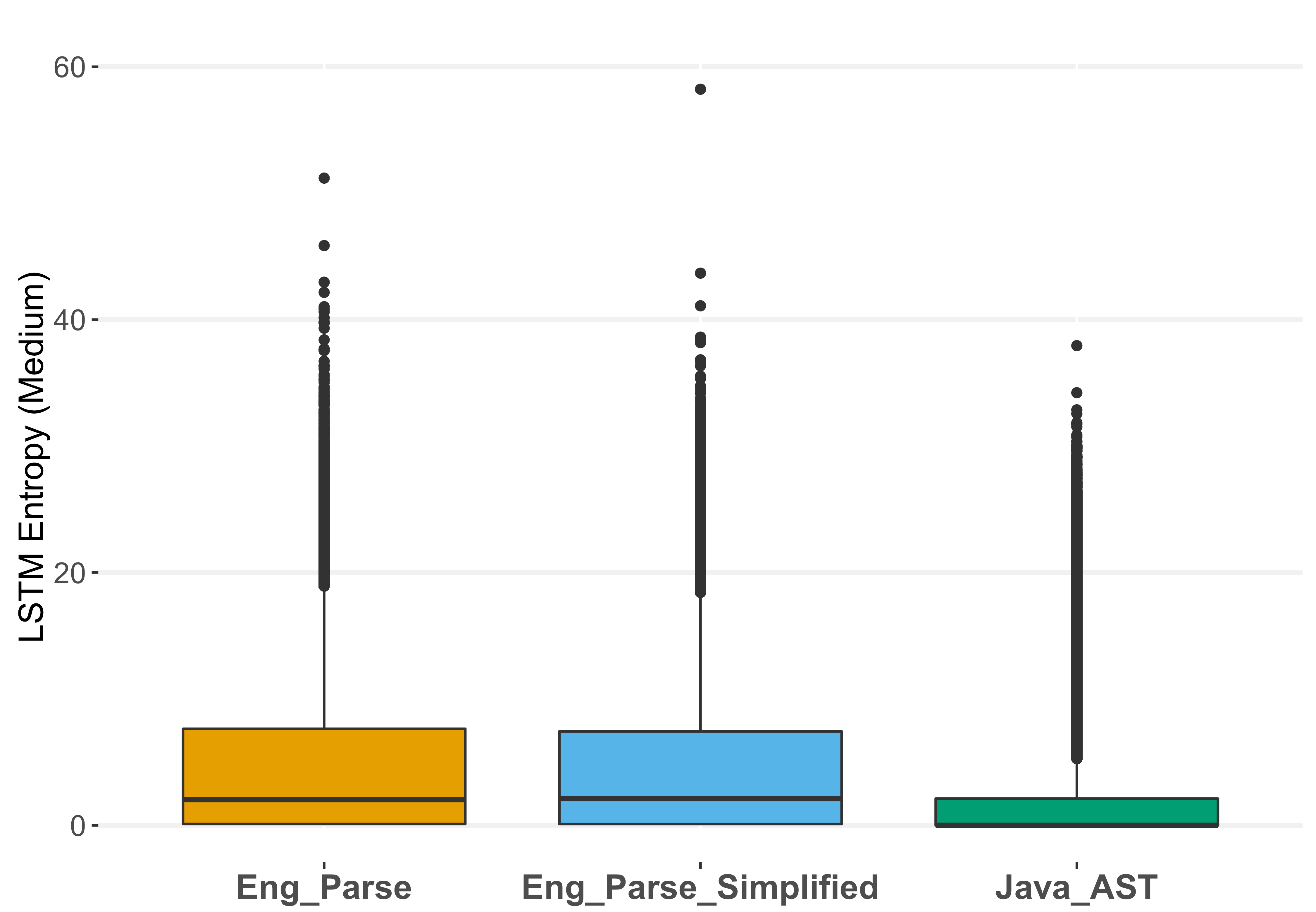}}
  \caption{Entropy comparisons of the terminal tokens in the parse trees using ngram and LSTM models}
  \label{fig:termEntropy}
\end{figure}

\begin{table*}[ht]
\centering
\caption{Summary of non-parametric effect sizes and 99\% confidence intervals (in bits) comparing the difference in the median of the entropy distributions of the terminal tokens in parse trees from Java and the Penn Treebank.  
The differences indicate how much smaller the Java distributions are compared to English.
Rows labelled with simplified are comparing English trees with simplified non-terminals to the Java trees, and rows without it use the original Treebank tags.   Numbers are marked with * if $p < .05$, ** if $p < .01$, *** if $p < .001$
from a Mann Whitney U test}  
  \begin{tabular}{| c |  c  c  |}
    \hline
    Model & Terminal Tokens in Tree & Original Text \\ \hline
    \multirow{2}{*}{Ngram Simplified} & (-0.351, -0.293) & (-3.411, -3.336)  \\ 
    & $0.078^{***}$ & $0.316^{***}$ \\ \hline
    \multirow{2}{*}{Ngram} & (-0.484, -0.435) &   \\ 
    & $0.096^{***}$ &  \\ \hline
    \multirow{2}{*}{Cache Simplified} & (-1.557, -1.499) & (-5.200, -5.116)  \\ 
    & $0.264^{***}$ & $0.479^{***}$ \\ \hline
     \multirow{2}{*}{Cache} & (-1.617, -1.559) & \\ 
    & $0.284^{***}$ &  \\ \hline
    \multirow{2}{*}{LSTM Simplified (Small)} & (-0.866, -0.821) & (-4.0680, -3.985)  \\ 
    & $0.288^{***}$ & $0.413^{***}$ \\ \hline
      \multirow{2}{*}{LSTM (Small)} & (-0.788, -0.748) &   \\ 
    & $0.286^{***}$ & \\ \hline
    \multirow{2}{*}{LSTM Simplified (Medium)} & (-0.907, -0.861) & (-3.441, -3.375)  \\ 
    & $0.352^{***}$ & $0.414^{***}$  \\ \hline
        \multirow{2}{*}{LSTM (Medium)} & (-0.847, -0.801) &  \\ 
    & $0.355^{***}$ &   \\ \hline
  \end{tabular}
  \label{tab:TerminalEntComp} 
\end{table*}

Fig. \ref{fig:termEntropy} shows the entropy comparisons of the \textit{terminal} token distribution
for both Java and English when parse trees are taken into account.
Though we focus primarily on the entropy distributions of the LSTM models, as they is best able to capture the linearized
tree structure, we will mention the ngram and cache model results briefly.
With the ngram model the difference between Java and English drops substantially, albeit not completely.
In contrast, the cache model is able to capture proportionally more of the grammar of Java.
However, neural models are better able to learn the grammar, and in both the smaller 1 layer LSTM and in the larger 2 layer LSTM
Java remains more predictable than English.

We confirm the intuition provided in the box plots in the upper part of Table \ref{tab:TerminalEntComp}.
Each of the differences between the English and Java terminals are significant, and have a \emph{small} effect
size in the more capable LSTM and cache models.
The effect size in the ngram model is very small, but it is questionable how well such a simple model can capture the tree syntax; the LSTM
results are the most reliable. 
The actual difference between Java and English runs from slightly less than a 1 to about 1.5 bits in the cache model.
The concerns about the effect of simplifying the types effecting the comparison of grammar were unfounded.
Using Wilcox tests to compare the simplified and the full non-terminal set revealed no significant difference in the more reliable LSTM
models, and a significant but extremely small effect in the ngram and cache models.
Finally, to ensure a fair comparison between these languages as parse trees and them as raw text, Table \ref{tab:TerminalEntComp}
has a column \emph{Original Text}.
These are the same set as the terminal tokens in the tree, but with all tree information removed before language model processing.
We see that in the original text, the effect sizes and confidence intervals are all larger, with almost \emph{medium effect sizes} and gaps far greater than 1 bit of difference.
Therefore, we can conclude that eliminating the ambiguity of English grammar explains \emph{some}, but not all of the difference in 
repetition of the language compared to Java.

Additionally, with our medium LSTM 23.4\% (small LSTM had 9.9\%) of Java terminals had entropy 0, meaning the choice was completely determined by the grammar.
In contrast, in the medium LSTMs only about 4.9/4.8\% (for the simplified and unsimplified tree) of English terminals had 0 entropy.
The small LSTMs had .8\%/1.6\% tokens that were completely predictable in the English simplified/unsimplified trees.
These tokens primarily consisted of the punctuation of each language, with occasionally stop words or reserved words in Java.
In English, the largest contributor to  low-entropy  tokens were commas, and in Java it was open parentheses, the dot operator,  open brackets, and
closing parentheses in decreasing order.  The other tokens only made much smaller portions of the 0 entropy tokens.

Both this experiment and the previous one
suggest that the differences seen between
source code and English consist of more than simply syntactic differences.
This leaves the possibility that at least some of the difference comes from human choices independent from the grammar, though it is unclear what
components may influence these choices.
As previously discussed, we now explore some possibilities by comparing some specialized English texts that share properties 
with source code to both our more typical English and Java corpus.
We limit the presentation of our results to Java comparisons, but we found similar results were found 
when comparing the other programming language corpora as well.

\subsection{Language Proficiency}
\label{sec:LangProfResult}

\begin{table*}[ht]
\centering
\caption{Summary of non-parametric effect sizes and 99\% confidence intervals (in bits) of the median entropy comparing the English Language Learner corpora with Java and the balanced English Brown corpus.  Numbers are marked with * if $p < .05$, ** if $p < .01$, *** if $p < .001$ from a Mann Whitney U test}
\begin{tabular}{| c | c c c |}
\hline
Brown \textgreater  Language &Ngram&Cache&LSTM\\ \hline
\multirow{2}{*}{Gachon}  & (-1.673, -1.595) & (-1.558, -1.475) & (-3.767, -3.674) \\ 
 & $0.152^{***}$ & $0.133^{***}$ & $0.294^{***}$  \\ \hline
\multirow{2}{*}{TECCL}  & (-1.729, -1.657) & (-1.643, -1.564) & (-4.194, -4.106) \\ 
 & $0.147^{***}$ & $0.132^{***}$ & $0.309^{***}$  \\ \hline
Language \textgreater  Java (Small)& & &\\ \hline
\multirow{2}{*}{Gachon}  & (-1.575, -1.501) & (-4.079, -4.004) & (-2.461, -2.398) \\ 
 & $0.153^{***}$ & $0.434^{***}$ & $0.341^{***}$  \\ \hline
\multirow{2}{*}{TECCL}  & (-1.501, -1.429) & (-4.046, -3.972) & (-2.043, -1.988) \\ 
 & $0.138^{***}$ & $0.403^{***}$ & $0.296^{***}$  \\ \hline
\end{tabular}
\label{tab:JavaBrownLangProficiency}
\end{table*}

\begin{figure}[ht]
\centering
\subfloat[Unigrams]{\includegraphics[width=.49\textwidth]{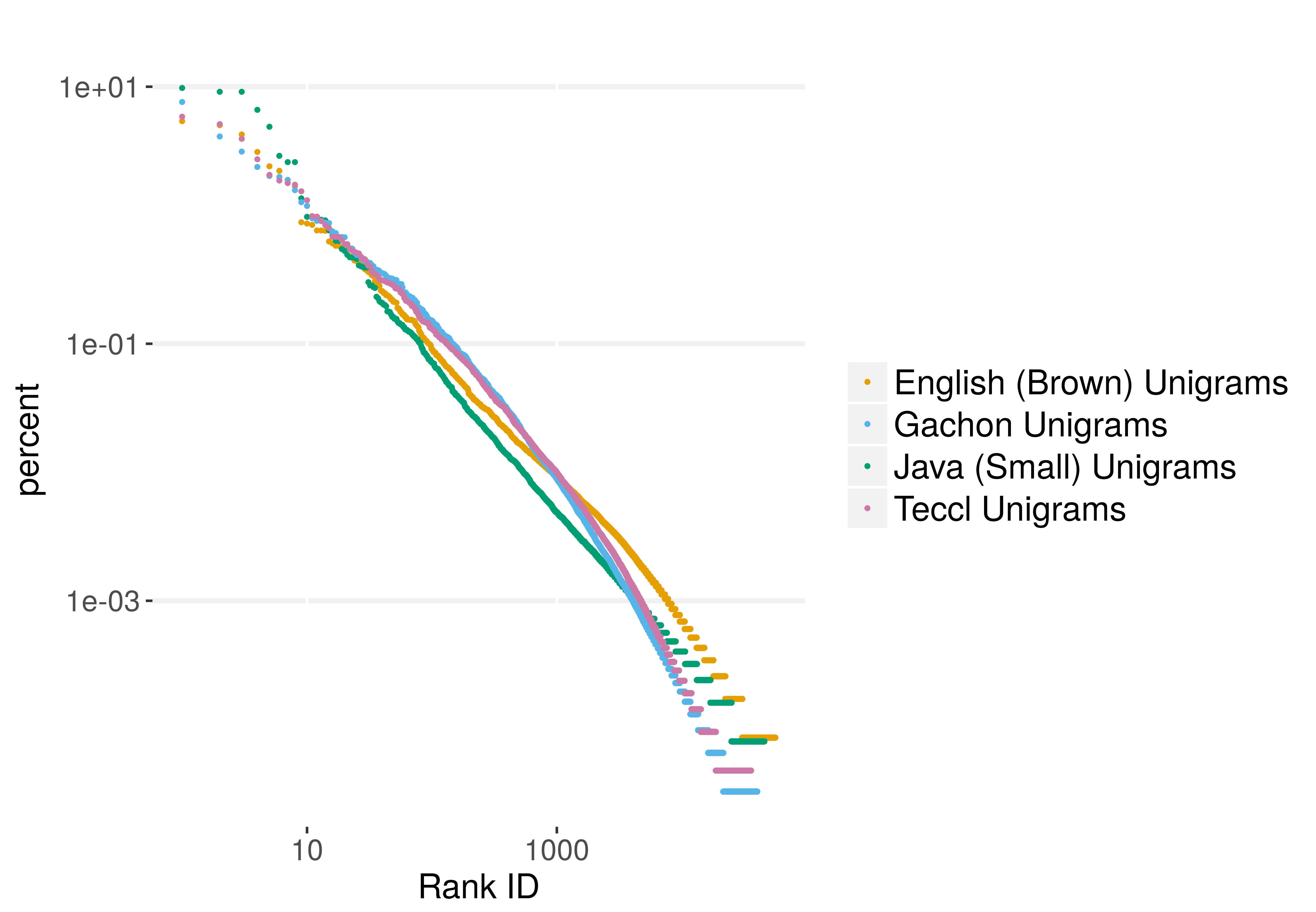}} 
\subfloat[Bigrams]{\includegraphics[width=.49\textwidth]{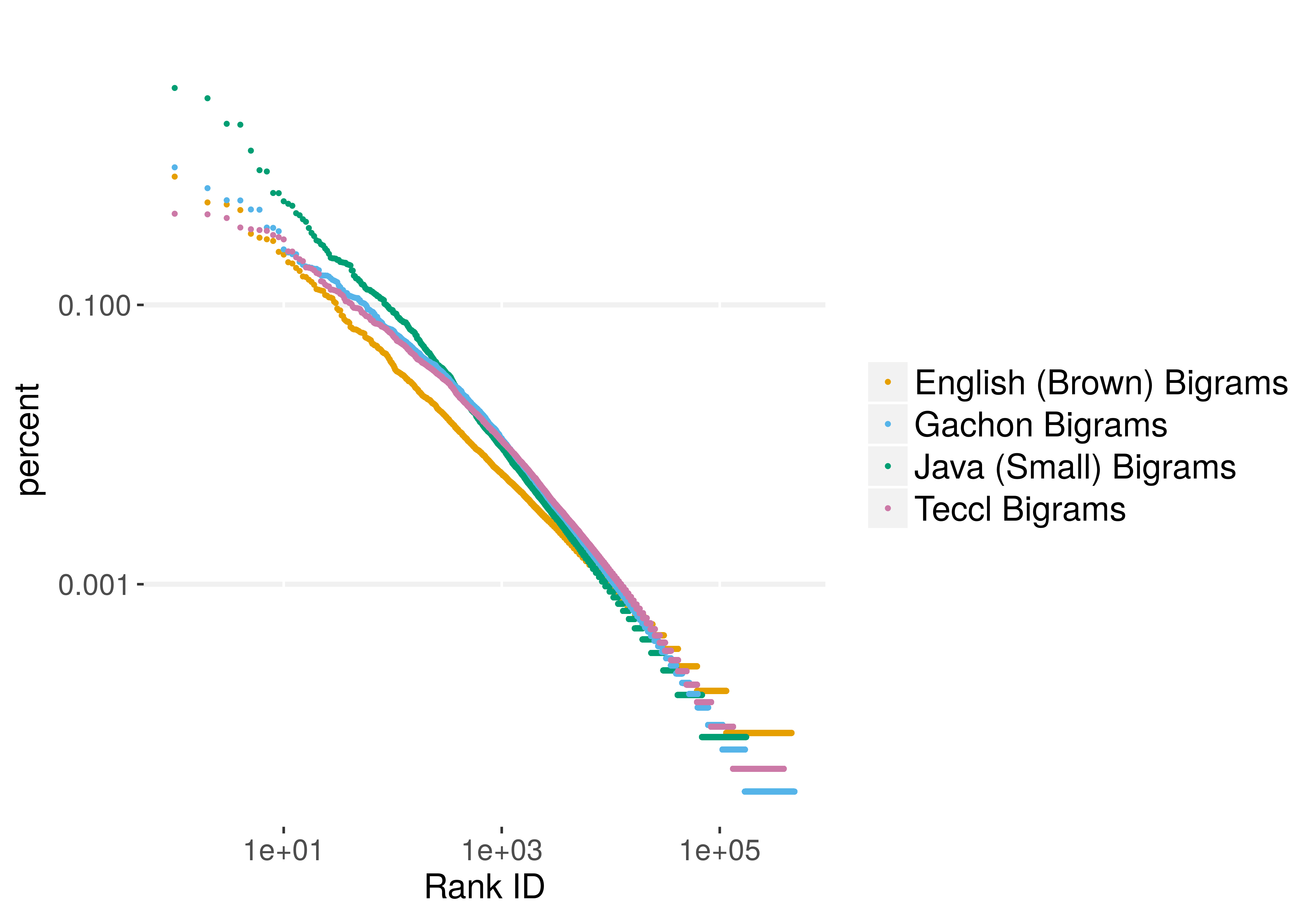}}\\
\subfloat[Trigrams]{\includegraphics[width=.49\textwidth]{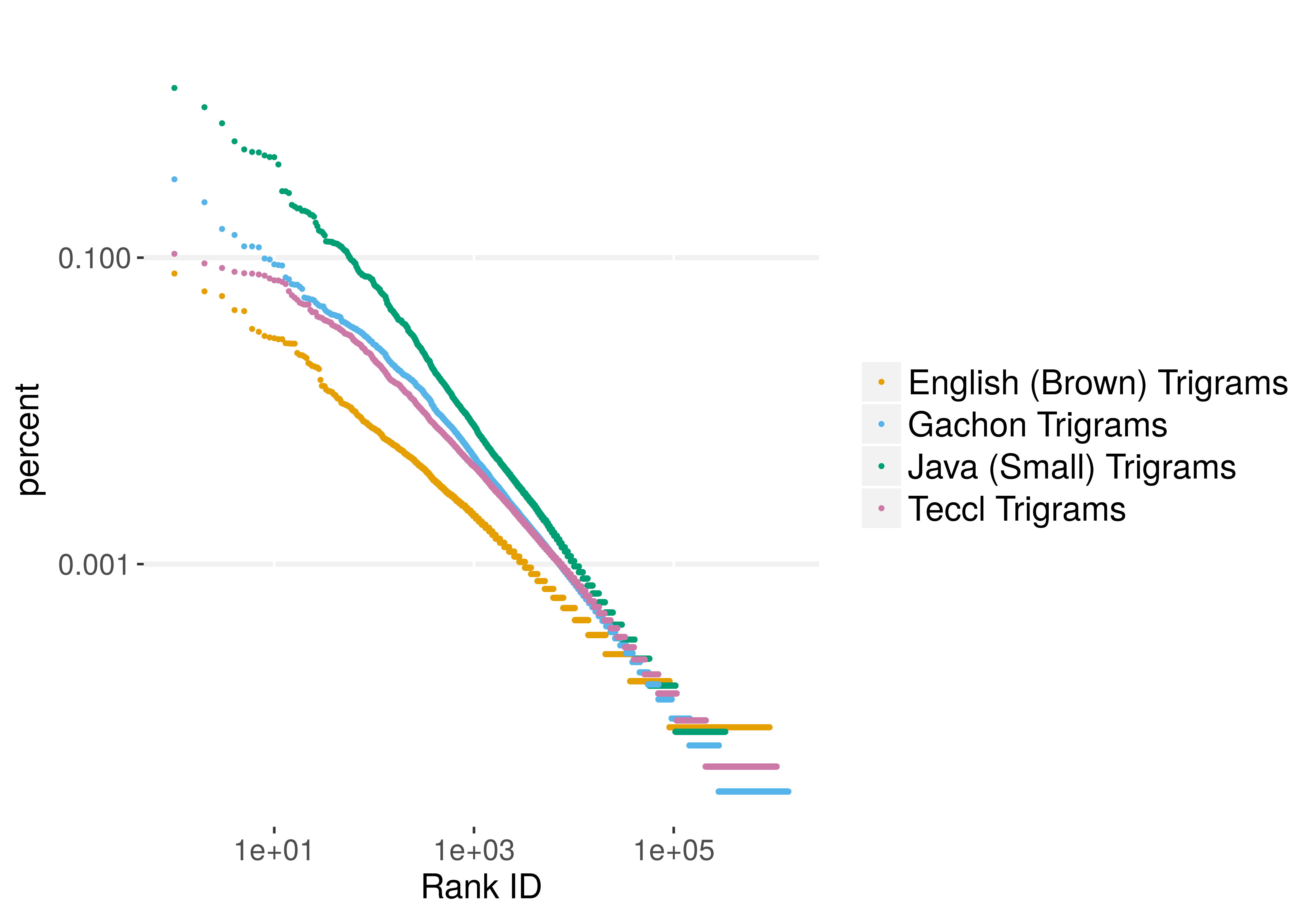}}
  \caption{Zipf plots for the unigrams, bigrams, and trigrams of the general English, Java, and English language learner corpora}
  \label{fig:termZipf}
\end{figure}

Fig. \ref{fig:termZipf} shows  Zipf plots comparing English with our ESL (English as a second language) corpora and Java. 
ESL is certainly more repetitive than general purpose English; however, it is not as repetitive as source code.
This behavior is confirmed with the language models displayed in Fig \ref{fig:termEntropy}.
Regardless of where the more basic trigram model or the increasing the context with the LSTM model,
the entropy, like the Zipf slope lines, fall in between source code and general native language written corpora.
Table \ref{tab:JavaBrownLangProficiency} reports p-values, confidence intervals, and effect sizes and confirms that the english language learner texts
fall fairly evenly between native English and Java.
The one exception is the when using the cache model, where code gains comparatively over both fluent and learner english.
Neither exhibits the locality needed to benefit from this model's assumptions.

Thus, the behaviors exhibited by the foreign language learner corpora are consistent with the the hypothesis that less fluency and
greater difficulty would result in utterances that more closely resemble source code.

 \begin{figure}[ht]
\centering
\subfloat[3 grams]{\includegraphics[width=.49\textwidth]{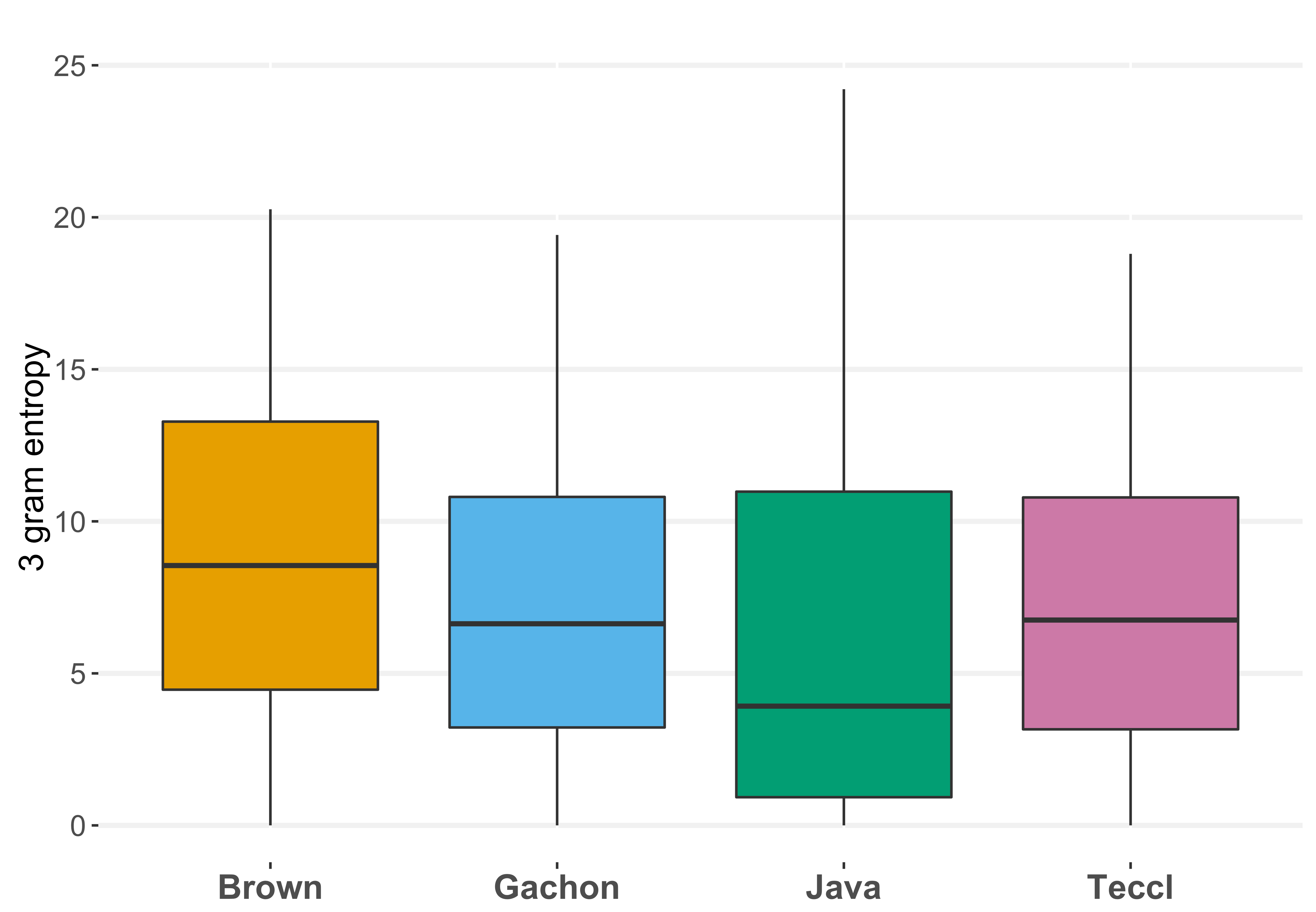}} 
\subfloat[3 grams with cache]{\includegraphics[width=.49\textwidth]{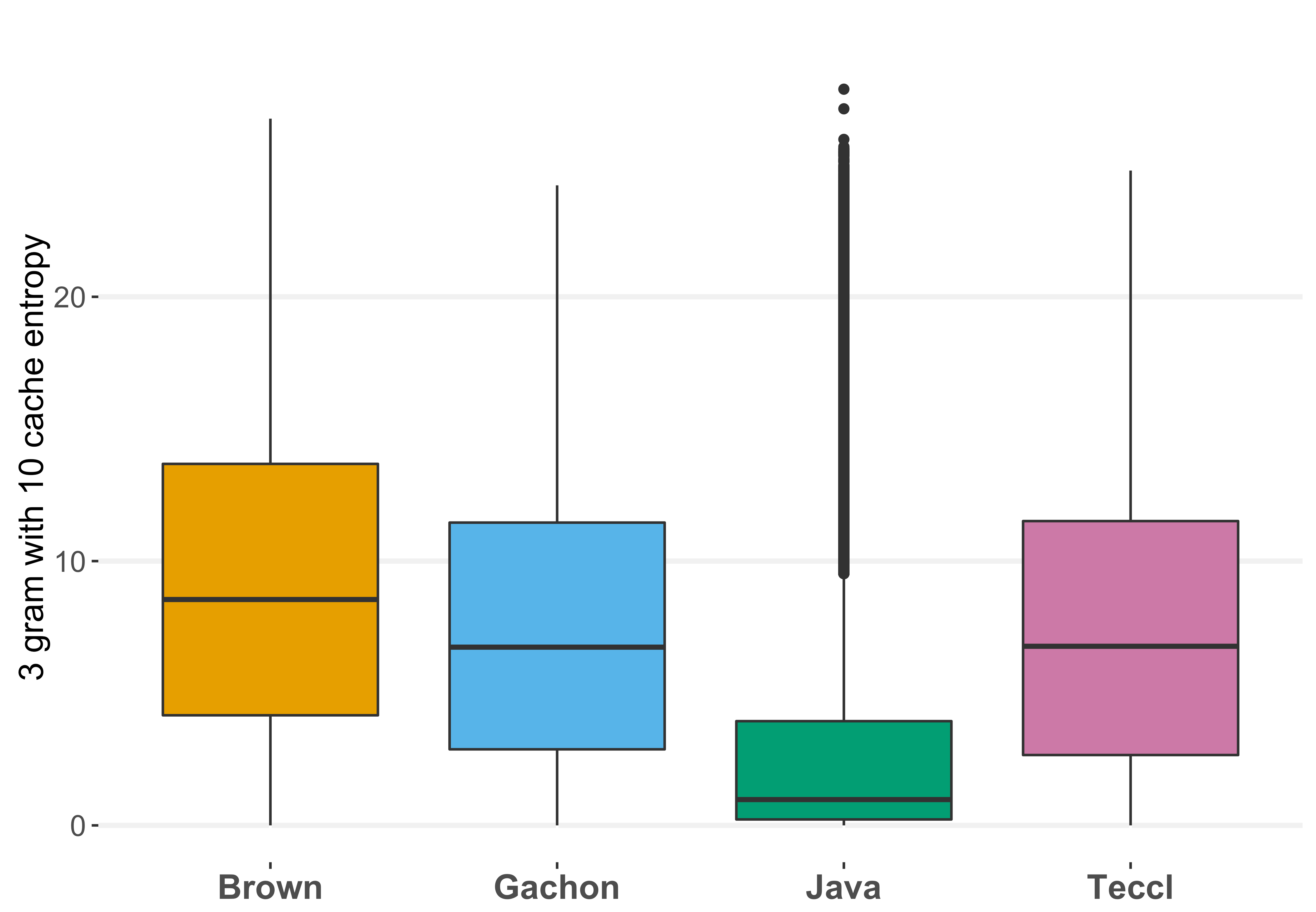}}\\
\subfloat[LSTM (Small)]{\includegraphics[width=.49\textwidth]{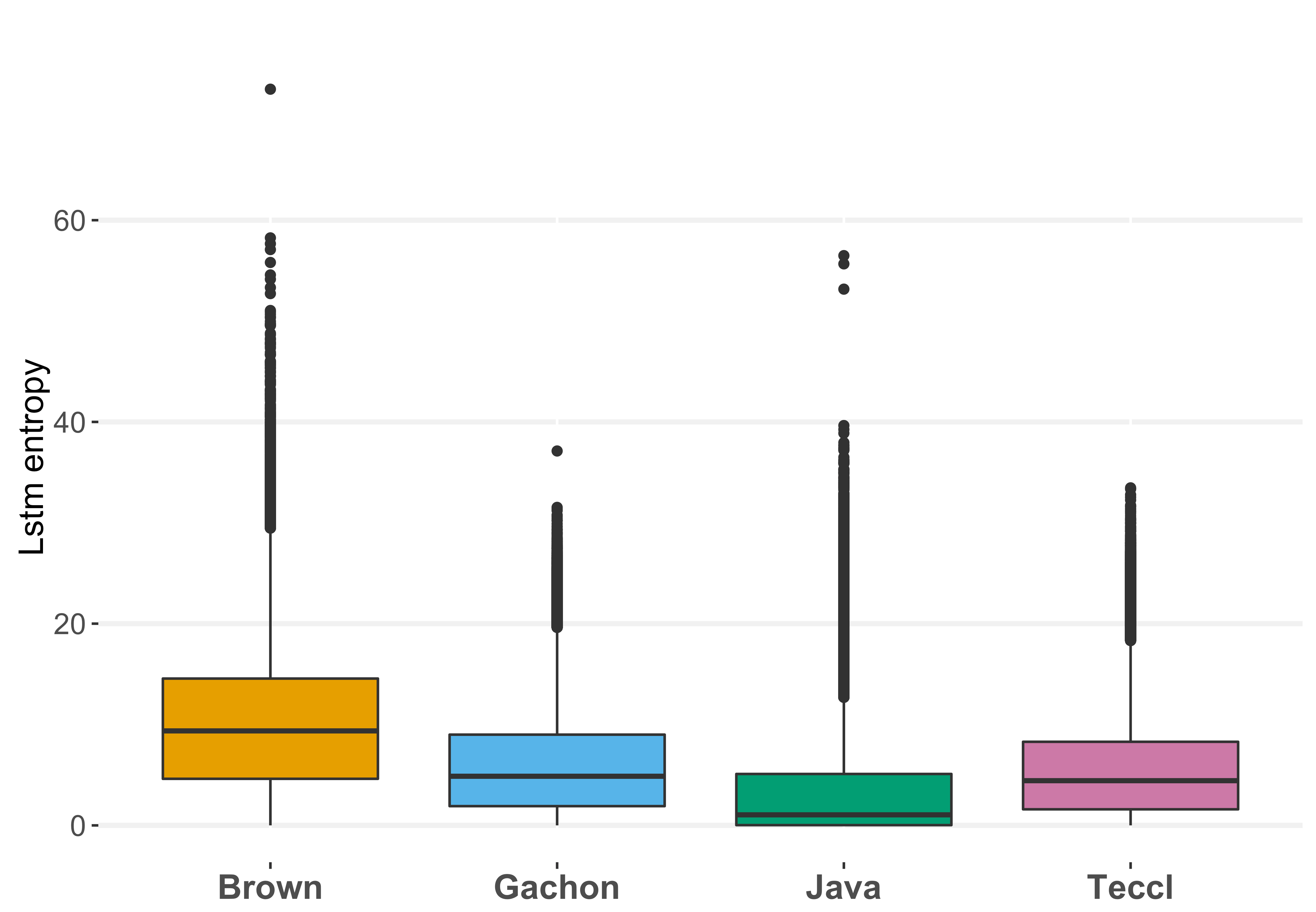}}
  \caption{Entropy comparisons of the of the English language learners corpora with Java and English Corpora using the LSTM and best trigram models}
  \label{fig:eflEntropy}
\end{figure}

\subsection{Comparing Technical and Non-Technical Corpora}
\label{sec:TechResult}

 \begin{figure}[ht]
\centering
\subfloat[Unigrams]{\includegraphics[width=.49\textwidth]{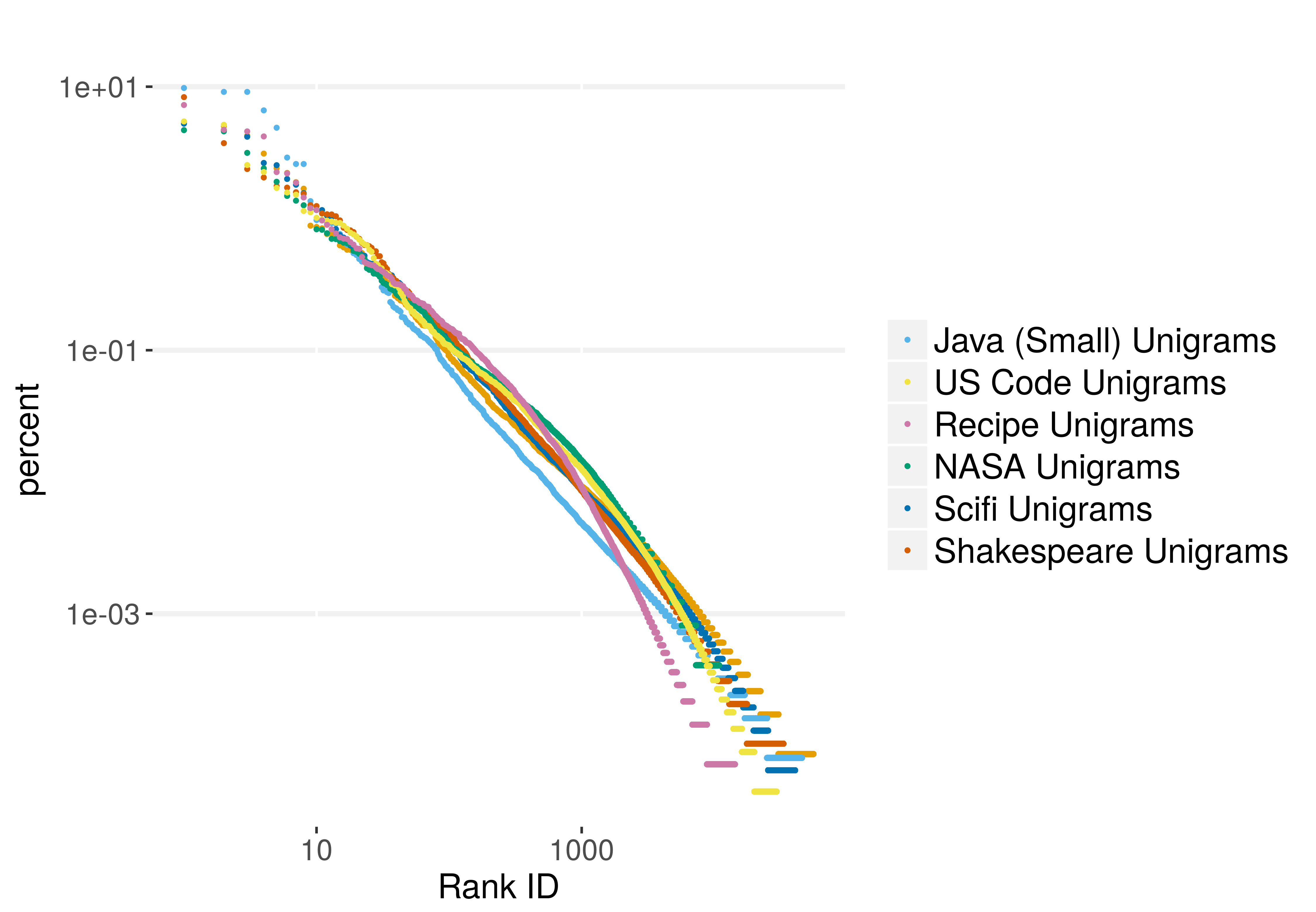}} 
\subfloat[Bigrams]{\includegraphics[width=.49\textwidth]{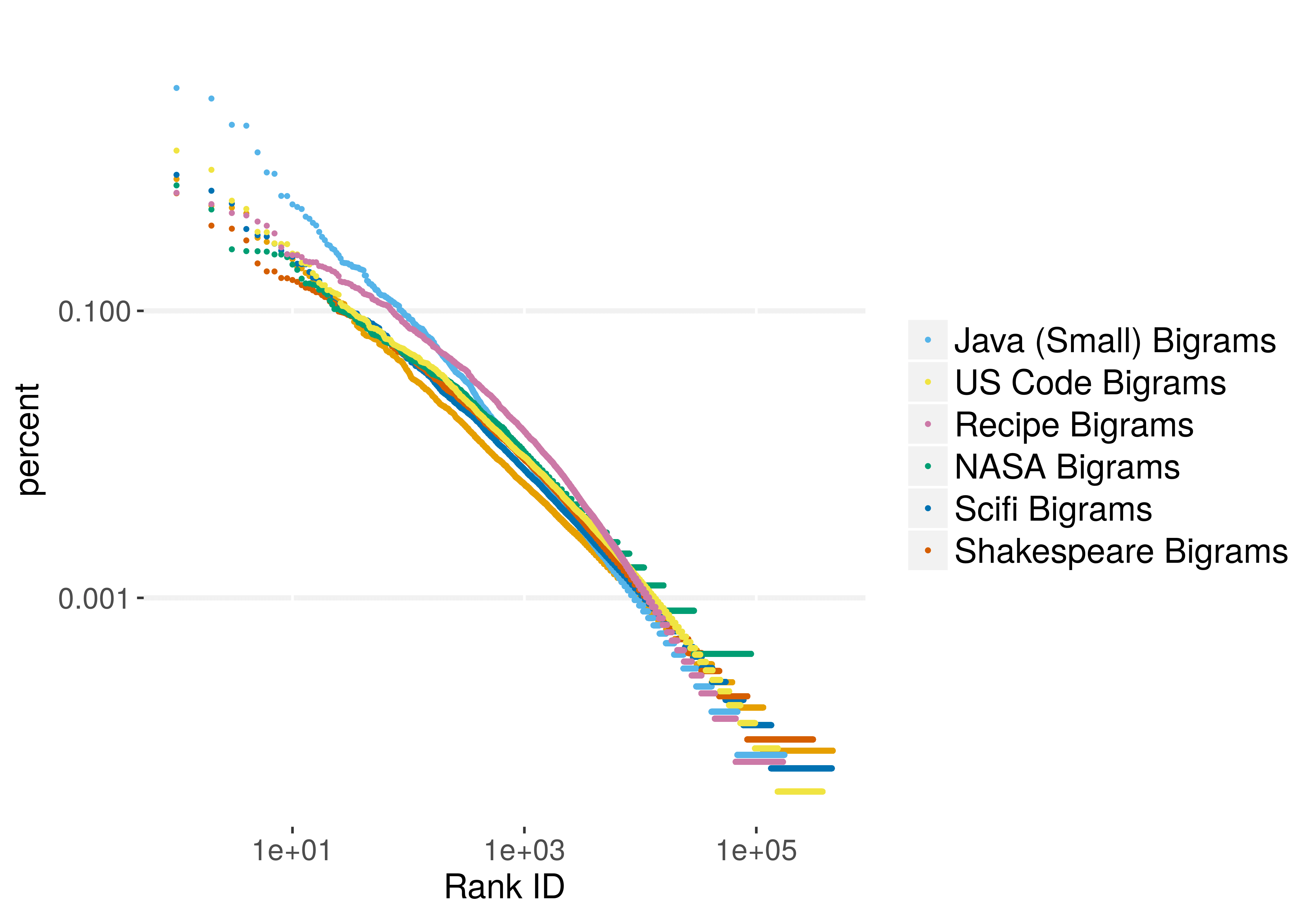}}\\
\subfloat[Trigrams]{\includegraphics[width=.49\textwidth]{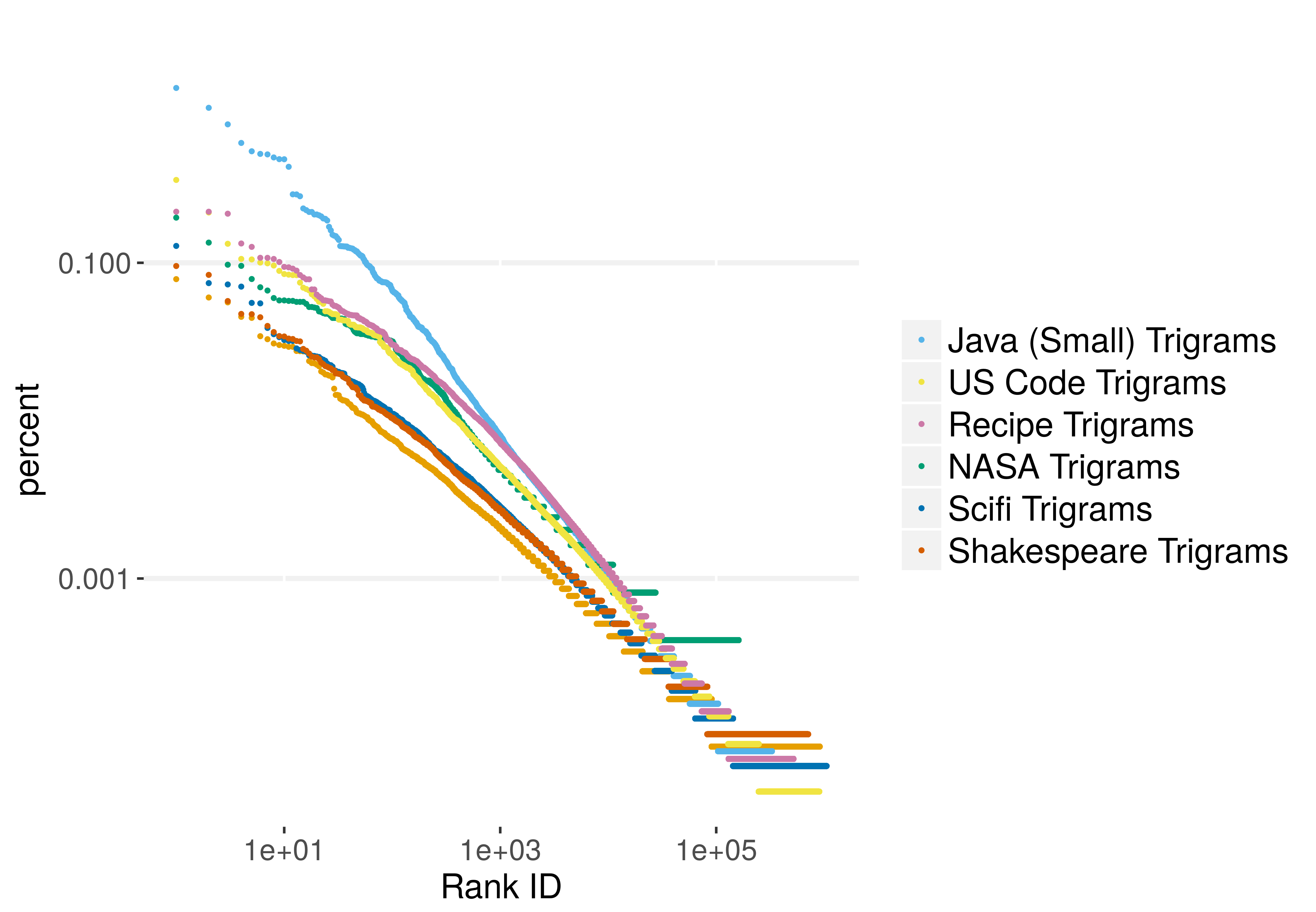}}
  \caption{Unigram, bigram, and trigram Zipf plot comparisons between the technical and imperative English corpora in comparison to the non technical English corpora and Java}
  \label{fig:TechZipf}
\end{figure}

 \begin{figure}[ht]
\centering
\subfloat[3 grams]{\includegraphics[width=.49\textwidth]{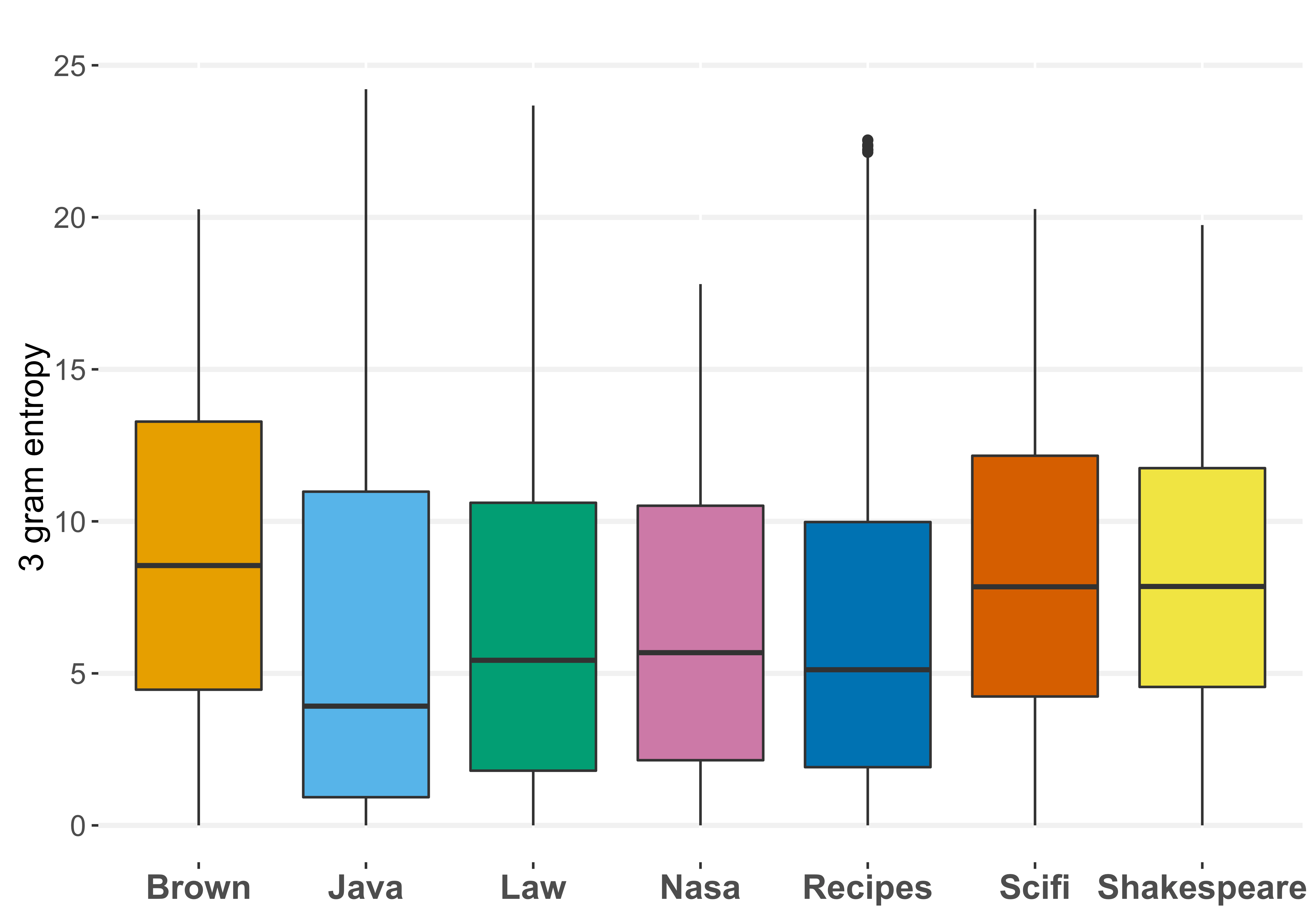}} 
\subfloat[3 grams with cache]{\includegraphics[width=.49\textwidth]{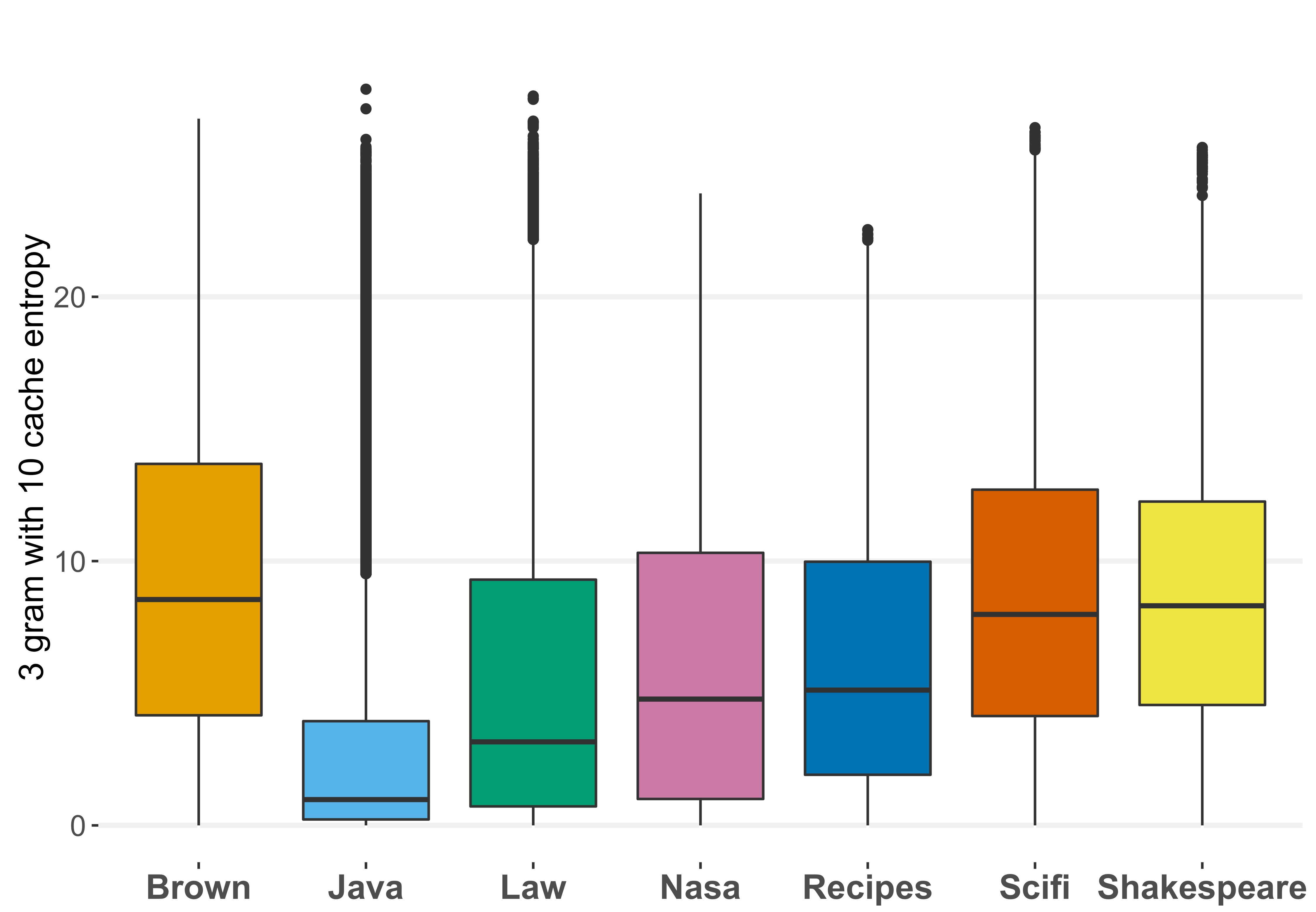}}\\
\subfloat[LSTM (Small)]{\includegraphics[width=.49\textwidth]{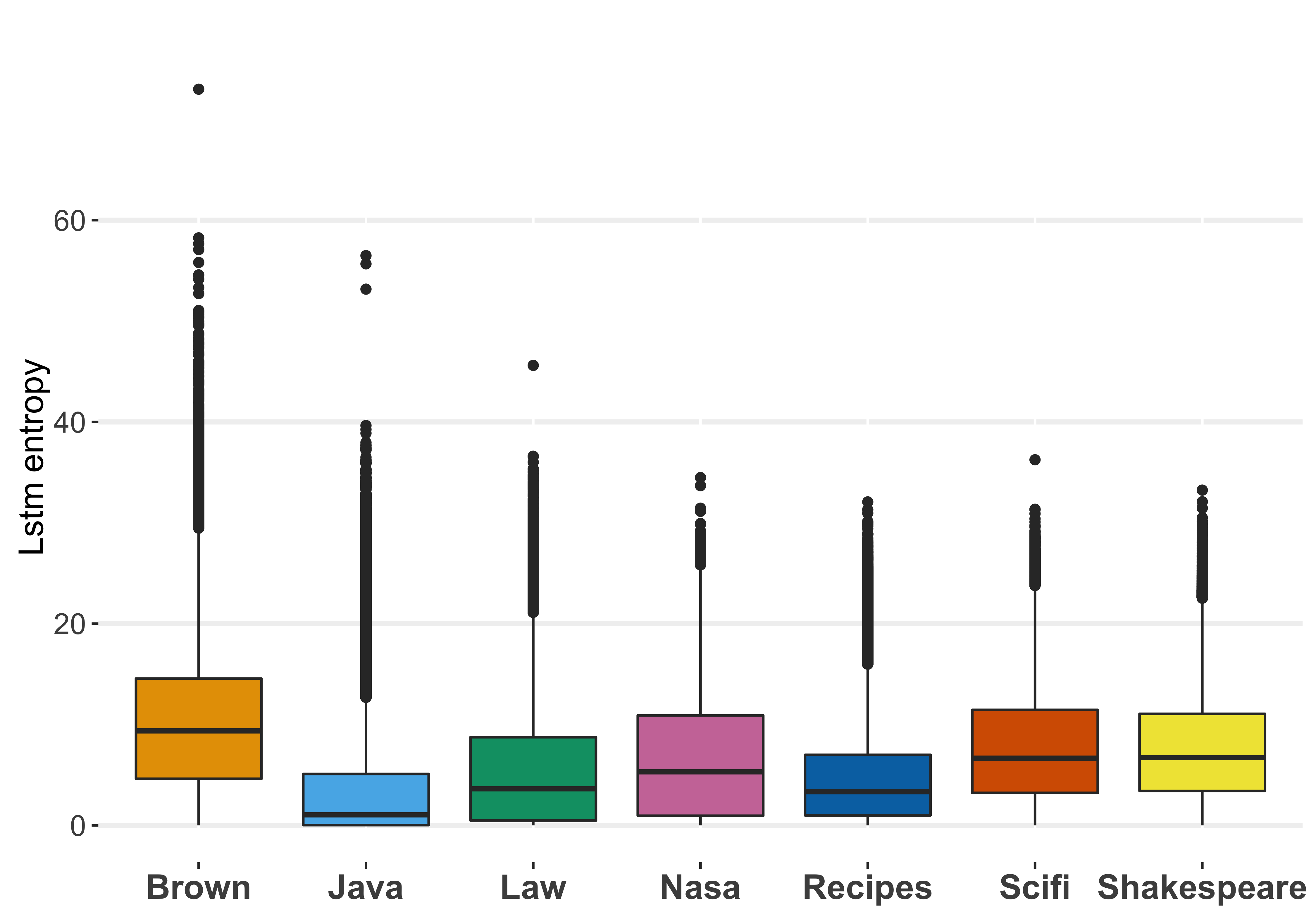}}
  \caption{Box plots of the distribution of entropy of the technical and imperative English corpora in comparison to the non technical English corpora and Java}
  \label{fig:TechEnt}
\end{figure}

Now, we compare technical and imperative English (such as law, recipes, or high-level requirements)
with non-technical English such as novels and plays; we expect technical English
to be more repetitive, since it's harder to read and write.  We also expect imperative English to be
more repetitive as to avoid ambiguity in communication.
Fig. \ref{fig:TechZipf} displays the unigram, bigram, and trigram Zipf curves for all of these corpora, the balanced English
Brown corpus, and our smaller sample of the Java corpus.  We see that once again, unigram slope behavior is
highly equivalent, but these slopes separate as the ngram length increases.  The science fiction novels and Shakespeare's
plays behave very similarly to the balanced Brown corpus.  The technical corpora fall between these nontechnical
English corpora and the Java code corpus, as we expected from our hypothesis.
The technical and imperative corpora of NASA directives, recipes, and US Code corpora exhibit more
code-like behavior than the Shakespeare, Science Fiction, and Brown
corpora.

In Fig. \ref{fig:TechEnt}, we verify these results with the ease of prediction via language model. 
The technical corpora are easier to predict than the non-technical corpora, but not as easy as the Java corpus,
regardless of which language model is used.
If we validate these distributions with Wilcox tests and effect sizes, shown in Table \ref{tab:JavaBrownTechEff}, 
which compare the effect size between brown and our other corpora, and Java and our other corpora.
We see that all corpora are more predictable than Brown, but that the non-technical corpora are proportionately much closer to the balanced
Brown corpus than the technical and imperative corpora.
Likewise, Java is significantly smaller than all corpora, but this effect size of this difference is sometimes small between
it and the technical english corpora.
In fact, with the best language models, the size of the difference between the median entropy of Java and both the corpus of US law and the corpus of recipes is only slightly over 1 bit.
In terms of confidence intervals, when using a cache or LSTM model, Java is about as twice as predictable as the these corpora.

\begin{table*}[ht]
\centering
\caption{Summary of non-parametric effect sizes and 99\% confidence intervals (in bits) comparing each technical and non-technical corpus with Brown and then Java.  Numbers are marked with * if $p < .05$, ** if $p < .01$, *** if $p < .001$ from a Mann Whitney U test}
\begin{tabular}{| c | c  c  c |}
\hline
Brown \textgreater Language&Ngram&Cache&LSTM\\ \hline
\multirow{2}{*}{NASA}  & (-2.514, -2.39) & (-2.96, -2.826) & (-3.374, -3.207) \\ 
 & $0.197^{***}$ & $0.224^{***}$ & $0.216^{***}$  \\ \hline
\multirow{2}{*}{Science Fiction}  & (-0.514, -0.421) & (-0.396, -0.295) & (-2.065, -1.949) \\ 
 & $0.045^{***}$ & $0.031^{***}$ & $0.158^{***}$  \\ \hline
\multirow{2}{*}{US Code}  & (-2.532, -2.456) & (-3.736, -3.655) & (-4.55, -4.457) \\ 
 & $0.219^{***}$ & $0.323^{***}$ & $0.343^{***}$  \\ \hline
\multirow{2}{*}{Shakespeare}  & (-0.592, -0.498) & (-0.391, -0.287) & (-2.157, -2.038) \\ 
 & $0.053^{***}$ & $0.031^{***}$ & $0.166^{***}$  \\ \hline
\multirow{2}{*}{Recipes}  & (-2.763, -2.683) & (-2.737, -2.651) & (-5.127, -5.028) \\ 
 & $0.268^{***}$ & $0.254^{***}$ & $0.426^{***}$  \\ \hline
    Language \textgreater  Java (Small)  &Ngram & Cache &LSTM \\ \hline
\multirow{2}{*}{NASA}  & (-0.636, -0.529) & (-2.237, -2.101) & (-2.478, -2.332) \\ 
 & $0.06^{***}$ & $0.253^{***}$ & $0.249^{***}$  \\ \hline
\multirow{2}{*}{Science Fiction}  & (-2.754, -2.67) & (-5.35, -5.257) & (-4.121, -4.037) \\ 
 & $0.247^{***}$ & $0.523^{***}$ & $0.457^{***}$  \\ \hline
\multirow{2}{*}{US Code}  & (-0.532, -0.468) & (-0.9, -0.868) & (-1.152, -1.09) \\ 
 & $0.061^{***}$ & $0.21^{***}$ & $0.204^{***}$  \\ \hline
\multirow{2}{*}{Shakespeare}  & (-2.801, -2.711) & (-5.651, -5.56) & (-4.202, -4.117) \\ 
 & $0.24^{***}$ & $0.519^{***}$ & $0.446^{***}$  \\ \hline
\multirow{2}{*}{Recipes}  & (-0.493, -0.431) & (-2.676, -2.603) & (-1.135, -1.085) \\ 
 & $0.063^{***}$ & $0.388^{***}$ & $0.255^{***}$  \\ \hline
\end{tabular}
\label{tab:JavaBrownTechEff}
\end{table*}

We also checked to see if there was any effect of a cache for the technical and non-technical corpora.
If technical language behaves like code, we would expect more local repetition, and hence improvements when
moving from an ngram to an ngram-cache model.
Table \ref{tab:TechModelEffect}, demonstrates confidence intervals and effect sizes for the cache improvements, with positive confidence intervals indicating an improvement over a basic ngram model.  
For our non-technical corpora, there is a extremely small negative effect on predictability when
using a cache, and no significant effect on the Brown corpus.
In comparison, the small Java corpus, the legal language corpus, and the NASA directive corpus all have significant increases in entropy when not including the cache,
though there is no cache effect in the recipe corpus.
This effect size extremely tiny in the NASA corpus, but is somewhat larger for the legal corpus.
This agrees with the notion of the restrictiveness of technical language, and especially that of legal language as the most restrictive technical language, as its
local repetitiveness allows a cache to improve about twice as much over the raw ngram score.
However, the cache effect in the legal corpus is still not as large as with the Java corpus.
These observations of cache behavior are consistent with any association with technical style rather than merely imperative texts, but a more in depth
experiment would be required to properly test this hypothesis.

\begin{table*}[ht]
\centering
\caption{Summary of non-parametric effect sizes and 99\% confidence intervals (in bits) comparing the locality effects of the cache in each language.   Positive values in the intervals indicate an improvement due to the cache, and negative values indicate worse performance compared to the pure ngram model. Numbers are marked with * if $p < .05$, ** if $p < .01$, *** if $p < .001$ from a paired Mann Whitney U test}  
  \begin{tabular}{| c |  c   |}
    \hline
    Language & Ngram $>$ Cache \\ \hline
     \multirow{2}{*}{Brown} & (-0.021,  0.049) \\ 
    & $0.001$   \\ \hline
    \multirow{2}{*}{Java} & (1.570, 1.631)  \\ 
    & $0.269^{***}$   \\ \hline
   \multirow{2}{*}{NASA} & (0.375, 0.523)  \\
   & $0.0566^{***}$  \\ \hline
      \multirow{2}{*}{Recipes} & (-0.030,  0.030)  \\
   & $0^{}$  \\ \hline
      \multirow{2}{*}{Science Fiction} & (-0.129, -0.028)  \\
   & $0.008^{***}$  \\ \hline
      \multirow{2}{*}{US Code} & (1.123, 1.180) \\
   & $0.138^{***}$  \\ \hline
      \multirow{2}{*}{Shakespeare} & (-0.255, -0.155)  \\
   & $0.021^{***}$  \\ \hline
  \end{tabular}
  \label{tab:TechModelEffect} 
\end{table*}

%% file: discussion.tex
Our study starts with the discovery first reported in~\citep{Hindle2012}, that software
is highly repetitive and predictable. While this is surprising in itself, the real surprise
is that it is \emph{far} more predictable than natural language; indeed, using the
perplexity measure, it's about 8 to 16 times more predictable. 
Why is this the case? 
Is it vocabulary? Syntax? Or something else? Does it depend on programming
language? Natural language? The type of corpora? 
While this paper does not provide a definitive identification of the exact reasons
for why code is \emph{so much} more predictable than English, we describe a series
of experiments that points in the direction of deliberate choice, rather language constraints, 
as the reason. 

First, we show that the differences observed between English and Java 
holds up with other natural and programming languages, and across different types
of natural language corpora. 
Programming language corpora in general are more repetitive than natural language
corpora.
Our models captured some common intuition about the programming languages, namely that Haskell's compact
design leads to more English-like texts -- that in fact it achieves in practice its goal of informationally dense content.

Next we address the question of whether the greater repetitiveness of code arises mainly
from the simpler syntax of code. To begin with, we remove the keywords operators and punctuation 
from code, and likewise the closed-category words and punctuation from English, and compare
the repetitiveness of the remaining content vocabulary, and find that in fact code gets \emph{more repetitive} when these syntactic markers are eliminated; thus suggesting that the additional
repetitiveness is not exclusively syntax-based. 
 
Diving deeper and examining the parse tree structure, we do find that some of the differences in predictability derive from differences between programming language and natural language syntax.
Once normalized for the number of expansions the grammar allows, writers of English and code choose among their immediate options equivalently in each language.
However, when accounting for all the available terminal choices and the long term operations,
code still remains more predictable than English. 
Thus, it seems while a significant portion of the difference between English and Java is determined by grammatical restrictions, these restrictions do not account for all of the difference.

We surmise that the residual differences between Code and English may arise from the 
greater difficulty of reading and writing code, and bring in for comparison English corpora
that might require greater effort: ESL (english as a second language) corpora, legal corpora, 
and NASA directives; we find that these are still less predictable than code, but more so than
English, thus constituting an intermediate level of predictability, as expected. Finally, we find
technical texts, such as both NASA directives and Law, along with a text of natural language instructions
are also likewise intermediate in predictability to code and English. 
Code is a unique form of human expression; as Allamanis \emph{et al} observe 
\citep{allamanis2017survey} it comprises two channels; one from human to human and the other from human to code. 
This dual-channel nature places special demands on readers and writers. That these specialized
English texts, where the style imposes greater effort and the cost of miscommunication is higher also become more code-like
is consistent with the theory that humans use repetitive but familiar structures to communicate
clearly under such constraints.

In a broader context, knowing that these differences between natural and programming languages come from choice rather than 
language constraints is useful.
It can provide theoretical grounding to choices when designing new languages and finding the right degree of
expressiveness.
After all, if humans mostly choose from a limited set of possible available constructs in code, then these
choices should impact how languages are created, documented, and taught.
Highlighting or including language options that are never used may be increase confusion and the potential for mistakes.
Likewise, this theory supports the notion that the limitations imposed by style are important for clear communication.
For example, existing research shows that pull requests that conform to project style are more readily accepted \citep{Hellendoorn2015}.
Finally, as it seems human choice does matter when it comes to programming, gaining a better grasp on how humans actually cognitively
process code is important.
As Knuth said, code is not merely for the machines~\citep{knuth1984literate}.